%% file: main.tex
\RequirePackage{fix-cm}
\documentclass[11pt]{article}

\usepackage[preprint]{acl}
\usepackage{times}
\usepackage[T1]{fontenc}
\usepackage{iftex}
\ifPDFTeX\usepackage[utf8]{inputenc}\fi
\usepackage{microtype,inconsolata,graphicx,booktabs,amsmath,amssymb,multirow}
\usepackage{placeins,tabularx,xcolor,colortbl}
\usepackage{bm}
\usepackage{enumitem}

\definecolor{dvgreen}{HTML}{E3F1EB}
\definecolor{dvyellow}{HTML}{FFF0BF}
\definecolor{dvlight}{HTML}{F0F3F7}

\newcommand{\method}{\textsc{DualView}}

\newcommand{\result}[2]{\ensuremath{#1_{\mbox{\fontsize{4.5}{5}\selectfont\textpm\,#2}}}}

\newcommand{\best}[2]{\ensuremath{\mathbf{#1}_{\mbox{\fontsize{4.5}{5}\selectfont\textpm\,#2}}}}

\title{\method{}: Leveraging Broad and Output-Specific Evidence for LLM-based Multi-Output Text Regression}

\author{Yiheng Su \hspace{2em} Junyi Jessy Li \hspace{2em} Matthew Lease \\
        The University of Texas at Austin \\
        \texttt{\{sam.su, jessy, ml\}@utexas.edu}}

\begin{document}

\maketitle

\begin{abstract}

While work on LLM-based regression has largely focused on scalar prediction, multi-output text regression remains less explored.
We present \method{}, a parameter-efficient approach to leverage LLMs for multi-output text regression. \method{} extracts broad and output-specific representations from a frozen LLM, then uses a dual-branch architecture to jointly predict numerical scores for all targets.
Across five regression tasks and eight LLMs (up to 122B), \method{} outperforms seven supervised regression baselines while requiring 82--1842$\times$ fewer fitted parameters than the three strongest baselines.
These improvements extend across individual output dimensions and persist even with limited training data.
\method{} atop a smaller open-weight LLM also outperforms unsupervised prediction with Gemini 3.1 Pro. Ablations show that broad and output-specific evidence provide complementary predictive benefits, while lightweight shared predictors achieve strong performance with fewer trainable parameters.

\end{abstract}

\section{Introduction}

Text regression aims to predict continuous numerical targets from textual inputs. While recent work has 
largely focused on predicting scalar targets \citep{nafar-etal-2025-learning, lukasik2025raft, su2026relish, chen2026genre2, akhauri2026regression, Zhang2026-mc, Zuo2026-jc, Chung2026-nj}, predicting multiple numerical targets for the same input---\emph{multi-output text regression} (MOTR)---remains less explored. 

\begin{figure}[t]
\centering\includegraphics[width=\columnwidth]{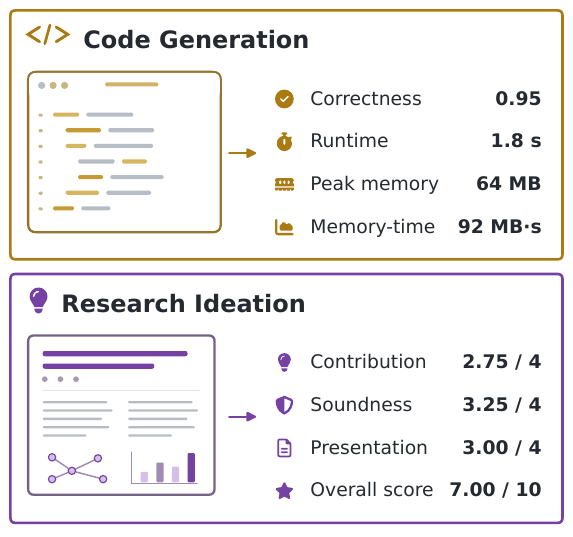}
\caption{\textbf{Multi-output text regression} predicts multiple numerical targets for the same textual artifact. Examples include code generation and research ideation, where the same program or paper is evaluated along dimensions with potentially different units and scales.}
\label{fig:task-overview}
\end{figure}

Building better MOTR systems can benefit fine-grained evaluations of generated text, where standard protocols assess free-form outputs along multiple dimensions such as fluency, factuality, coherence, and relevance \citep{liu2023geval, hashemi-etal-2024-llm, kim-etal-2025-biggen, chen-etal-2026-multi-agent}. More specialized tasks exhibit the same multi-output structure: machine translation can be evaluated for accuracy, fluency, and style \citep{popov-etal-2025-refined,shayegh-etal-2025-feeding}, generated code for correctness and efficiency \citep{ICLR2025_06694da0,ye2026problemoriented}, and research ideas for novelty, relevance, and feasibility \citep{ICLR2025_ea94957d,baek-etal-2025-researchagent}. \textbf{Figure~\ref{fig:task-overview}} illustrates the latter two examples.


However, developing general-purpose MOTR methods presents several challenges. First, existing text regression methods often fine-tune all LLM parameters, making them costly to scale to larger models \citep{lukasik2025raft,chen2026genre2}. 
Second, training one scalar predictor for each output scales adaptation costs with the number of targets, motivating more compact multi-output predictors \citep{do-etal-2024-autoregressive}.
Third, existing supervised MOTR methods often employ specialized architectures that may be less suitable across heterogeneous domains. For example, GAT-AES \citep{li-ng-2025-graph} uses handcrafted features and a fully connected graph that exploits strong correlations among essay-quality criteria. However, this architecture is less suitable for code quality estimation, where essay-specific features do not apply, and targets such as code correctness and peak memory may not be strongly correlated. 

\begin{figure*}[t]
\centering\includegraphics[width=\textwidth]{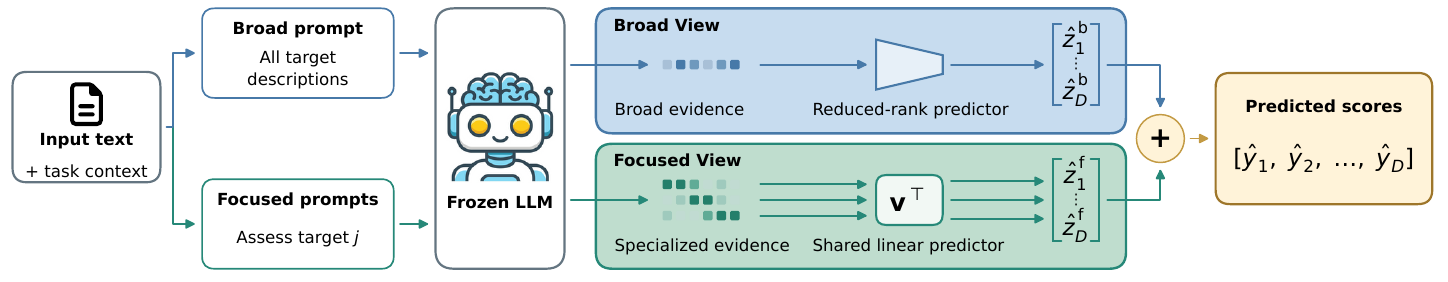}
\caption{\textbf{\method{} architecture.}
\method{} queries a frozen LLM with a broad prompt and focused prompts to extract complementary representations. 
The ``Broad View'' jointly predicts all output dimensions from one representation using a reduced-rank predictor. 
On the other hand, the ``Focused View'' predicts each output dimension individually using a shared linear predictor with target-specific representations.
The two branches predict standardized scores, which are summed and rescaled to obtain the final predictions.
}
\label{fig:architecture}
\end{figure*}

In this work, we propose \method{}, a supervised MOTR approach that shifts specialization from domain-specific prediction architectures to evidence extraction. Specifically, \method{} extracts complementary broad and output-specific representations from an LLM. A unified dual-branch model leverages these representations to predict all output dimensions, sharing predictors across targets instead of training separate models (see \textbf{Figure~\ref{fig:architecture}}). Finally, by freezing the LLM, \method{} trains only lightweight modules and scales better to larger LLMs.


We evaluate \method{} on five quality estimation tasks spanning summarization \citep{min-etal-2025-towards}, machine translation \citep{popov-etal-2025-refined}, code generation \citep{ouyang2026dscodebench}, dialogue \citep{NEURIPS2024_be2e1b68}, and research ideation \citep{lu-etal-2026-rethinking}. \method{} outperforms seven supervised state-of-the-art regression baselines across the five datasets and eight LLM backbones (8--122B). 
Relative to the second-best method, GAT-AES \citep{li-ng-2025-graph}, \method{} improves aggregate Pearson correlation by 18.4\% and NRMSE by 2.8\% (\textbf{Table~\ref{tab:overall}}). 
Output-level results further show that \method{} outperforms the strongest supervised baselines on nearly all targets (\textbf{Figure~\ref{fig:output-profiles}}).

By freezing the LLM and sharing predictors across targets, \method{} requires only 5.1K--57.4K fitted parameters, compared with 82--1842$\times$ more for the strongest supervised baselines (\textbf{Table~\ref{tab:parameters}}). Training-data efficiency experiments show that \method{} continues to outperform the strongest baselines even with limited supervision (\textbf{Figure~\ref{fig:training-efficiency}}). \method{} outperforms Gemini 3.1 Pro under zero- and few-shot LLMaaJ (\textbf{Table~\ref{tab:gemini-comparison}}).



\section{Related Work}\label{sec:related-work}

\subsection{LLM-Based Text Regression}
\label{sec:rw_regression}

Existing supervised, LLM-based text regression methods broadly fall into three families.
i) \emph{Autoregressive decoding} generates numerical values as text \citep{vacareanu2024words,song2025decoding,chen2026genre2}.
ii) \emph{Regression-aware inference} derives point estimates from the model's output distribution under a regression objective \citep{lukasik-etal-2024-regression,lukasik2025raft}.
iii) \emph{Predictive heads} map LLM hidden representations directly to continuous values \citep{tang2025understanding,su2026relish}.

Our work differs in three key respects. 1) While most supervised methods focus on scalar prediction, we study the joint prediction of multiple numerical targets. 2) Existing multi-output methods often introduce specialized prediction architectures such as target-specific experts and graph-based interactions \citep{wang-liu-2025-mes,li-ng-2025-graph}. In contrast, \method{} shifts specialization to evidence extraction, using simple predictors over complementary representations from a frozen LLM. 3) Recent supervised methods only evaluate LLMs with up to 32B parameters \citep{su2026relish, Zhang2026-mc}, whereas we experiment with LLMs up to 122B parameters.





\subsection{Fine-Grained Text Evaluation}
Fine-grained text evaluation (also known as multi-dimensional, multi-criteria, or multi-trait evaluation) assesses the same text input along multiple numerical axes \citep{fabbri-etal-2021-summeval,liu2023geval}. 
By assessing distinct aspects of quality, fine-grained evaluation helps identify specific weaknesses, reveal performance trade-offs, and guide targeted revisions \citep{hashemi-etal-2024-llm, kim-etal-2025-biggen, kola-etal-2026-saja}. However, obtaining human annotations or execution-based metrics (e.g., code correctness and peak memory) across multiple dimensions can be expensive, motivating automated evaluation methods that learn from limited supervision. We show that \method{} consistently outperforms supervised baselines even with as few as 64 training examples ($\S$\ref{sec:data-efficiency}).

\subsection{Multi-Task Learning}
Multi-task learning (MTL) trains a single model to perform multiple related tasks simultaneously \citep{ding-etal-2023-score,hu-etal-2025-impartial,hu-etal-2026-multi}. Multi-output text regression is closely related but imposes a more structured prediction problem. Whereas MTL may combine tasks with different inputs, outputs, or objectives, MOTR predicts the \textit{same} set of continuous targets from the \textit{same} text input.
This shifts the focus from knowledge transfer across heterogeneous tasks to balancing shared and target-specific information within the same prediction problem.
Regression-capable MTL methods such as UniSO-N \citep{tan2025uniso} and MT-ICL~\mbox{\citep{zhu2026multitask}} provide general mechanisms for sharing across tasks, but do not specifically exploit this same-input, fixed-target structure. 
In contrast, \method{} exploits this structure directly through a dual-branch architecture that leverages complementary evidence extracted from frozen LLMs. $\S$\ref{sec:related-extended} details additional comparisons.
 

\section{\method{}}
\label{sec:method}

\subsection{Problem Formulation}\label{sec:problem-formulation}
Let $\mathcal S=\{(x_i,\mathbf y_i)\}_{i=1}^N$ denote a training set of textual inputs $x_i\in\mathcal X$ paired with target vectors $\mathbf y_i=(y_{i1},\ldots,y_{iD})\in\mathbb R^D$.
Let $\mathcal Q=(d_1,\ldots,d_D)$ denote an ordered set of target descriptions, where $d_j$ specifies the meaning and numerical scale of target $j$.
Given $\mathcal Q$, multi-output text regression aims to learn a predictor $f_{\mathrm{REG}}:\mathcal X\to\mathbb R^D$ that jointly predicts all $D$ targets for each input.

Following \citet{tang2025understanding} and \citet{su2026relish}, we standardize each target using its training-set mean $\mu_j$ and standard deviation $\sigma_j$:
{
\setlength{\abovedisplayskip}{7pt}
\setlength{\belowdisplayskip}{7pt}
\setlength{\abovedisplayshortskip}{0pt}
\setlength{\belowdisplayshortskip}{0pt}
\begin{equation*}
    z_{ij}=\frac{y_{ij}-\mu_j}{\sigma_j},
    \qquad
    \widehat y_{ij}=\mu_j+\sigma_j\widehat z_{ij}.
\end{equation*}
}
All predictors are trained on standardized targets. At inference, predictions are rescaled by $\sigma_j$ and shifted by $\mu_j$ to restore their original scales.

\subsection{Evidence Extraction}
\label{sec:evidence-extraction}

Rather than tailoring architectures to each domain, \method{} applies simple predictors to complementary evidence extracted from a frozen LLM. 
We first extract a broad representation to capture evidence relevant to all targets. 
Since different targets may depend on different aspects of the input, we also extract output-specific representations to identify cues particularly relevant to each target.


Let $s(\mathcal Q)$ denote a shared task instruction specifying all prediction targets, and let $q(d_j)$ denote a short instruction requesting only an assessment of target $j$. For each input $x_i$, we construct a broad prompt and $D$ focused prompts:
{
\setlength{\abovedisplayskip}{7pt}
\setlength{\belowdisplayskip}{7pt}
\setlength{\abovedisplayshortskip}{0pt}
\setlength{\belowdisplayshortskip}{0pt}
\begin{equation*}
\begin{aligned}
P_i^b &= [s(\mathcal Q);x_i],\\
P_{ij}^f &= [s(\mathcal Q);x_i;q(d_j)].
\end{aligned}
\end{equation*}
}

For the broad view, we mean-pool model hidden states across all tokens of the broad prompt:
{
\setlength{\abovedisplayskip}{7pt}
\setlength{\belowdisplayskip}{7pt}
\setlength{\abovedisplayshortskip}{0pt}
\setlength{\belowdisplayshortskip}{0pt}
\begin{equation*}
\mathbf{J}_i
=
\frac{1}{|P_i^b|}
\sum_{t=1}^{|P_i^b|}\mathbf{h}_t(P_i^b)
\in\mathbb{R}^H.
\end{equation*}
}

For the focused view, we extract the hidden state at the last token of each focused prompt, immediately before generation:
{
\setlength{\abovedisplayskip}{7pt}
\setlength{\belowdisplayskip}{7pt}
\setlength{\abovedisplayshortskip}{0pt}
\setlength{\belowdisplayshortskip}{0pt}
\begin{equation*}
\mathbf{C}_{ij}
=
\mathbf{h}_{|P_{ij}^f|}(P_{ij}^f)
\in\mathbb{R}^H.
\end{equation*}
}
Here, we use the final hidden state rather than mean pooling because most tokens precede the target-specific instruction and are shared across targets. Averaging their representations thus could dilute the effect of the focused instruction.

\paragraph{Shared-prefix reuse.}
The broad and focused prompts share the task instructions,
target descriptions, and input. We therefore cache the
shared prefix's model states (e.g., attention KV pairs) and reuse them across prompts. Rather than independently evaluating $D+1$ complete prompts, we process the shared prefix once and evaluate only the remaining suffix of each prompt. See $\S$\ref{sec:app_implementation} for details on extraction and normalization.

\subsection{Broad Vector Predictor}\label{sec:broad-predictor}
The broad branch predicts all $D$ targets jointly from $\mathbf J_i\in\mathbb R^H$ using a linear predictor:
{
\setlength{\abovedisplayskip}{7pt}
\setlength{\belowdisplayskip}{7pt}
\setlength{\abovedisplayshortskip}{0pt}
\setlength{\belowdisplayshortskip}{0pt}
\begin{equation*}
    \widehat{\mathbf z}^{\,\mathrm b}_i
    = \mathbf W_{\mathrm b}^{\top}\mathbf J_i.
\end{equation*}
}
An unconstrained $\mathbf W_{\mathrm b}\in\mathbb R^{H\times D}$ effectively fits a separate linear predictor for each target. To obtain a compact multivariate predictor, we adopt reduced-rank regression \citep{IZENMAN1975248}, factorizing $\mathbf W_{\mathrm b}=\mathbf A\mathbf B$, where $\mathbf A\in\mathbb R^{H\times k}$ and $\mathbf B\in\mathbb R^{k\times D}$, with $k<D$. Each target $j$ uses a coefficient vector $\mathbf A\mathbf b_j$, where $\mathbf b_j$ is the $j$-th column of $\mathbf B$. Thus, all targets are predicted from the same $k$ linear projections of $\mathbf J_i$, but each target assigns different weights to these projections. This factorization provides a compact multivariate predictor by sharing $k$ linear projections across outputs.

\subsection{Shared Focused Predictor}\label{sec:focused-predictor}
The focused branch uses a shared linear predictor to estimate target $j$ from $\mathbf C_{ij}\in\mathbb R^H$:
{
\setlength{\abovedisplayskip}{7pt}
\setlength{\belowdisplayskip}{7pt}
\setlength{\abovedisplayshortskip}{0pt}
\setlength{\belowdisplayshortskip}{0pt}
\begin{equation*}
    \widehat z^{\,\mathrm f}_{ij}
    = \mathbf v^\top\mathbf C_{ij}.
\end{equation*}
}
Because $\mathbf C_{ij}$ captures evidence specific to target $j$, a shared coefficient vector $\mathbf v\in\mathbb R^H$ can produce distinct predictions for different output targets. Sharing one predictor also allows supervision from all targets to inform a common scoring function, while requiring fewer parameters than fitting separate predictors for each target.

Notably, the two branches serve complementary purposes: the broad branch learns common predictive directions across different targets, while the focused branch pools supervision across targets to score their specialized evidence.

\subsection{Training}\label{sec:training-inference}
We jointly train the broad and focused predictors to exploit their complementary evidence. The final prediction for target $j$ combines both branches:
{
\setlength{\abovedisplayskip}{7pt}
\setlength{\belowdisplayskip}{7pt}
\setlength{\abovedisplayshortskip}{0pt}
\setlength{\belowdisplayshortskip}{0pt}
\begin{equation*}
    \widehat z_{ij}
    =\mathbf J_i^\top\mathbf A\mathbf b_j
    +\mathbf v^\top\mathbf C_{ij}+a_j,
\end{equation*}
}
where $a_j$ is a target-specific intercept. We minimize regularized squared error on the combined predictions, subject to $\operatorname{rank}(\mathbf W_{\mathrm b})\leq k$:
{
\setlength{\abovedisplayskip}{7pt}
\setlength{\belowdisplayskip}{7pt}
\setlength{\abovedisplayshortskip}{0pt}
\setlength{\belowdisplayshortskip}{0pt}
\begin{equation*}
\begin{aligned}
    \min_{\mathbf W_{\mathrm b},\,\mathbf v,\,\mathbf a}\quad
    &\frac{1}{ND}\sum_{i=1}^{N}\sum_{j=1}^{D}
    (z_{ij}-\widehat z_{ij})^2\\
    &+\lambda_{\mathrm b}\|\mathbf W_{\mathrm b}\|_F^2
    +\lambda_{\mathrm f}\|\mathbf v\|_2^2,
\end{aligned}
\end{equation*}
}
where $\mathbf W_{\mathrm b}=\mathbf A\mathbf B$, $\mathbf a=(a_1,\ldots,a_D)$, and $\lambda_{\mathrm b},\lambda_{\mathrm f}$ control regularization. We optimize this objective through alternating minimization. Starting from a reduced-rank ridge solution with $\mathbf v=\mathbf 0$, we alternate between fitting the shared predictor to the broad predictor's residuals and refitting the broad predictor to the focused predictor's residuals. We repeat until convergence and select the rank and regularization parameters using the validation split.


\section{Experimental Setup}\label{sec:experimental-setup}

\paragraph{Models.} We assess \method{} over eight LLM backbones: Gemma 4 E4B and 31B \citep{gemmateam2026gemma4}, Qwen3.5 9B and 122B-A10B \citep{team2026qwen3}, Qwen3.6 27B, Llama 3.3 70B Instruct \citep{grattafiori2024llama}, Qwen2.5 72B Instruct \citep{qwen2024qwen25}, and GPT OSS 120B \citep{openai2025gptoss}. We freeze all LLM backbones and fit only lightweight regression predictors. We further compare against Gemini 3.1 Pro in $\S$\ref{sec:frontier-comparison}.


\paragraph{Datasets.} We evaluate \method{} on five fine-grained quality estimation tasks: MSumBench for summarization \citep{min-etal-2025-towards}, RATE for translation \citep{popov-etal-2025-refined}, CreativeEval for research ideation \citep{lu-etal-2026-rethinking}, DSCodeBench for code generation \citep{ouyang2026dscodebench}, and PRISM for dialogue \citep{NEURIPS2024_be2e1b68}. $\S$\ref{sec:datasets} reports dataset split sizes and target ranges (\textbf{Table~\ref{tab:dataset-splits}}).


\paragraph{Baselines.} We evaluate against seven supervised baselines: RELISH \citep{su2026relish}, GenRe2 \citep{chen2026genre2}, T-MES \citep{wang-liu-2025-mes}, GAT-AES \citep{li-ng-2025-graph}, SAJA \citep{kola-etal-2026-saja}, UniSO-N \citep{tan2025uniso}, and MT-ICL \citep{zhu2026multitask}. We use frozen LLMs for all supervised methods and adapt their prediction modules to the multi-output setting where necessary (\S~\ref{sec:baseline-adaptations}).
We also report zero- and four-shot LLM-as-a-judge prompting baselines.

\paragraph{Metrics.} We report Pearson correlation and normalized root mean squared error (NRMSE). Following prior work \citep{lin-etal-2023-counterfactual, NEURIPS2023_3c2b60a3, NEURIPS2024_2ee1c872}, we normalize RMSE by the training-set standard deviation to fairly compare across datasets and targets with different scales. We additionally report Spearman correlation in the appendix to assess rank-order agreement. $\S$\ref{sec:metric-definition} provides further details.

\begin{figure*}[!t]
\centering
\includegraphics[width=\textwidth]{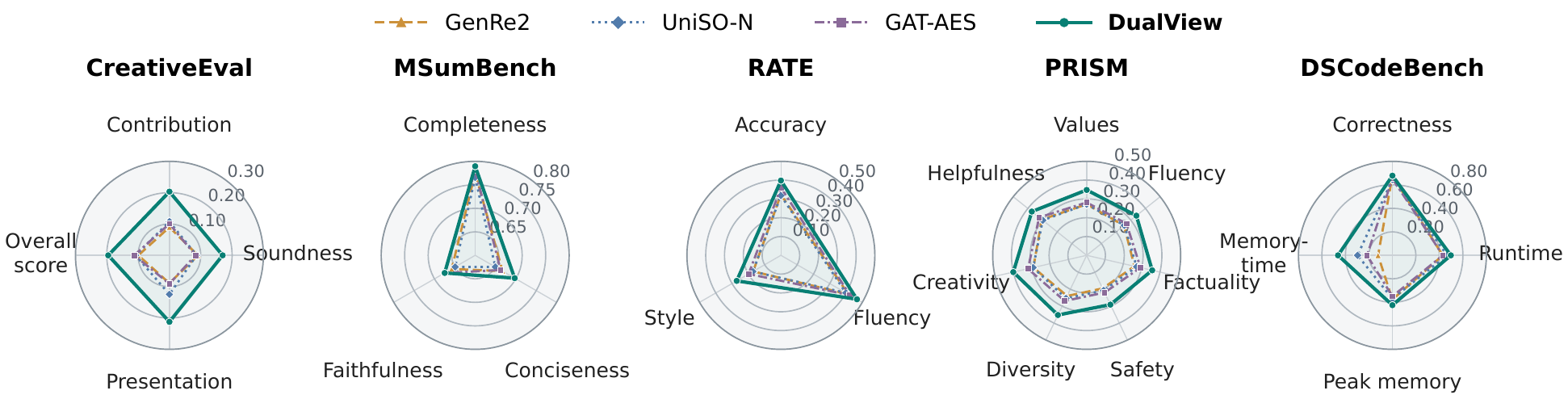}
\caption{\textbf{Predictive Performance across datasets and dimensions.} Each spoke reports Pearson correlation for one dataset aspect, macro-averaged over 8 LLM backbones and 3 training seeds. We compare \method{} against the three strongest supervised baselines. Larger radial values are better, where scales vary across datasets for visibility.}
\label{fig:output-profiles}
\end{figure*}

\section{Key Results}\label{sec:results}

\subsection{Predictive Performance}

\begin{table}[!t]
\centering
\small
\begingroup
\renewcommand{\arraystretch}{0.90}
\begin{tabular}{@{}l@{\hspace{1.4em}}c@{\hspace{1.4em}}c}
\toprule
\textbf{Method} & \textbf{NRMSE} $\bm{\downarrow}$ & \textbf{Pearson} $\bm{\uparrow}$ \\
\midrule
\multicolumn{3}{@{}l}{\textbf{Unsupervised}} \\
Zero shot & \result{17.197}{21.944} & \result{0.237}{0.013} \\
Few shot  & \result{61.673}{103.726} & \result{0.260}{0.025} \\
\midrule
\multicolumn{3}{@{}l}{\textbf{Supervised}} \\
SAJA     & \result{0.930}{0.070} & \result{0.263}{0.014} \\
T-MES    & \result{0.956}{0.065} & \result{0.297}{0.033} \\
RELISH   & \result{0.953}{0.070} & \result{0.340}{0.034} \\
MT-ICL   & \result{0.914}{0.063} & \result{0.348}{0.025} \\
GenRe2   & \result{0.908}{0.072} & \result{0.351}{0.024} \\
UniSO-N  & \result{0.924}{0.071} & \result{0.360}{0.042} \\
GAT-AES  & \result{0.898}{0.070} & \result{0.365}{0.032} \\
\textbf{\method{} (Ours)}
         & \cellcolor{dvgreen}\best{0.873}{0.073}
         & \cellcolor{dvgreen}\best{0.432}{0.038} \\
\bottomrule
\end{tabular}
\endgroup
\caption{\textbf{Predictive performance} aggregated across 8 LLMs and 5 datasets spanning 21 output targets. Values are macro-averaged over datasets and LLMs, with standard deviations across three seeds ($\pm$). Lower ($\downarrow$) NRMSE and higher ($\uparrow$) Pearson correlation are better.}
\label{tab:overall}
\end{table}

\paragraph{Aggregate performance.} \textbf{Table~\ref{tab:overall}} shows that \method{} achieves the strongest aggregate performance across five datasets and eight LLM backbones. Compared with GAT-AES, the strongest supervised baseline, \method{} improves Pearson correlation by 18.4\% and NRMSE by 2.8\%. The substantially higher NRMSE of zero- and few-shot prompting reflects large out-of-range predictions.

\paragraph{Performance across targets.}
Since our MOTR setting emphasizes predicting multiple targets per dataset, a robust method should ideally perform well on \textit{all} targets. While \textbf{Table~\ref{tab:overall}}'s results usefully summarize predictive performance on average, simple arithmetic averaging across target dimensions risks strong performance on some dimensions masking weak performance on others.
Thus, \textbf{Figure~\ref{fig:output-profiles}} compares \method{} with the three strongest supervised baselines across all 21 output targets and reports Pearson correlations macro-averaged over eight LLMs.
\method{} achieves the strongest performance on 20 of 21 targets, with MSumBench faithfulness effectively tied at 0.675. 

\paragraph{Performance across LLM--target pairs.}
Figure~\ref{fig:output-profiles} still averages performance across LLMs, which can hide inconsistent results for individual LLM--target pairs. \textbf{Table~\ref{tab:output-spread-main}} in $\S$\ref{sec:strict_wins} reports two additional fine-grained measures against the three strongest supervised baselines. The \textit{strict win rate} is the percentage of LLM--target pairs on which a method outperforms all competitors. The \textit{worst-output gap} is the largest performance difference from the best method across targets within each dataset--LLM setting, averaged across settings. \method{} achieves the highest win rates of 65.9\% for Pearson and 71.4\% for NRMSE, compared with 8.9\% and 7.7\% for GAT-AES. Our mean worst-output gaps are also smaller: 0.039 versus 0.149 for Pearson and 0.016 versus 0.055 for NRMSE. These results show that \method{} wins more consistently with smaller target-level gaps.

\method{} also outperforms all supervised baselines under aggregation schemes that place greater weight on poorly predicted targets (\textbf{Table~\ref{tab:power-mean-full}}). Detailed results for every dataset--LLM combination are reported in \textbf{Tables~\ref{tab:expanded-pearson}, \ref{tab:expanded-spearman}, and~\ref{tab:expanded-nrmse}} ($\S$\ref{sec:complete}), with complete breakdowns in $\S$\ref{sec:targets}.


\subsection{Trainable Footprint}
\label{sec:footprint}

\begin{table}[!t]
\centering
\small
\renewcommand{\arraystretch}{1.08}
\begin{tabular}{@{}lrr@{}}
\toprule
\textbf{Method} & \textbf{\# Params} & \textbf{Ratio} \\
\midrule
\textbf{\method{} (Ours)} & 5.1K--57.4K & -- \\
\midrule
GAT-AES  & 1.84M--4.73M  & 82--360$\times$ \\
GenRe2   & 3.89M--5.44M  & 93--779$\times$ \\
UniSO-N  & 9.44M--20.98M & 366--1842$\times$ \\
\bottomrule
\end{tabular}
\caption{\textbf{Trainable parameter counts} for \method{} and the three strongest supervised baselines. Ranges span 8 LLMs, 5 datasets, and 3 training seeds. Baseline ratios are computed relative to \method{} for each LLM, then macro-averaged.}
\label{tab:parameters}
\end{table}

\begin{figure*}[!t]
\centering
\includegraphics[width=\textwidth]{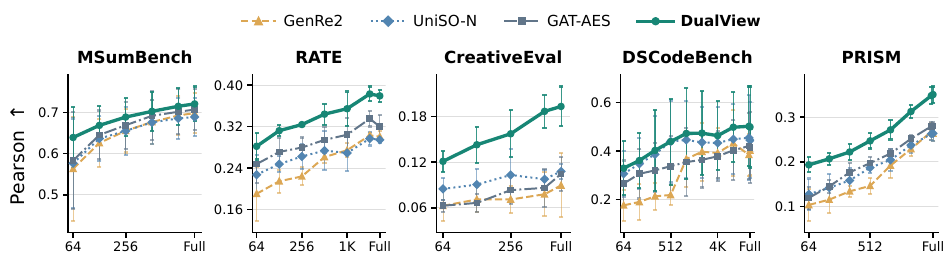}
\caption{\textbf{Predictive Performance as a function of training-set size.} Curves show Pearson correlation averaged over 8 LLMs within each seed, then averaged across three seeds, with whiskers denoting one standard deviation. We compare \method{} with the three strongest aggregate supervised baselines across five datasets. The horizontal axis is logarithmic, and Full denotes each dataset's complete training split. $\S$\ref{sec:data-efficiency} provides additional details.}
\label{fig:training-efficiency}
\end{figure*}

\begin{table*}[!ht]
\centering
\footnotesize
\renewcommand{\arraystretch}{0.90}
\setlength{\tabcolsep}{5pt}
\setlength{\aboverulesep}{0.25ex}
\setlength{\belowrulesep}{0.25ex}
\begin{tabular}{@{}lcccccc}
\toprule
\textbf{Dataset}
& \multicolumn{3}{c}{\textbf{NRMSE} $\bm{\downarrow}$}
& \multicolumn{3}{c}{\textbf{Pearson} $\bm{\uparrow}$} \\
\cmidrule(lr){2-4} \cmidrule(lr){5-7}
& \textbf{zero shot} & \textbf{few shot} & \textbf{\method{}}
& \textbf{zero shot} & \textbf{few shot} & \textbf{\method{}} \\
\midrule
CreativeEval & \result{1.976}{0.103} & \result{1.192}{0.086} & \cellcolor{dvgreen}\best{0.980}{0.043} & \result{0.099}{0.025} & \result{0.164}{0.028} & \cellcolor{dvgreen}\best{0.207}{0.030} \\
MSumBench    & \result{1.240}{0.126} & \result{1.021}{0.075} & \cellcolor{dvgreen}\best{0.682}{0.060} & \result{0.629}{0.062} & \result{0.630}{0.065} & \cellcolor{dvgreen}\best{0.729}{0.040} \\
RATE         & \result{1.147}{0.035} & \result{1.103}{0.039} & \cellcolor{dvgreen}\best{0.911}{0.020} & \result{0.331}{0.037} & \result{0.342}{0.028} & \cellcolor{dvgreen}\best{0.453}{0.020} \\
PRISM        & \result{1.208}{0.025} & \result{1.167}{0.059} & \cellcolor{dvgreen}\best{0.965}{0.019} & \result{0.162}{0.010} & \result{0.173}{0.030} & \cellcolor{dvgreen}\best{0.340}{0.022} \\
DSCodeBench  & \result{0.896}{0.294} & \result{0.888}{0.292} & \cellcolor{dvgreen}\best{0.771}{0.316} & \result{0.400}{0.147} & \result{0.433}{0.196} & \cellcolor{dvgreen}\best{0.548}{0.206} \\
\midrule
\textbf{Average} & \result{1.293}{0.084} & \result{1.074}{0.052} & \cellcolor{dvgreen}\best{0.862}{0.073} & \result{0.324}{0.027} & \result{0.348}{0.041} & \cellcolor{dvgreen}\best{0.455}{0.041} \\
\bottomrule
\end{tabular}
\caption{\textbf{Comparisons to Gemini 3.1 Pro.} We compare \method{} atop Gemma 4 31B (best open LLM) to Gemini 3.1 Pro under zero-shot and few-shot LLMaaJ across five datasets. Metrics are macro-averaged across all output dimensions within each dataset, with standard deviations across three seeds ($\pm$).}
\label{tab:gemini-comparison}
\end{table*}

\textbf{Table~\ref{tab:parameters}} shows that \method{} requires only 5.1K--57.4K fitted parameters, compared with 1.84M--20.98M for the three strongest supervised baselines (82--1842$\times$ more). Reduced-rank and shared predictors avoid fitting separate high-dimensional models for each target. Notably, the trainable footprint \textit{scales with representation width rather than backbone parameter count}, facilitating adaptation to larger LLMs. \textbf{Table~\ref{tab:parameters-absolute}} in $\S$\ref{sec:cost} provides detailed counts for all baselines across datasets and LLMs.

\subsection{Training-Data Efficiency}\label{sec:sample-efficiency-main}

We also investigate training-data efficiency by examining how predictive performance varies as the number of training examples increases. We compare \method{} against the three strongest supervised baselines and report Pearson correlation macro-averaged across all 8 LLM backbones and 3 seeds for each dataset. We refit methods on nested training subsets (e.g., larger subsets must contain all examples from smaller ones), starting at 64 instances and doubling the budget up to each dataset's full training split. We hold the validation and test splits fixed across budgets. 

\textbf{Figure~\ref{fig:training-efficiency}} shows that \method{} consistently outperforms the three strongest baselines at every evaluated training budget. 
\method{}'s lead on MSumBench narrows but remains stable or widens on the other four datasets. 
\textbf{Figures~\ref{fig:efficiency-pearson-all}, \ref{fig:efficiency-spearman-test}, and~\ref{fig:efficiency-nrmse-test}} ($\S$\ref{sec:data-efficiency}) compare \method{} against all seven supervised baselines for the three metrics.

\subsection{Comparison with Gemini 3.1 Pro}\label{sec:frontier-comparison}

\textbf{Table~\ref{tab:gemini-comparison}} compares \method{} using Gemma 4 31B with Gemini 3.1 Pro under zero-shot and few-shot prompting. Relative to the stronger few-shot condition, \method{} improves Pearson correlation and NRMSE by 30.7\% and 19.8\%, respectively. This result shows that supervised prediction with a smaller open LLM can outperform unsupervised prompting using a frontier proprietary LLM.

\section{Ablations}\label{sec:ablations}

$\S$\ref{sec:evidence-ablation} examines the complementarity of broad and focused evidence. $\S$\ref{sec:optimization-ablation} ablates other training strategies. $\S$\ref{sec:target-specificity-ablation} investigates the importance of target-specific evidence alignment. Finally, $\S$\ref{sec:sharing-ablation} assesses parameter sharing across outputs. Unless otherwise stated, ablation results are macro-averaged across five datasets and eight LLM backbones with standard deviations across three training seeds ($\pm$).

\begin{table}[!t]
\centering
\small
\begingroup
\renewcommand{\arraystretch}{0.95}
\begin{tabular}{@{}llcc}
\toprule
\textbf{Evidence} & \textbf{Predictor}
& \textbf{NRMSE} $\bm{\downarrow}$ & \textbf{Pearson} $\bm{\uparrow}$ \\
\midrule
Broad   & Reduced rank
        & \result{0.893}{0.073} & \result{0.383}{0.030} \\
        & Full rank
        & \result{0.892}{0.070} & \result{0.388}{0.026} \\
\addlinespace[2pt]
Focused & Shared
        & \result{0.912}{0.068} & \result{0.377}{0.033} \\
        & Independent
        & \result{0.893}{0.070} & \result{0.379}{0.028} \\
\midrule
\multicolumn{2}{@{}l}{\textbf{\method{} (Ours)}}
        & \cellcolor{dvgreen}\best{0.873}{0.073}
        & \cellcolor{dvgreen}\best{0.432}{0.038} \\
\bottomrule
\end{tabular}
\endgroup
\caption{\textbf{Evidence complementarity.}
We compare broad and focused evidence individually, using
both constrained and unconstrained predictors, against their
combination in \method{}.}
\label{tab:evidence-ablation}
\end{table}

\subsection{Evidence Complementarity}
\label{sec:evidence-ablation}

We first investigate whether broad and focused representations provide complementary predictive information. We compare the reduced-rank broad and shared focused predictors individually against their combination, using the same regularized squared-error objective. To assess whether single-view performance is limited by predictor capacity, we also evaluate full-rank broad regression and independent focused predictors for each target. All variants are refitted using their respective features.

\textbf{Table~\ref{tab:evidence-ablation}} shows that combining both evidence outperforms either alone, even with more capacity. Compared with the strongest single-view configuration (full-rank broad ridge), \method{} improves Pearson correlation by 11.3\% and reduces NRMSE by 2.1\%. These results support the complementarity of the two evidence views, suggesting that target-focused representations capture predictive information not fully captured by the broad view.

\begin{table}[!t]
\centering
\small
\begingroup
\renewcommand{\arraystretch}{0.90}
\setlength{\tabcolsep}{3pt}
\resizebox{\columnwidth}{!}{%
\begin{tabular}{@{}lcc}
\toprule
\textbf{Training strategy}
& \textbf{NRMSE} $\bm{\downarrow}$ & $\bm{r\uparrow}$ \\
\midrule
Separate fitting + $\alpha$ fusion
    & \result{0.880}{0.068} & \result{0.419}{0.026} \\
Fixed-basis joint fitting
    & \result{0.881}{0.069} & \result{0.414}{0.025} \\
\textbf{Alternating joint fitting (ours)}
    & \cellcolor{dvgreen}\best{0.873}{0.073}
    & \cellcolor{dvgreen}\best{0.432}{0.038} \\
\bottomrule
\end{tabular}%
}
\endgroup
\caption{\textbf{Training strategy ablations.}
We compare our alternating optimization to two simpler strategies.
Separate fitting and $\alpha$ fusion train both predictors
independently and combine their predictions using a
validation-tuned weight. Fixed-basis fitting holds the broad
basis $\mathbf{A}$ fixed while jointly optimizing broad
projection weights $\mathbf{B}$ and focused ridge weights
$\mathbf{v}$.}
\label{tab:optimization-ablation}
\end{table}

\subsection{Training Strategy Ablations}
\label{sec:optimization-ablation}

We also examine whether our alternating optimization improves over two simpler training strategies. \textit{Separate fitting + fusion} trains the broad and focused predictors independently, then combines their predictions using a validation-tuned weight $\alpha$. \textit{Fixed-basis joint fitting} first trains the broad predictor alone, then fixes its learned basis $\mathbf{A}$ while jointly fitting the target-specific weights $\mathbf{B}$ and focused predictor $\mathbf{v}$. All strategies use the same extracted representations and rank constraints.

\textbf{Table~\ref{tab:optimization-ablation}} shows that alternating joint fitting achieves the strongest performance, improving Pearson correlation by 3.1\% and 4.3\% over separate fitting and fixed-basis joint fitting, respectively, while reducing NRMSE by 0.8\% and 0.9\%. Notably, fixed-basis joint fitting does not improve over separate fitting, suggesting that joint optimization alone is insufficient. Instead, the results suggest that adapting the broad predictive directions to focused evidence helps the two branches make more complementary contributions.

\begin{table}[!t]
\centering
\small
\begingroup
\renewcommand{\arraystretch}{0.9}
\setlength{\tabcolsep}{3pt}
\resizebox{\columnwidth}{!}{%
\begin{tabular}{@{}llcc}
\toprule
\textbf{Evidence} & \textbf{Alignment}
& \textbf{NRMSE} $\bm{\downarrow}$ & $\bm{r\uparrow}$ \\
\midrule
Focused only & Mismatched
             & \result{0.923}{0.072} & \result{0.357}{0.032} \\
             & Matched
             & \result{0.912}{0.068} & \result{0.377}{0.033} \\
\addlinespace[2pt]
\textbf{*Broad + focused} & Mismatched
             & \result{0.889}{0.075} & \result{0.389}{0.031} \\
             & \textbf{*Matched}
             & \cellcolor{dvgreen}\best{0.873}{0.073}
             & \cellcolor{dvgreen}\best{0.432}{0.038} \\
\bottomrule
\end{tabular}%
}
\endgroup
\caption{\textbf{Target-specific evidence alignment.}
We compare matched and mismatched focused evidence,
with and without broad evidence. Mismatching assigns each
target another target's focused representation through a fixed
cyclic permutation. \method{} (*) uses both evidence views with matched semantics.}
\label{tab:target-specificity-ablation}
\end{table}

\subsection{Target-Specific Evidence Alignment}
\label{sec:target-specificity-ablation}

The focused view assumes that target-specific prompting extracts evidence particularly useful for the corresponding output. To test this, we introduce a ``mismatched'' control that assigns each target to another target's focused representation through cyclic permutation. We refit focused predictors using their permuted features with and without the broad predictor, applying the permutation consistently during training and evaluation.

\textbf{Table~\ref{tab:target-specificity-ablation}} shows that mismatched focused representations degrade predictive performance in both the focused-only and the dual-branch configurations. The performance loss is more pronounced in the combined setting: Pearson drops by 10.0\% versus 5.3\%, and NRMSE increases by 1.8\% versus 1.2\%, respectively. 

\begin{table*}[!t]
\centering
\small
\begingroup
\renewcommand{\arraystretch}{1.05}
\begin{tabular*}{\textwidth}{@{\extracolsep{\fill}}llcccc}
\toprule
\textbf{Broad} & \textbf{Focused}
& \textbf{Parameter Count}
& \textbf{Parameter Range}
& \textbf{NRMSE} $\bm{\downarrow}$
& $\bm{r\uparrow}$ \\
\midrule
Full rank & Independent
& $2HD+D$ & 15.4K--114.7K
& \result{0.874}{0.072} & \result{0.429}{0.038} \\

Full rank & Shared
& $HD+H+D$ & 10.2K--65.5K
& \result{0.873}{0.070}
& \cellcolor{yellow!20}\best{0.433}{0.031} \\

Reduced rank & Independent
& $k(H+D)+HD+D$ & 10.2K--106.5K
& \result{0.873}{0.071} & \result{0.432}{0.034} \\

\textbf{*Reduced rank} & \textbf{*Shared}
& $\bm{k(H+D)+H+D}$
& \cellcolor{dvgreen}\textbf{5.1K--57.4K}
& \cellcolor{dvgreen}\result{\textbf{0.873}}{0.073}
& \result{0.432}{0.038} \\
\bottomrule
\end{tabular*}
\endgroup
\caption{\textbf{Parameter-sharing ablations.}
We compare reduced- and full-rank broad predictors with shared
and independent focused predictors. \method{} (*) uses reduced-rank broad and shared focused predictors.}
\label{tab:sharing-ablation}
\end{table*}

\subsection{Sharing Across Outputs}
\label{sec:sharing-ablation}

Finally, we assess the accuracy--parameter trade-off introduced by the two forms of sharing in \method{}. 
We compare reduced-rank and full-rank broad predictors, crossed with shared and target-specific focused predictors. 
All configurations use the same extracted evidence and are fitted jointly using the $ L_2$-regularized objective in $\S$\ref{sec:training-inference}. 
\textbf{Table~\ref{tab:sharing-ablation}} shows that \method{} achieves comparable predictive performance to less constrained alternatives while fitting the fewest parameters.

Parameter sharing reduces the trainable footprint in two ways.  
First, the reduced-rank broad predictor uses $k(H+D)$ rather than $HD$ coefficients for an unrestricted predictor, with $k(H+D)<HD$ in all evaluated configurations.
Second, sharing the focused predictor reduces its coefficients from $HD$ to $H$. Including $D$ target-specific intercepts, \method{} requires only $(k + 1)(H+D)$ parameters, compared with $2HD+D$ for full-rank broad and independent focused predictors.

\section{Discussion and Future Work}

\paragraph{When does supervised adaptation pay off?}
Comparisons with Gemini raise a practical cost question beyond predictive performance. Direct prompting requires no task-specific training, whereas \method{} incurs an upfront fitting cost but can use a smaller open-weight LLM for repeated inference. In principle, if the supervised system has lower marginal inference cost, this upfront cost can be amortized over sufficient requests. For \method{}, however, the trade-off also depends on the number of outputs: processing $N$ inputs requires $N$ broad and $ND$ focused representations, although shared-prefix reuse avoids redundant backbone computation. The preferred configuration, potentially including hybrid strategies that balance latency and predictive performance, therefore depends on the practical deployment context.

\paragraph{Extensions to distribution elicitation.}
Our multi-output formulation may extend from predicting multiple semantic targets to predicting multiple points of a conditional distribution. 
\citet{vedula-etal-2025-quantile} explored LLM-based quantile regression for text-based price prediction, and \citet{zhu-etal-2026-text} extend it to general text-to-distribution settings using dense quantile grids. 
Specifically, for quantiles ${\tau_1,\ldots,\tau_N}$, a text input maps to a vector of quantile estimates $[q_{\tau_1}(x),\ldots,q_{\tau_N}(x)]$.
To leverage \method{}, we can treat each quantile level as a separate target and extract broad and quantile-specific representations from the LLM. Downstream ridge predictors can be applied as-is since they are domain-agnostic. Training could use pinball loss with a monotonicity constraint to prevent quantile crossing. We leave text-to-distribution extensions to future work.

\paragraph{Integrating multi-objective optimization.}
Multi-output regression can also be viewed through a multi-objective lens: \method{} uses the same $L_2$-regularized objectives across targets, but each target contributes separate loss terms that can induce conflicting updates to shared parameters. Multi-objective optimization (MOO) methods offer a natural way to address such conflicts by reweighting or modifying output-wise gradients during shared-parameter training \citep{sener2018multiobjective, efroni2025aligned, li-etal-2025-gradient}. MOO methods are thus complementary to \method{}, modifying how supervision is balanced across outputs without affecting how to extract features from the LLM backbone. Integrating MOO strategies is left for future work.


\paragraph{Transfer to unseen targets.} The shared focused predictor may generalize to unseen target descriptions without retraining. However, the broad predictor uses coefficients learned for a fixed output set, so extending the full \method{} model to new targets would require additional adaptation. This would be another interesting direction to pursue.

\section{Conclusion}
We proposed \method{}, a lightweight approach to multi-output text regression that leverages broad and output-specific representations from a frozen LLM. A rank-constrained multivariate predictor captures shared structure across outputs, while a shared scalar predictor uses evidence specialized to each target. Across five tasks and eight LLM backbones, \method{} outperforms seven supervised baselines while fitting substantially fewer parameters. These advantages extend across nearly all output dimensions and persist even with limited supervision. Ablations show that the two evidence views are complementary and that parameter sharing preserves performance with fewer parameters.

\clearpage

\section*{Limitations}


\paragraph{Sensitivity to prompt design.}
Because \method{} extracts representations from LLMs, the extracted vectors may be sensitive to changes in prompt formatting, which can affect predictions. We use fixed prompt templates for each dataset and do not evaluate alternative formulations.

\paragraph{Dependence on LLM representations.}
\method{} requires access to model internals to extract representations and thus may not work with black-box models. Because the LLM backbone is frozen, \method{}'s predictive performance also depends on the quality of its pretrained representations and may degrade on unfamiliar domains.

\paragraph{Nonconvex optimization.}
The rank-constrained joint objective specified in $\S$\ref{sec:training-inference} is nonconvex. Although each alternating update optimally fits one predictor while holding the other fixed, our optimization strategy does not guarantee a globally optimal joint solution.

\paragraph{Computational cost.}
While \method{} fits fewer parameters ($\S$\ref{sec:footprint}) versus supervised baselines, actual computational costs depend on the practical deployment environment. Specifically, \method{} requires one broad representation and $D$ focused representations per input. Although shared-prefix reuse avoids redundant computation, the $D$ focused suffixes still require separate processing. Thus, inference computation grows approximately linearly with the number of targets, and fewer fitted parameters do not necessarily translate into lower inference latency or memory use.


\bibliography{references}
\clearpage
\appendix


\section{Appendix Organization}

We organize the appendix as follows:

\begin{enumerate}
\item \textbf{Extended Related Work} ($\S$\ref{sec:related-extended}) further discusses LLM-based text regression methods, fine-grained text evaluation, and multi-task learning.

\item \textbf{Additional Method Details} ($\S$\ref{sec:app_implementation}) details evidence extraction, feature standardization, broad and focused predictors, and alternating training optimization.

\item \textbf{Additional Experimental Setup} ($\S$\ref{sec:setup-details}) describes the LLMs, datasets, baselines, evaluation metrics, and hyperparameters.

\item \textbf{Additional Key Results} ($\S$\ref{sec:additional-results}) provides more fine-grained predictive results,  parameter counts, training-data efficiency results, and comparisons with Gemini 3.1 Pro.

\item Additional Ablations ($\S$F) examines strong linear readout baselines and the accuracy--parameter-sharing trade-off in \method{}.

\item \textbf{Expanded Discussion} ($\S$\ref{sec:discussion-extended}) discusses multi-objective optimization as a possible extension of \method{}.

\item \textbf{Dataset Prompt Templates} ($\S$\ref{sec:dataset-prompts}) specifies the broad prompts and target-specific suffixes used for all five datasets.

\item \textbf{Output-Level Performance Tables} ($\S$\ref{sec:targets}) report NRMSE, Pearson, and Spearman results for each output, dataset, and LLM across all supervised methods.
\end{enumerate}

\section{Extended Related Work}\label{sec:related-extended}

\subsection{LLM-based Text Regression}\label{sec:rw-regression}

The three families described in $\S$\ref{sec:rw_regression} mainly categorize LLM-based text regression methods by how they obtain numerical predictions. Another important consideration is how they fine-tune the underlying LLM. Autoregressive decoding and regression-aware inference typically update the LLM through supervised fine-tuning or reinforcement learning that updates all model parameters, whereas predictive heads typically freeze the base LLM and only train the auxiliary modules.

Since \method{} freezes the LLM, we adopt the same setting for all supervised baselines. We therefore exclude methods whose primary contributions involve fine-tuning the LLM, since freezing would eliminate the central mechanism of their original formulations. Specifically, we acknowledge but exclude RAFT \citep{lukasik2025raft}, TRACT \citep{chiang-etal-2025-tract}, REAL \citep{Zhang2026-mc}, SmoothMMD \citep{Zuo2026-jc}, and Dist2Loss \citep{Chung2026-nj}.

\method{} uses lightweight regression modules trained atop frozen LLMs. This approach is motivated by prior work demonstrating the effectiveness of LLM hidden representations for downstream regression \citep{tang2025understanding}, recovering numerical predictive statistics directly from internal states \citep{piskorz2026eliciting}, and showing that numerical information encoded internally may not be reliably expressed through generated answers \citep{yuchi-etal-2026-llms}. Building on these findings, \method{} extends existing scalar text regression work using simple predictors based on extracted broad and target-specific representations.



\subsection{Fine-Grained Text Evaluation}\label{sec:rw-evaluation}

As a core application of \method{}, works like SummEval \citep{fabbri-etal-2021-summeval} and G-Eval \citep{liu2023geval} have long motivated assessing machine-generated text across multiple quality dimensions. 
One supervised baseline we compare to is SAJA \citep{kola-etal-2026-saja}, which learns to align judge-derived rubric features with human annotations using traditional machine learning models such as random forest. In contrast, \method{} directly uses prompt-conditioned hidden representations from a frozen LLM rather than generated rubric scores, extracting both broad and target-specific evidence for downstream regression.

Most supervised methods for fine-grained text evaluation focus on automated essay scoring (AES), where each essay receives scores for traits such as grammar, vocabulary, coherence, and fluency. Representative approaches include autoregressive trait-score generation \citep{do-etal-2024-autoregressive}, T-MES's trait-aware experts \citep{wang-liu-2025-mes}, and GAT-AES's graph modeling \citep{li-ng-2025-graph}. We adapt T-MES and GAT-AES to the frozen-backbone setting and evaluate them across heterogeneous domains. The key difference for \method{} is that we use simple predictors (ridge) that can work well across heterogeneous domains. Notably, \method{} shifts the burden of specialization to extracting complementary input representations (features) from frozen LLMs.


\subsection{Sharing Across Tasks and Outputs}\label{sec:rw-mtl}

Multi-task learning (MTL) typically entails learning one model to perform heterogeneous tasks. Examples include representation sharing across language-understanding tasks \citep{hu-etal-2025-impartial,hu-etal-2026-multi} and joint argument-span detection, type prediction, and quality scoring \citep{ding-etal-2023-score}. 

Multi-output text regression can be viewed as a structured instance of MTL in which every input has the same fixed set of continuous targets. This implies that information sharing between ``tasks" may be more intuitive, since output formats are homogeneous (e.g., the same numerical loss applies to all output dimensions). Nevertheless, it remains unclear how to share supervision across outputs while still allowing each prediction to retain information relevant to its specific target.

We compare with two regression-capable MTL approaches. MT-ICL learns a transformer-based Bayesian predictor from sequences of tasks \citep{zhu2026multitask}, whereas UniSO-N learns numerical surrogates across heterogeneous optimization tasks using language-model representations \citep{tan2025uniso}. \method{} differs by exploiting MOTR's shared-input structure through two complementary forms of sharing. Namely, a reduced-rank broad predictor learns common predictive patterns across outputs, while a shared predictor predicts individual targets from focused representations.


\section{Additional Method Details}\label{sec:app_implementation}

\subsection{Feature Extraction}\label{sec:feature-extraction}

Per $\S$\ref{sec:evidence-extraction}, for each input $x_i$, we construct a broad prompt $P_i^b=[s(\mathcal Q);x_i]$ and $D$ focused prompts $P_{ij}^f=[s(\mathcal Q);x_i;q(d_j)]$. We then extract representations:
\begin{equation*}
\mathbf J_i
=\frac{1}{|P_i^b|}\sum_{t=1}^{|P_i^b|}
\mathbf h_t(P_i^b),\qquad
\mathbf C_{ij}
=\mathbf h_{|P_{ij}^f|}(P_{ij}^f).
\end{equation*}
Both representations have shape $\mathbb R^H$ and use hidden states from the penultimate LLM layer (directly preceding the language modeling head). The broad view effectively mean-pools all prompt tokens, while the focused view retrieves the last-token hidden state after appending the target-specific instruction. Notably, during extraction, we do not autoregressively generate numerical answers. Our design of using pooled LLM representations for downstream regression follows prior work \citep{tang2025understanding, su2026relish}.

\paragraph{Shared-prefix reuse.}
By design, the focused prompt is just the broad prompt plus instructions to identify target-specific evidence. This means that we can cache the broad prompt's model activations (e.g., attention key--value pairs, recurrent states) to avoid redundant computation. Each focused representation then requires processing only the target-specific continuations.

\subsection{Evidence and Label Standardization}\label{sec:method-coordinates}

We standardize representations using training-split statistics. For the broad evidence, define
\begin{align*}
\mathbf m_J
&=\frac{1}{N}\sum_i\mathbf J_i,\\
\ell_J
&=\max\!\left\{\sqrt{\frac{1}{N}\sum_i
\|\mathbf J_i-\mathbf m_J\|_2^2},\epsilon_J\right\},\\
\widetilde{\mathbf J}_i
&=(\mathbf J_i-\mathbf m_J)/\ell_J.
\end{align*}
For focused evidence, one mean vector and one scalar scale are shared across targets:
\begin{align*}
\mathbf m_C
&=\frac{1}{ND}\sum_{i,j}\mathbf C_{ij},\\
\ell_C
&=\max\!\left\{\sqrt{\frac{1}{NDH}\sum_{i,j}
\|\mathbf C_{ij}-\mathbf m_C\|_2^2},\epsilon_C\right\},\\
\widetilde{\mathbf C}_{ij}
&=(\mathbf C_{ij}-\mathbf m_C)/\ell_C.
\end{align*}
The positive constants $\epsilon_J$ and $\epsilon_C$ prevent division by zero. For each target $j$, we also compute the training-label mean $\mu_j$ and population standard deviation $\sigma_j$, then set $z_{ij}=(y_{ij}-\mu_j)/\sigma_j$. All statistics are fixed at evaluation time. In the main-text equations, we omit the tildes on the preprocessed representations for readability.

\subsection{Rank-Constrained Broad Predictor}\label{sec:method-ridge}

Let $\mathbf X\in\mathbb R^{N\times H}$ contain the preprocessed broad representations $\widetilde{\mathbf J}_i^\top$ as rows. In each broad fitting step, let $\mathbf R\in\mathbb R^{N\times D}$ denote the residual targets after subtracting the current focused predictions. We solve
\begin{equation*}
\min_{\substack{\operatorname{rank}(\mathbf W_{\mathrm b})\leq k\\
\mathbf a\in\mathbb R^D}}
\|\mathbf R-\mathbf X\mathbf W_{\mathrm b}
-\mathbf 1\mathbf a^\top\|_F^2
+\gamma_{\mathrm b}\|\mathbf W_{\mathrm b}\|_F^2,
\end{equation*}
where $\gamma_{\mathrm b}=ND\lambda_{\mathrm b}$ and $\mathbf a$ is an unregularized, target-specific intercept vector. Since the columns of $\mathbf X$ are centered, the optimal intercept equals the column means of $\mathbf R$:
\begin{equation*}
\mathbf a=\frac{1}{N}\mathbf R^\top\mathbf 1,
\qquad
\mathbf R_c=\mathbf R-\mathbf 1\mathbf a^\top.
\end{equation*}
We then solve reduced-rank ridge on $\mathbf R_c$:
\begin{align*}
\mathbf C_\gamma
&=(\mathbf X^\top\mathbf X+\gamma_{\mathrm b}\mathbf I)^{-1}
\mathbf X^\top\mathbf R_c,\\
\mathbf K
&=\mathbf R_c^\top\mathbf X\mathbf C_\gamma.
\end{align*}
If $\mathbf Q_k\in\mathbb R^{D\times k}$ contains the leading orthonormal eigenvectors of $\mathbf K$, a reduced-rank ridge solution is
\begin{equation*}
\mathbf A=\mathbf C_\gamma\mathbf Q_k,
\qquad
\mathbf B=\mathbf Q_k^\top,
\qquad
\mathbf W_{\mathrm b}=\mathbf A\mathbf B.
\end{equation*}
Here $\mathbf A\in\mathbb R^{H\times k}$ learns $k$ common predictive directions and $\mathbf B\in\mathbb R^{k\times D}$ assigns weights to them for each target. The broad contribution to the standardized prediction is
\begin{equation*}
\widehat{\mathbf z}^{\,\mathrm b}_i
=\mathbf B^\top\mathbf A^\top
\widetilde{\mathbf J}_i\in\mathbb R^D.
\end{equation*}

\paragraph{Rank and storage.}
The reduced-rank factorization stores $k(H+D)$ coefficients, versus $HD$ for a full-rank broad predictor. This reduces model complexity when $k(H+D)<HD$, as we empirically confirm in our evaluated configurations.

\subsection{Shared Focused Predictor}\label{sec:method-focused}

The focused branch uses one linear coefficient vector $\mathbf v\in\mathbb R^H$ for all targets:
\begin{equation*}
\widehat z_{ij}^{\,\mathrm f}
=\mathbf v^\top\widetilde{\mathbf C}_{ij}.
\end{equation*}
Because the focused representations are conditioned on target instructions, the same predictor can assign different scores to different targets. The shared coefficients are estimated from all $ND$ example--target pairs.

With the broad predictor and intercepts fixed, the residuals are:
\begin{equation*}
e_{ij}=z_{ij}-\widetilde{\mathbf J}_i^\top
\mathbf A\mathbf b_j-a_j.
\end{equation*}
The focused fitting step solves
\begin{equation*}
\min_{\mathbf v}\ \sum_{i,j}
(e_{ij}-\mathbf v^\top\widetilde{\mathbf C}_{ij})^2
+\gamma_{\mathrm f}\|\mathbf v\|_2^2,
\end{equation*}
where $\gamma_{\mathrm f}=ND\lambda_{\mathrm f}$. Its ridge solution is
\begin{equation*}
\mathbf v=\left(\sum_{i,j}
\widetilde{\mathbf C}_{ij}\widetilde{\mathbf C}_{ij}^\top
+\gamma_{\mathrm f}\mathbf I\right)^{-1}
\sum_{i,j}\widetilde{\mathbf C}_{ij}e_{ij}.
\end{equation*}
Thus, the focused branch has $H$ fitted coefficients, rather than $HD$ for separate target-specific linear predictors.

\subsection{Training and Inference}\label{sec:method-procedure}

\paragraph{Joint objective.}
The standardized prediction combines the two branches additively:
\begin{equation*}
\widehat z_{ij}
=\widetilde{\mathbf J}_i^\top\mathbf A\mathbf b_j
+\mathbf v^\top\widetilde{\mathbf C}_{ij}+a_j.
\end{equation*}
As specified in $\S$\ref{sec:training-inference}, we minimize
\begin{equation*}
\begin{aligned}
\min_{\substack{\mathbf W_{\mathrm b},\,\mathbf v,\,\mathbf a\\
\operatorname{rank}(\mathbf W_{\mathrm b})\leq k}}
\quad&\frac{1}{ND}\sum_{i,j}(z_{ij}-\widehat z_{ij})^2\\
&+\lambda_{\mathrm b}\|\mathbf W_{\mathrm b}\|_F^2
+\lambda_{\mathrm f}\|\mathbf v\|_2^2,
\end{aligned}
\end{equation*}
with $\mathbf W_{\mathrm b}=\mathbf A\mathbf B$.

\paragraph{Alternating residual fitting.}
We initialize $\mathbf v=\mathbf 0$ and fit the reduced-rank broad predictor to the standardized targets. We then alternate two ridge updates: (i) fit $\mathbf v$ to the residuals of the current broad predictor and intercepts, using the solution in $\S$\ref{sec:method-focused}; and (ii) subtract the current focused predictions from the standardized targets,
\begin{equation*}
R_{ij}=z_{ij}-\mathbf v^\top\widetilde{\mathbf C}_{ij},
\end{equation*}
then refit $\mathbf A$, $\mathbf B$, and $\mathbf a$ using the rank-constrained solution in $\S$\ref{sec:method-ridge}. Both steps minimize the same combined objective with the other branch fixed. We repeat until convergence and select $k$, $\lambda_{\mathrm b}$, and $\lambda_{\mathrm f}$ on the validation split.

\paragraph{Inference.}
During inference, we restore the predicted standardized scores back to their original target units:
\begin{equation*}
\widehat y_{ij}=\mu_j+\sigma_j\widehat z_{ij}.
\end{equation*}

\section{Additional Experimental Setup}\label{sec:setup-details}

\subsection{Models}\label{sec:extraction-records}

We evaluate \method{} and baselines on eight frozen LLM backbones: Gemma 4 E4B Instruct \footnote{\url{https://huggingface.co/google/gemma-4-E4B-it}} and Gemma 4 31B Instruct\footnote{\url{https://huggingface.co/google/gemma-4-31B-it}} \citep{gemmateam2026gemma4}, Qwen3.5 9B\footnote{\url{https://huggingface.co/Qwen/Qwen3.5-9B}} and Qwen3.5 122B-A10B\footnote{\url{https://huggingface.co/Qwen/Qwen3.5-122B-A10B}} \citep{team2026qwen3}, Qwen3.6 27B\footnote{\url{https://huggingface.co/Qwen/Qwen3.6-27B}}, Llama 3.3 70B Instruct\footnote{\url{https://huggingface.co/meta-llama/Llama-3.3-70B-Instruct}} \citep{grattafiori2024llama}, Qwen2.5 72B Instruct\footnote{\url{https://huggingface.co/Qwen/Qwen2.5-72B-Instruct}} \citep{qwen2024qwen25}, and GPT OSS 120B\footnote{\url{https://huggingface.co/openai/gpt-oss-120b}} \citep{openai2025gptoss}.

\subsection{Datasets}\label{sec:datasets}
\textbf{Table~\ref{tab:dataset-splits}} reports the partitions for each dataset. \textbf{Tables~\ref{tab:prompt-creativeeval}--\ref{tab:prompt-dscodebench}} in $\S$\ref{sec:dataset-prompts} give the broad and focused prompt templates for these datasets. 

\paragraph{Partitioning} Several benchmarks contain multiple model outputs associated with the same underlying source. Randomly partitioning individual examples could therefore place related instances in both training and evaluation sets. To avoid this form of overlap, we construct the training, validation, and test partitions at the source-group level. Specifically, we partition examples by programming problem (DSCodeBench), source document (MSumBench), source segment (RATE), conversation (PRISM), or paper (CreativeEval), keeping all examples from each group in the same split. 

We repeat experiments using three random seeds (42, 1234, and 2026). For each seed, we independently construct train, validation, and test partitions at the source-group level, following the grouping strategy described above. All supervised methods use identical partitions within each seed to ensure matched comparisons. Counts are identical across seeds.

Since \method{} uses closed-form regression updates with a fixed initialization, its fitting procedure is deterministic given the extracted representations and data partitions. Thus, variation across seeds reflects differences in data partitioning rather than stochastic optimization of the proposed predictor for \method{}. For the other supervised baselines, however, differences might reflect both partitioning and stochastic optimization from gradient descent. 

\begin{table*}[t]
\centering\small
\setlength{\tabcolsep}{4pt}
\renewcommand{\arraystretch}{1.15}
\begin{tabularx}{\textwidth}{@{}lrrr>{\raggedright\arraybackslash}p{0.30\textwidth}>{\raggedright\arraybackslash}X@{}}
\toprule
\textbf{Dataset} & \textbf{Train} & \textbf{Validation} & \textbf{Test} & \textbf{Outputs} & \textbf{Range} \\
\midrule
CreativeEval & 718 & 103 & 205 & 4: Contribution, Soundness, Presentation, Overall & $[1,4]$ for Contribution, Soundness, Presentation\newline $[1,10]$ for Overall \\
\addlinespace[3pt]
MSumBench & 1,575 & 225 & 450 & 3: Completeness, Conciseness, Faithfulness & All $[0,1]$ \\
\addlinespace[3pt]
RATE & 2,783 & 400 & 792 & 3: Accuracy, Fluency, Style & All $[0,100]$ \\
\addlinespace[3pt]
PRISM & 4,331 & 619 & 1,237 & 7: Values, Fluency, Factuality, Safety, Diversity, Creativity, Helpfulness & All $[1,100]$ \\
DSCodeBench & 17,211 & 2,472 & 2,346 & 4: Correctness, Runtime, Memory, Integral & $[0,1]$ for Correctness\newline $[0,\infty)$ for Runtime, Memory, Integral \\
\addlinespace[3pt]
\bottomrule
\end{tabularx}
\caption{\textbf{Dataset split sizes and outputs.} Multiple candidate outputs associated with a source group are assigned to the same partition to avoid leakage. For DSCodeBench, Memory denotes peak resident memory, and Integral denotes its time integral.}
\label{tab:dataset-splits}
\end{table*}


\paragraph{MSumBench \citep{min-etal-2025-towards}.}
MSumBench\footnote{\url{https://github.com/DISL-Lab/MSumBench}} contains machine-generated summaries in English and Chinese across six domains: CNN/DailyMail news, GovReport reports, PubMed medical literature, MultiWOZ booking dialogue, MeetingBank meetings, and MediaSum interviews. 

The dataset provides three human-annotated targets: \textit{completeness}, the fraction of annotated source key facts covered by the summary; \textit{conciseness}, the fraction of summary sentences containing at least one source key fact; and \textit{faithfulness}, the fraction of summary sentences without factual errors. All fractions are in 0--1. 

\paragraph{RATE \citep{popov-etal-2025-refined}.}
The WMT24 English--Russian task\footnote{\url{https://huggingface.co/datasets/yandex/wmt24-en-ru-rate}} predicts expert assessments of a candidate translation along human-annotated dimensions of accuracy, fluency, and style, each on a 0--100 scale. Accuracy concerns preserving source meaning, fluency concerns the quality of the Russian translation, and style concerns preserving the source's manner of expression. 

\paragraph{CreativeEval \citep{lu-etal-2026-rethinking}.}
CreativeEval\footnote{\url{https://github.com/lichun-19/creative_eval}} contains human-written introductions from ICLR 2024 submissions. The authors form a set of 513 accepted and 513 rejected papers and match them across 16 soundness/presentation bins. The target dimensions are real average human reviewer scores spanning contribution, soundness, and presentation scores on a 1--4 scale, together with an overall score on a 1--10 scale. Notably, averaging reviewer scores yields continuous targets even when individual judgments use a discrete Likert scale. 

\paragraph{DSCodeBench \citep{ouyang2026dscodebench}.}
DSCodeBench\footnote{\url{https://github.com/ShuyinOuyang/DSCodeBench}} asks a model to implement a Python solution to a realistic data-science programming problem derived from GitHub. 
We collect execution-based targets for this study using a sandbox during dataset construction. Our motivation is to evaluate whether we can predict machine-generated code behavior without execution at inference. 

Targets include the fraction of unit tests passed (from a fixed
200-test suite), wall-clock time (seconds), peak sampled
process-tree resident memory (bytes), and the time integral
of resident memory (byte-seconds). We sample the summed resident set size (RSS) at a nominal interval of 5 ms, taking the maximum sampled RSS as peak memory and using trapezoidal integration to compute the memory integral.

Correctness lies in $[0,1]$, while the resource measurements
are nonnegative. We impose a 300-second execution timeout. We drop programs with execution exceptions and programs that execute beyond the 300-second timeout to avoid artificially inflating or deflating the measured wall-clock. We apply all preprocessing rules before partitioning.

\paragraph{PRISM \citep{NEURIPS2024_be2e1b68}.}

PRISM\footnote{\url{https://github.com/HannahKirk/prism-alignment}} evaluates an AI assistant's first-turn response to unguided, values-guided, and controversy-guided conversations. Responses are evaluated along seven dimensions on a 1--100 scale: 

\begin{enumerate}[topsep=2pt,itemsep=2pt,parsep=0pt,partopsep=0pt]
    \item \textbf{Values}: the extent to which the response reflected the participant's values or cultural perspective
    \item \textbf{Fluency}: the extent to which the response was well written and coherent
    \item \textbf{Factuality}: the extent to which the response was factual and informative
    \item \textbf{Safety}: the extent to which the response was safe and did not risk harm to the participant or others
    \item \textbf{Diversity}: the extent to which the response summarized multiple viewpoints
    \item \textbf{Creativity}: the extent to which the response was creative and inspiring
    \item \textbf{Helpfulness}: the extent to which the response was helpful and relevant.
\end{enumerate}

For regression, each retained conversation contributes
its selected top-rated first-turn response and the corresponding
seven performance-attribute ratings. The dataset only releases the seven-dimensional annotations for the response to the opening message, and not for the full conversation or every candidate response. We retain only examples with an available selected response and valid ratings for all seven attributes, excluding examples with missing ratings.

We partition examples by conversation, ensuring that each
conversation appears in only one split. Participant-level
disjointness is not enforced, so the evaluation measures
generalization to unseen conversations, potentially involving
the same participants represented in the training data.

\subsection{Baseline Adaptations}\label{sec:baseline-adaptations}

Because the supervised baselines differ in their original training procedures (supervised fine-tuning of all LLM parameters) and prediction settings (mostly scalar), we adapt them to operate with frozen LLM backbones and multi-output text regression targets.

\paragraph{RELISH \citep{su2026relish}.}
RELISH iteratively refines a learned latent state through cross-attention over frozen LLM token representations and maps the final state to a \textit{scalar} numerical prediction. To support joint prediction of multiple target dimensions, we replace the scalar linear readout with a $D$-dimensional linear readout. All other components and loss functions are not changed.

\paragraph{GenRe2 \citep{chen2026genre2}.}
GenRe2 trains autoregressive numerical decoders for \textit{scalar} regression using supervised initialization followed by sequence-level reinforcement learning. While most experiments fine-tune the LLM backbone, we note that its code-regression experiment already freezes the LLM encoder, making GenRe2 amenable to our frozen LLM setting. 
To support multi-output settings, we train a compact decoder on pooled broad-prompt features and condition it on output identity. One shared decoder predicts all outputs, with a rank-16 residual adapter for each output after each decoder block to permit specialization. We use normalized base-2 numerical tokens and median aggregation of sampled predictions.

\paragraph{T-MES \citep{wang-liu-2025-mes}.}
T-MES is designed for multi-trait automated essay scoring with a shared pretrained text encoder, token-level mixture-of-experts routing, and trait-specific scalar readouts. Its objective combines regression with trait-consistency, representation-correlation, and expert-diversity terms. We only substitute frozen broad-prompt LLM token states for its essay encoder representations since its original multi-output structure carries over directly.
We also adapt the output scalar readouts to our datasets' target ranges using scaled sigmoid predictions for bounded outputs and softplus predictions for nonnegative unbounded outputs.

\paragraph{GAT-AES \citep{li-ng-2025-graph}.}
GAT-AES is also designed for multi-trait essay scoring by passing messages among essay-embedding nodes, handcrafted-feature nodes, and learned trait nodes. 
Each training example supplies one embedding node obtained by projecting the last-token state of the broad prompt, with edges from this node to the output nodes and between output nodes. We omit the essay-specific handcrafted-feature nodes. 
We freeze the LLM while updating the other components, including the projection module, graph layers, and scalar readouts for each output dimension. 
As in T-MES, output links accommodate our bounded and nonnegative unbounded targets.

\paragraph{SAJA \citep{kola-etal-2026-saja}.}
SAJA aligns LLM judgments with human labels by prompting for a structured vector of rubric scores and fitting a lightweight calibration model. Thus, SAJA already operates without updating the LLM or accessing its hidden states. We preserve this generated-score interface and use its random-forest regression approach. 
Specifically, each forest shares the same extracted inputs but has separate parameters. 

\paragraph{UniSO-N \citep{tan2025uniso}.}
UniSO-N functions as a numerical surrogate for offline black-box optimization across heterogeneous generative tasks. Its numerical-regression variant maps language-model embeddings to a scalar using a shared MLP, where the embedder is fine-tuned with contrastive and Lipschitz objectives. After freezing the LLM, the embedder is no longer updated. Like RELISH, we replace the final projection layer of the scalar MLP with a $D$-dimensional linear projection to avoid fitting separate models for each output target.

\paragraph{MT-ICL \citep{zhu2026multitask}.}
Multi-Task Bayesian In-Context Learning trains a numerical Transformer to predict from labeled examples of a target task and auxiliary tasks. The original work studies synthetic inference problems and spatiotemporal regression. 
To apply MT-ICL to text, we replace numerical covariates with frozen output-specific LLM representations and treat each output dimension as a task. 
One shared Transformer serves every output and returns its predictive mean in the original target units. 

\subsection{Metrics and Aggregation}\label{sec:metric-definition}
We report normalized root mean squared error (NRMSE), Pearson $r$, and Spearman $\rho$. Because the criteria use different units and score scales, raw RMSEs are not directly comparable. We therefore normalize RMSE by the training standard deviation, following prior work \citep{NEURIPS2023_3c2b60a3, NEURIPS2024_2ee1c872,lin-etal-2023-counterfactual}:

\begin{equation*}
 \mathrm{NRMSE}_j=\frac{\sqrt{\frac1n\sum_i(\hat y_{ij}-y_{ij})^2}}{\sigma_j^{\mathrm{train}}}.
\end{equation*}

NRMSE is unitless, where lower error is better. For aggregation, we first compute metrics within each dataset--LLM--seed setting, averaging the relevant output dimensions and settings within that seed, then report the mean and sample standard deviations across three seeds. Outputs receive equal weight within a dataset, and aggregate results weight datasets and LLMs equally. That is, datasets with more dimensions or sample sizes do not receive more weight. 

\subsection{Hyperparameters and Model Selection}\label{sec:fitting-settings}

\paragraph{\method{}} We fit the reduced-rank broad predictor and shared linear focused predictor through alternating residual minimization, as described in $\S$\ref{sec:method-procedure}. Both ridge subproblems are solved analytically, with the broad predictor's rank and both regularization penalties selected on the validation split. \textit{We use a validation utility that averages NRMSE and one minus Pearson correlation.} \textbf{Table~\ref{tab:fitting-hyperparameters}} summarizes the hyperparameter settings.

\begin{table*}[!t]
\centering
\small
\begingroup
\renewcommand{\arraystretch}{1.05}
\setlength{\tabcolsep}{6pt}
\begin{tabular}{@{}
    p{0.27\textwidth}
    p{0.67\textwidth}
@{}}
\toprule
\textbf{Baselines} & \textbf{Values} \\
\midrule

\multicolumn{2}{@{}l}{\textbf{RELISH}} \\
Latent predictor
    & 1 latent token, width 256, 3 layers, 8 attention heads \\
Objective
    & Huber loss ($\delta=1$) \\
Optimization
    & Learning rates $\{10^{-5},3\times10^{-5},10^{-4}\}$, weight decay 0, batch size 8 \\
Configuration
    & Up to 30 epochs with patience 2. Minimizes the same validation objective as \method{}. \\
\addlinespace[4pt]

\multicolumn{2}{@{}l}{\textbf{GenRe2}} \\
Decoder
    & Width 256, 3 layers, 4 heads, FFN width 1024,
      dropout 0.1, adapter rank 16 \\
Objective
    & Numerical-token cross-entropy followed by ReMax refinement \\
Optimization
    & Learning rate $10^{-5}$, weight decay 0, batch size 8 \\
Configuration
    & Up to 20 epochs per stage with patience 2. Minimizes 
      validation loss (supervised) and maximizes reward (refinement) \\
\addlinespace[4pt]

\multicolumn{2}{@{}l}{\textbf{T-MES}} \\
Predictor
    & Hidden width 256, dropout 0.1 \\
Objective
    & We use weights 0.7, 0.3, and 0.1 for MSE, consistency, and diversity loss, respectively
       \\
Optimization
    & Learning rates $\{10^{-5},3\times10^{-5},10^{-4}\}$, batch size 16 \\
Configuration
    & Up to 100 epochs with patience 10. Minimizes the same validation objective as \method{}. \\
\addlinespace[4pt]

\multicolumn{2}{@{}l}{\textbf{GAT-AES}} \\
Graph predictor
    & Width 512, 2 attention layers (4 and 1 heads),
      dropout 0.1 \\
Objective
    & MSE \\
Optimization
    & Learning rates $\{10^{-5},3\times10^{-5},10^{-4}\}$, batch size 16 \\
Configuration
    & Up to 100 epochs with patience 10. Minimizes the same validation objective as \method{}. \\
\addlinespace[4pt]

\multicolumn{2}{@{}l}{\textbf{UniSO-N}} \\
Predictor
    & 2-layer ReLU MLP, width 2048 \\
Objective
    & MSE \\
Optimization
    & Learning rates $\{10^{-5},3\times10^{-5},10^{-4}\}$, batch size 128 \\
Configuration
    & Up to 200 epochs with patience 10. Minimizes the same validation objective as \method{}. \\
\addlinespace[4pt]

\multicolumn{2}{@{}l}{\textbf{MT-ICL}} \\
Transformer
    & Width 128, 6 layers, 4 heads, dropout 0.1 \\
Objective
    & Gaussian negative log-likelihood \\
Optimization
    & Learning rates
      $\{10^{-6},5\times10^{-6},10^{-5},
      5\times10^{-5},10^{-4},5\times10^{-4},10^{-3}\}$ \\
Configuration
    & Up to 1000 epochs with patience 10.
      Minimizes validation NLL \\
\addlinespace[4pt]

\multicolumn{2}{@{}l}{\textbf{SAJA}} \\
Predictor
    & Random forest with $\{100, 200, 400\}$ trees \\
Optimization
    & Bagging using 5-fold grouped CV \\
Configuration
    & Depth $\{5,10,15\}$,
      minimum split $\{5,10\}$,
      minimum leaf $\{2,4\}$, 
      maximum features $\{\texttt{sqrt}, 0.5\}$.
      Minimizes the same validation objective as \method{}. \\
\bottomrule
\end{tabular}
\endgroup
\caption{\textbf{Baseline Hyperparameters.}
Braced values denote searched alternatives. Whenever possible, we try to follow standard selection criteria except for methods that explicitly minimize some text-based generative loss.}
\label{tab:baseline-tuning}
\end{table*}

\paragraph{Baseline training and selection.}
\textbf{Table~\ref{tab:baseline-tuning}} summarizes the recorded configurations of the seven supervised baselines. The methods use different objectives and model-selection criteria, reflecting their respective prediction mechanisms. MT-ICL uses 32 training support examples per task, with up to 64 query examples per episode.

\begin{table}[t]
\centering
\footnotesize
\setlength{\tabcolsep}{4pt}
\renewcommand{\arraystretch}{1.08}
\begin{tabular}{@{}p{0.45\columnwidth}p{0.48\columnwidth}@{}}
\toprule
\textbf{Parameter} & \textbf{Values} \\
\midrule

\multicolumn{2}{@{}l}{\textit{Universal}} \\[-1pt]
Random seeds
    & $42, 1234, 2026$ \\
Validation Utility
    & (NRMSE + 1 - Pearson ) / 2 \\

\addlinespace[2pt]
\multicolumn{2}{@{}l}{\textit{Broad predictor}} \\[-1pt]
Ridge penalty $\gamma_{\mathrm b}$
    & $10^{-6}, 10^{-5}, \ldots, 10^{4}$ \\
Rank $k$
    & $\{1,2,4,8\}$, capped at $D-1$ \\

\addlinespace[2pt]
\multicolumn{2}{@{}l}{\textit{Focused predictor}} \\[-1pt]
Ridge penalty $\gamma_{\mathrm f}$
    & $10^{-6}, 10^{-5}, \ldots, 10^{4}$  \\

\bottomrule
\end{tabular}

\caption{\textbf{\method{} Hyperparameters.}
The broad rank $k$ constrains the output coefficient matrix to rank at most $k<D$. The ridge penalties $\gamma_{\mathrm b}$ and $\gamma_{\mathrm f}$ correspond to the unaveraged squared-error subproblems in $\S$\ref{sec:method-ridge} and $\S$\ref{sec:method-focused}, respectively.}
\label{tab:fitting-hyperparameters}
\end{table}

\subsection{Computational Budget}

We conduct experiments on NVIDIA Grace Hopper GH200 nodes, with a 72-core Arm-based Grace CPU and approximately 237GB of memory. All supervised experiments are completed within one thousand GPU hours. 

\section{Additional Key Results}\label{sec:additional-results}

The following analyses supplement the main results in the same order as the main text: predictive performance and output-level analyses, trainable footprint, training-data efficiency, and comparisons with Gemini 3.1 Pro.

\subsection{Predictive Performance}\label{sec:complete}

\textbf{Tables~\ref{tab:expanded-pearson}, \ref{tab:expanded-spearman}, and~\ref{tab:expanded-nrmse}} expand \textbf{Table~\ref{tab:overall}} with Pearson correlation, Spearman correlation, and NRMSE, respectively, for all seven supervised baselines across every dataset--LLM combination. Each dataset and LLM receives equal weight in the overall average.

\begin{figure*}[!t]
\centering
\includegraphics[width=\textwidth]{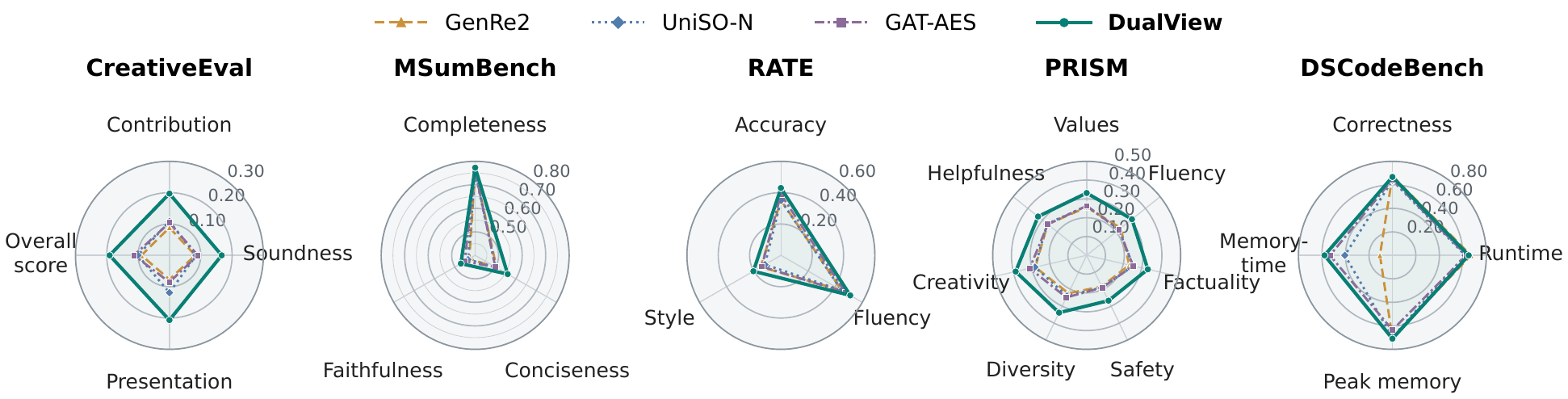}
\caption{\textbf{Output-level Spearman correlation.} Each spoke reports Spearman correlation for one dataset output, macro-averaged over 8 LLM backbones and 3 training seeds. We compare \method{} against the three strongest supervised baselines. Larger radial values are better, where scales vary across datasets for visibility.}
\label{fig:output-profiles-spearman}
\end{figure*}

\begin{figure*}[!t]
\centering
\includegraphics[width=\textwidth]{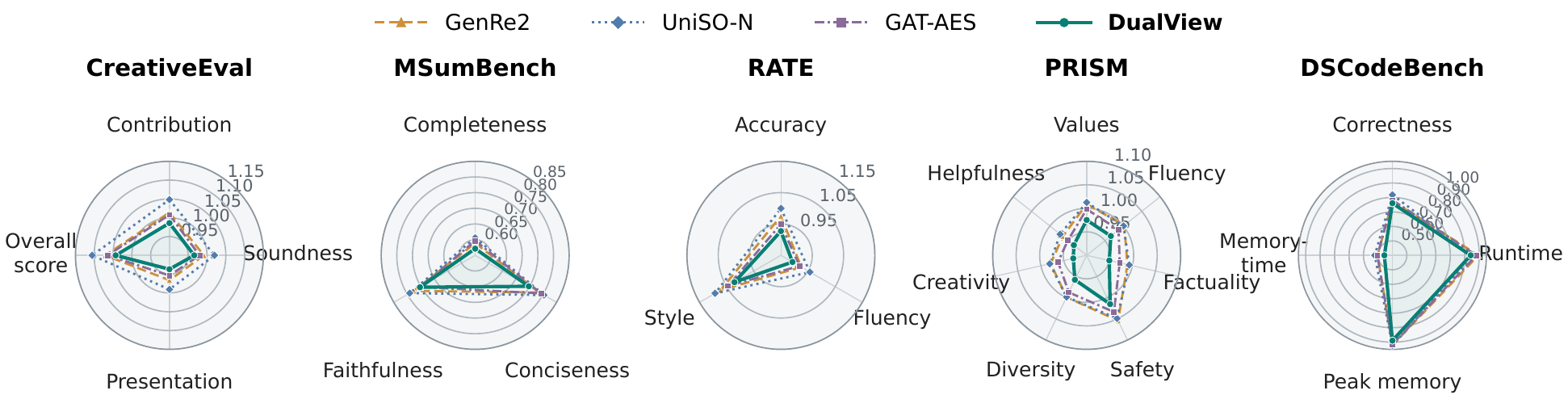}
\caption{\textbf{Output-level normalized RMSE (NRMSE).} Each spoke reports NRMSE for one dataset output, macro-averaged over 8 LLM backbones and 3 training seeds. We compare \method{} against the three strongest supervised baselines. Smaller radial values are better, where scales vary across datasets for visibility.}
\label{fig:output-profiles-nrmse}
\end{figure*}

\begin{table*}[p]
\centering\fontsize{7.3}{8.2}\selectfont\setlength{\tabcolsep}{2pt}\renewcommand{\arraystretch}{1.03}
\resizebox{\textwidth}{!}{\input{tables/dualview_pearson}}
\caption{\textbf{Pearson correlation across five datasets} (MSumBench, RATE, CreativeEval, DSCodeBench, and PRISM) and eight LLMs (Gemma 4 E4B, Qwen3.5 9B, Qwen3.6 27B, Gemma 4 31B, Llama 3.3 70B Instruct, Qwen2.5 72B Instruct, GPT OSS 120B, and Qwen3.5 122B A10B). Values are means with standard deviations across three seeds ($\pm$). Bold marks the best value in each row. \method{} wins are marked in light green whereas baseline wins are highlighted in yellow. Average LLM gives the within-dataset average, and Overall gives the average of all settings.}
\label{tab:expanded-pearson}
\end{table*}

\begin{table*}[p]\centering\scriptsize\setlength{\tabcolsep}{2pt}
\resizebox{\textwidth}{!}{\input{tables/dualview_spearman}}
\caption{\textbf{Spearman correlation across five datasets} (MSumBench, RATE, CreativeEval, DSCodeBench, and PRISM) and eight LLMs (Gemma 4 E4B, Qwen3.5 9B, Qwen3.6 27B, Gemma 4 31B, Llama 3.3 70B Instruct, Qwen2.5 72B Instruct, GPT OSS 120B, and Qwen3.5 122B A10B). Values are means with standard deviations across three seeds ($\pm$). Bold marks the best value in each row. \method{} wins are marked in light green whereas baseline wins are highlighted in yellow. Average LLM gives the within-dataset average, and Overall gives the average of all settings.}
\label{tab:expanded-spearman}
\end{table*}

\begin{table*}[p]\centering\scriptsize\setlength{\tabcolsep}{2pt}
\resizebox{\textwidth}{!}{\input{tables/dualview_relative_rmse}}
\caption{\textbf{Normalized RMSE (NRMSE) across five datasets} (MSumBench, RATE, CreativeEval, DSCodeBench, and PRISM) and eight LLMs (Gemma 4 E4B, Qwen3.5 9B, Qwen3.6 27B, Gemma 4 31B, Llama 3.3 70B Instruct, Qwen2.5 72B Instruct, GPT OSS 120B, and Qwen3.5 122B A10B). Values are means with standard deviations across three seeds ($\pm$). Bold marks the best value in each row. \method{} wins are marked in light green whereas baseline wins are highlighted in yellow. Average LLM gives the within-dataset average, and Overall gives the average of all settings.}
\label{tab:expanded-nrmse}
\end{table*}

\subsubsection{Fine-grained Performance Analyses}\label{sec:dimension-robustness}

To complement Table~\ref{tab:overall}, we examine how \method{} performs across different output targets and whether gains persist under aggregation schemes that penalize weak targets.

\paragraph{Output-level analyses.}
Similar to \textbf{Figure~\ref{fig:output-profiles}}, \textbf{Figures~\ref{fig:output-profiles-spearman} and~\ref{fig:output-profiles-nrmse}} report Spearman correlation and NRMSE results for \method{} across five datasets spanning 21 output targets. \method{} still wins most targets, leading Spearman correlation on 19 of 21 outputs and NRMSE on 20 of 21 compared to the three strongest supervised baselines. The two Spearman deficits are DSCodeBench correctness and runtime, where GenRe2 outperforms \method{} by 0.008 and 0.018, respectively. For NRMSE, the sole exception is MSumBench faithfulness, where \method{} is effectively tied with GAT-AES (< 0.01 difference).

\paragraph{Strict wins and worst-output gaps.}\label{sec:strict_wins}

These measures assess consistency across individual LLM--target comparisons since averaging across LLMs can still mask weak performance. A \textit{strict win} occurs when a method outperforms every other method on a given output, dataset, LLM, metric, and seed (excluding ties). We compute the winning percentage within each dataset--LLM setting and macro-average these percentages across settings.

To measure loss severity, we define each method's \textit{output-level gap} relative to the best-performing method on the same output. This gap is the nonnegative difference in Pearson or Spearman correlation, or the excess NRMSE. For each dataset--LLM setting and seed, the \textit{worst-output gap} is the largest such gap across targets. We then macro-average the worst-output gaps across dataset--LLM settings and report mean $\pm$ sample standard deviation across three seeds.

\textbf{Table~\ref{tab:output-spread-main}} shows that \method{} has higher strict win rates and smaller mean worst-output gaps than the three strongest supervised baselines. For Pearson and NRMSE, its win rates are 65.9\% and 71.4\%, respectively, versus 8.9\% and 7.7\% for GAT-AES. Its corresponding mean worst-output gaps are 0.039 and 0.016, versus 0.149 and 0.055 for GAT-AES. These measures indicate that \method{} wins more frequently and incurs smaller worst-target deficits when it does not rank first. Win rates do not add up to 100\% because we compared to all seven supervised baselines but only show the three strongest. 

\begin{table}
\centering\small\setlength{\tabcolsep}{2.5pt}
\input{tables/dualview_output_spread_main}
\caption{\textbf{Output-level wins and gaps.} Top: percentage of LLM--target comparisons on which a method strictly outperforms all competitors ($\uparrow$). Win rates do not sum to 100\% because we only show values for the three strongest supervised baselines, while win rates are computed by comparing \method{} to all seven supervised baselines. Bottom: largest target-level gap relative to the best method ($\downarrow$). Results are macro-averaged across 5 datasets and 8 LLMs within each seed, with standard deviations ($\pm$) over three seeds.}
\label{tab:output-spread-main}
\end{table}

\paragraph{Power-mean aggregation.}

For a given dataset, LLM, seed, and evaluation metric, let $a_j$ denote the score on output $j$. We aggregate across its $D$ outputs using the power mean
\begin{equation*}
M_q(\mathbf a)=\left(\frac{1}{D}\sum_{j=1}^{D}a_j^q\right)^{1/q}.
\end{equation*}
The main results use the arithmetic mean ($q=1$). To emphasize weak outputs, we instead use the harmonic mean ($q=-1$) for positive correlation scores and the root mean square ($q=2$) for NRMSE. When any correlation is zero, its negative-order power mean is defined as zero. We observe no negative correlations empirically. Unlike the arithmetic mean, the harmonic mean penalizes small correlations more strongly. Conversely, RMS assigns greater weight to large prediction errors.

Power means are computed within each dataset--LLM--seed setting before macro-averaging datasets and LLMs within each seed. We report means and sample standard deviations across three seeds. \textbf{Table~\ref{tab:power-mean-full}} shows that \method{} outperforms other supervised baselines even when weaker outputs are penalized more heavily.

\begin{table*}[t]\centering\small\setlength{\tabcolsep}{3pt}
\resizebox{\textwidth}{!}{\input{tables/dualview_power_mean_full}}
\caption{\textbf{Predictive performance under weak-output penalties.} We report harmonic means ($q=-1$, $\uparrow$) for Pearson and Spearman and root mean square for NRMSE ($q=2$, $\downarrow$). Average Dataset weights the five datasets equally. Entries report mean $\pm$ sample standard deviation across three seeds.}\label{tab:power-mean-full}
\end{table*}

\paragraph{Sensitivity to stronger penalties on weak correlations.}
We further examine power means using $q=-2$, which place greater emphasis on outputs with low positive correlations. Since $M_{-2}\leq M_{-1}\leq M_1$, this provides a more stringent check than harmonic aggregation. \textbf{Table~\ref{tab:power-mean-sensitivity}} shows that \method{} remains the strongest method even when weak Pearson and Spearman performances are penalized more heavily.

\begin{table*}\centering\small\setlength{\tabcolsep}{3pt}
\resizebox{\textwidth}{!}{\input{tables/dualview_power_mean_sensitivity}}
\caption{\textbf{Sensitivity to a stronger weak-output correlation penalty.} We report power means using order $q = -2$ to penalize small correlations more strongly than the harmonic means ($q = -1$) in \textbf{Table~\ref{tab:power-mean-full}}. Values are macro-averaged across eight LLMs with sample standard deviation across three seeds ($\pm$).}\label{tab:power-mean-sensitivity}
\end{table*}

\subsection{Trainable Footprint}\label{sec:cost}

\textbf{Table~\ref{tab:parameters-absolute}} supplements \textbf{Table~\ref{tab:parameters}} with fitted parameter counts for the supervised baselines across all eight backbones. Ranges reflect variation across datasets and seeds. SAJA is omitted because it uses random forests rather than neural regression modules.

For \method{}, the reduced-rank broad predictor has $k(H+D)$ coefficients, the shared focused predictor has $H$, and the target-specific intercepts have $D$, yielding $(k+1)(H+D)$ fitted parameters total. Thus, the trainable footprint depends on the model hidden size $H$, number of targets $D$, and selected rank $k$, but not directly on the size of the underlying LLM backbone.

\begin{table*}[tbp]\centering\scriptsize\setlength{\tabcolsep}{3pt}
\resizebox{\textwidth}{!}{\input{tables/dualview_parameters}}
\caption{\textbf{Absolute fitted head coefficients for all neural comparators.} Baseline counts are in millions (M), whereas \method{} counts are in thousands (K). Ranges span different configurations across datasets and seeds, if any. SAJA fits random forests and therefore has no neural coefficient count.}
\label{tab:parameters-absolute}\end{table*}

\subsection{Training-Data Efficiency}\label{sec:data-efficiency}

\begin{figure*}[p]\centering
\includegraphics[width=\textwidth]{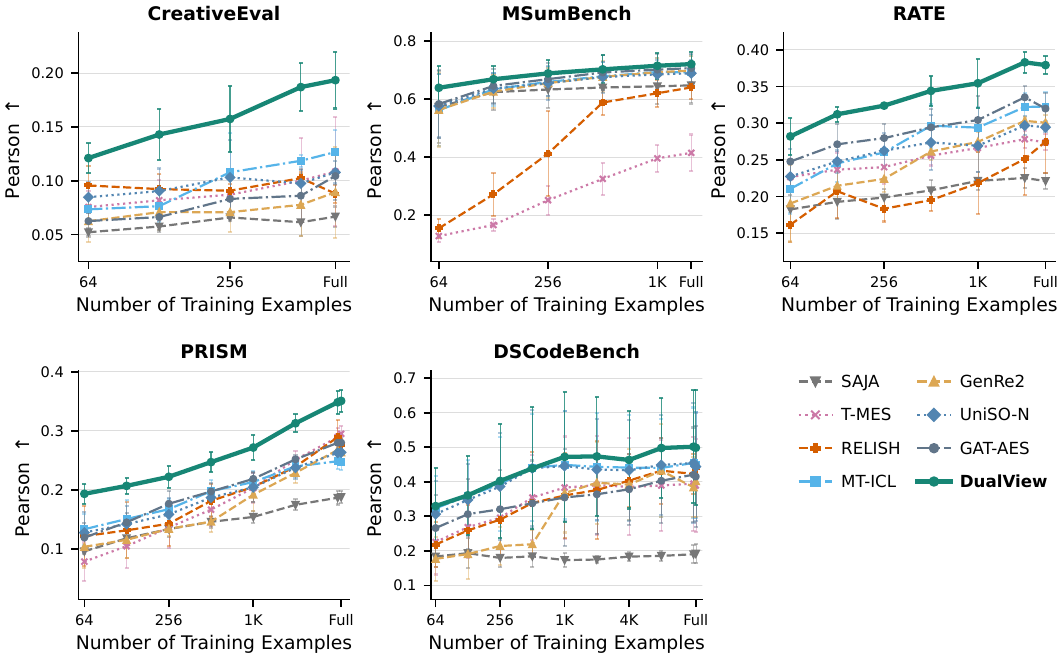}
\caption{\textbf{Training-data efficiency: Pearson.} Each curve averages outputs across eight LLMs within each seed, then shows the mean and sample standard deviation across three seeds. Full denotes the complete training split.}\label{fig:efficiency-pearson-all}
\end{figure*}

\begin{figure*}[p]\centering
\includegraphics[width=\textwidth]{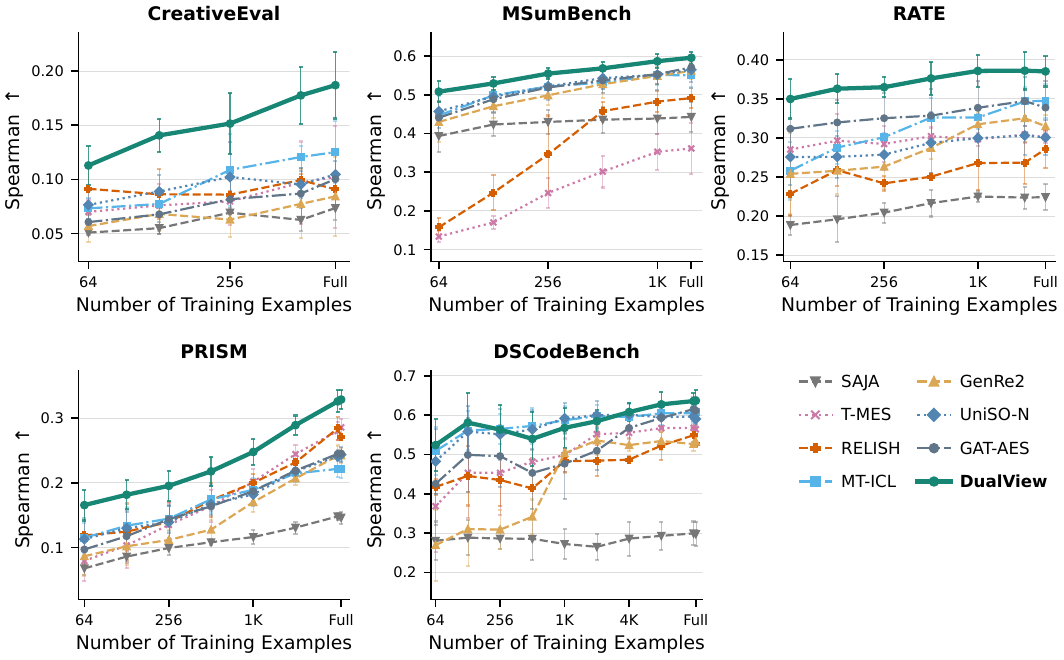}
\caption{\textbf{Training-data efficiency: Spearman.} Each curve averages outputs across eight LLMs within each seed, then shows the mean and sample standard deviation across three seeds. Full denotes the complete training split.}
\label{fig:efficiency-spearman-test}\end{figure*}

\begin{figure*}[p]\centering
\includegraphics[width=\textwidth]{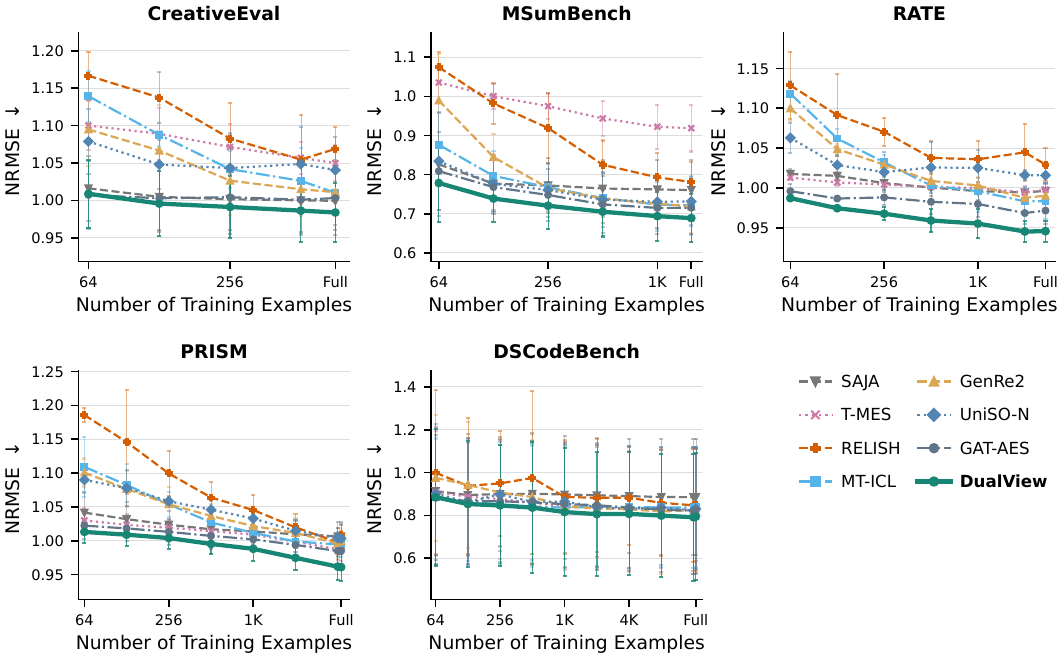}
\caption{\textbf{Training-data efficiency: NRMSE.} Each curve averages outputs across eight LLMs within each seed, then shows the mean and sample standard deviation across three seeds. Full denotes the complete training split.}
\label{fig:efficiency-nrmse-test}\end{figure*}

\textbf{Figures~\ref{fig:efficiency-pearson-all}, \ref{fig:efficiency-spearman-test}, and~\ref{fig:efficiency-nrmse-test}} extend \textbf{Figure~\ref{fig:training-efficiency}} to all seven supervised baselines, reporting Pearson correlation, Spearman correlation, and NRMSE as functions of training-set size. We vary the training budget but do not change the validation and test splits. Each dot represents macro-averages across eight LLMs within each seed, with whiskers showing standard deviation across three seeds.

\subsection{Comparison with Gemini 3.1 Pro}

\textbf{Table~\ref{tab:gemini-comparison-full}} extends \textbf{Table~\ref{tab:gemini-comparison}} by reporting Spearman correlation in addition to NRMSE and Pearson correlation. We compare zero- and four-shot Gemini 3.1 Pro with \method{} using Gemma 4 31B on each dataset. We selected this model for \method{} based on validation results, since comparing Gemini against results aggregated across multiple \method{} backbones would not be an apples-to-apples comparison.

\begin{table*}[!t]
\centering\small\setlength{\tabcolsep}{3pt}
\resizebox{\textwidth}{!}
{\input{tables/dualview_gemini_by_dataset}}
\caption{\textbf{Comparisons to Gemini 3.1 Pro.} We compare \method{} to Gemini 3.1 Pro under zero-shot and few-shot (4 demonstrations) settings across five datasets. \method{} results use Gemma 4 31B. Metrics are macro-averaged across dimensions within each dataset with standard deviations across three seeds ($\pm$).}
\label{tab:gemini-comparison-full}
\end{table*}

\section{Additional Ablations}\label{sec:additional-ablations}

\begin{table}[!t]
\centering
\small
\begingroup
\renewcommand{\arraystretch}{0.95}
\setlength{\tabcolsep}{3pt}
\resizebox{\columnwidth}{!}{%
\begin{tabular}{@{}lcc@{}}
\toprule
\textbf{Predictor}
& \textbf{NRMSE} $\bm{\downarrow}$
& $\bm{r\uparrow}$ \\
\midrule
\multicolumn{3}{@{}l}{\textit{Strongest adapted baseline}} \\
GAT-AES
& \result{0.898}{0.070} & \result{0.365}{0.032} \\
\addlinespace[2pt]
\multicolumn{3}{@{}l}{\textit{Single-view ridge}} \\
Broad (full rank)
& \result{0.892}{0.070} & \result{0.388}{0.026} \\
Focused (shared)
& \result{0.912}{0.068} & \result{0.377}{0.033} \\
Focused (independent)
& \result{0.893}{0.070} & \result{0.379}{0.028} \\
\addlinespace[2pt]
\multicolumn{3}{@{}l}{\textit{Joint two-view predictors}} \\
Full rank + independent
& \result{0.874}{0.072} & \result{0.429}{0.038} \\
Full rank + shared
& \best{0.873}{0.070} & \best{0.433}{0.031} \\
\textbf{\method{} (ours)}
& \best{0.873}{0.073} & \result{0.432}{0.038} \\
\bottomrule
\end{tabular}%
}
\endgroup
\caption{\textbf{Linear readout comparisons.}
Single-view controls use either broad or focused representations. Joint two-view configurations vary the broad rank constraint and focused parameter sharing. Values are macro-averaged across five datasets and eight LLMs, with standard deviations across three seeds. We include GAT-AES as the strongest adapted supervised baseline.}
\label{tab:linear-readout-summary}
\end{table}

\subsection{Strong Linear Baselines}
\label{sec:linear-baseline-results}

To assess how much of \method{}'s advantage comes from the extracted representations and regularized linear prediction, we compare against single-view ridge baselines and less constrained two-view predictors. \textbf{Table~\ref{tab:linear-readout-summary}} consolidates the corresponding results from the evidence-complementarity and parameter-sharing ablations.

\paragraph{Strength of single-view linear predictors.}
Full-rank ridge regression on broad representations achieves a Pearson correlation of 0.388 and NRMSE of 0.892, outperforming all seven adapted supervised baselines in aggregate. Focused-only ridge predictors are also competitive: the shared version achieves a Pearson correlation of 0.377, while independent predictors achieve 0.379 and an NRMSE of 0.893. These results establish that regularized linear readouts of frozen LLM representations are strong baselines by themselves.

\paragraph{Value of combining evidence.}
Nevertheless, \method{} outperforms full-rank broad ridge, with a 11.3\% higher Pearson and a 2.1\% lower NRMSE. Although a simple ridge predictor on broad representations already outperforms the strongest supervised baseline (GAT-AES), combining broad and target-specific representations yields further gains. This finding supports the complementary predictive value of the two evidence views.

\paragraph{Role of parameter sharing.}
Less constrained two-view predictors perform similarly to \method{}. Consequently, the parameter-sharing constraints should primarily be interpreted as an efficiency mechanism: they substantially reduce the trainable footprint while preserving comparable observed predictive performance. Nevertheless, these comparisons are not measurements of total computational savings.

\subsection{Statistical Reliability of Ablation Improvements}
\label{sec:bootstrap_ablations}

We assess the statistical reliability of two key ablation
findings: a) the benefit of combining broad and
output-specific evidence compared to full rank broad-only ridge (Table~\ref{tab:evidence-ablation}), and b) the benefit of alternating joint optimization over separate fitting with $\alpha$ fusion (Table~\ref{tab:optimization-ablation}).
Specifically, we conduct a paired, source-group-level bootstrap analysis on test predictions. We resample source groups with replacement within each dataset, using identical resamples for each pair of compared methods to preserve within-group dependencies. For each of 5,000 bootstrap replicates, we recompute Pearson correlation and NRMSE and aggregate across all outputs, LLM backbones, datasets, and seeds following the protocol in \S\ref{sec:metric-definition}.

We report 95\% confidence intervals (CIs) for paired performance differences, computed as \method{} minus the baseline for Pearson correlation and the reverse for NRMSE, so that positive values indicate improvements in both metrics. Table~\ref{tab:bootstrap_ablations} shows that \method{}'s gains are likely statistically significant, since all intervals exclude zero. 

\begin{table}[t]
    \centering
    \footnotesize
    \setlength{\tabcolsep}{3pt}
    \begin{tabular}{@{}lcc@{}}
        \toprule
        Comparator & $\Delta$Pearson & $\Delta$NRMSE \\
        \midrule
        Broad-only ridge
        & $[0.0328, 0.0497]$
        & $[0.0134, 0.0229]$ \\
        Separate + fusion
        & $[0.0053, 0.0177]$
        & $[0.0039, 0.0109]$ \\
        \bottomrule
    \end{tabular}
    \caption{\textbf{95\% Confidence Intervals} for improvements of \method{} compared to ablation alternatives. Positive values indicate that differences are likely statistically significant.}
    \label{tab:bootstrap_ablations}
\end{table}

Compared with full-rank broad-only ridge, \method{} improves aggregate Pearson correlation from 0.3876 to 0.4322 (+0.0447) and reduces NRMSE from 0.8919 to 0.8732 (+0.0186). The corresponding 95\% CIs are
$[0.0328, 0.0497]$ and $[0.0134, 0.0229]$, respectively, supporting the complementary predictive value of broad and output-specific representations. 

Compared with separate fitting followed by $\alpha$ fusion, \method{} further improves Pearson correlation from 0.419 to 0.432 (+0.013) and reduces NRMSE from 0.880 to 0.873 (a reduction of 0.007). The corresponding 95\% CIs are $[0.0053, 0.0177]$ and $[0.0039, 0.0109]$, respectively, further supporting that alternating joint optimization outperforms separate fitting and fusion.

\section{Expanded Discussion}\label{sec:discussion-extended}

\subsection{Multi-Objective Optimization}\label{sec:rw-moo}
Multi-objective formulations typically apply to balance different losses under multi-task learning \citep{sener2018multiobjective}. PCGrad projects conflicting gradients, CAGrad regularizes updates through worst-task local improvement, Nash-MTL negotiates a joint update, and FAMO adaptively balances loss reduction \citep{yu2020pcgrad,liu2021cagrad,navon2022nash,liu2023famo}. FairGrad controls fairness of task progress, while ExcessMTL addresses excessive weighting of noisy tasks \citep{ban2024fairgrad,he2024excessmtl}. Recent work also studies aligned objectives, selective task-group updates, and task-affinity prediction \citep{efroni2025aligned,jeong2025selective,ayman2026etap}.

Some MOO works cater to text regression. SaMRL trains an autoregressive multi-trait essay scorer with QWK-based rewards and an MSE penalty, learning weights for multiple reward-specific PPO losses \citep{do-etal-2024-autoregressive-multi}. LMUnit combines preference, direct-rating, and rationale supervision for output-conditioned evaluation \citep{saad-falcon-etal-2025-lmunit}. Multi-objective reward models fit attribute scores \citep{wang-etal-2024-interpretable}, and recent work combines Bradley--Terry preference and multi-attribute regression objectives in a shared embedding space \citep{zhang2026complementary}. Thus, these MOO methods are compatible with \method{}'s evidence design by balancing optimization across outputs, whereas our two-view design specifies the evidence available to each prediction.



\input{sections/dataset_prompts}

\section{Output-Level Performance Tables}\label{sec:targets}

Tables~\ref{tab:targets-creativeeval-gemma-e4b}--\ref{tab:targets-creativeeval-qwen-122b},
\ref{tab:targets-msumbench-gemma-e4b}--\ref{tab:targets-msumbench-qwen-122b},
\ref{tab:targets-rate-gemma-e4b}--\ref{tab:targets-rate-qwen-122b},
\ref{tab:targets-prism-gemma-e4b}--\ref{tab:targets-prism-qwen-122b}, and
\ref{tab:targets-dscodebench-gemma-e4b}--\ref{tab:targets-dscodebench-qwen-122b}
report output-level NRMSE, Pearson, and Spearman for CreativeEval, MSumBench, RATE, PRISM, and DSCodeBench, respectively, across all eight LLM backbones. Values are averaged across three training seeds with sample standard deviation ($\pm$).

\input{tables/dualview_targets}

\end{document}

%% file: tables/dualview_pearson.tex
\begin{tabular}{llcccccccc}
\toprule
\textbf{Dataset} & \textbf{LLM} & \textbf{SAJA} & \textbf{T-MES} & \textbf{RELISH} & \textbf{MT-ICL} & \textbf{GenRe2} & \textbf{UniSO-N} & \textbf{GAT-AES} & \textbf{\method{} (Ours)} \\
\midrule
CreativeEval & Gemma E4B & \result{0.044}{0.015} & \result{0.080}{0.046} & \result{0.073}{0.020} & \result{0.088}{0.024} & \result{0.056}{0.012} & \result{0.089}{0.042} & \result{0.060}{0.022} & \cellcolor{dvgreen}\best{0.154}{0.049} \\
 & Qwen 9B & \result{0.041}{0.012} & \result{0.084}{0.048} & \result{0.098}{0.017} & \result{0.087}{0.040} & \result{0.101}{0.061} & \result{0.087}{0.026} & \result{0.122}{0.057} & \cellcolor{dvgreen}\best{0.200}{0.044} \\
 & Qwen 27B & \result{0.075}{0.004} & \result{0.129}{0.053} & \result{0.083}{0.041} & \result{0.139}{0.061} & \result{0.082}{0.042} & \result{0.069}{0.013} & \result{0.152}{0.080} & \cellcolor{dvgreen}\best{0.217}{0.035} \\
 & Gemma 31B & \result{0.077}{0.028} & \result{0.110}{0.057} & \result{0.126}{0.036} & \result{0.141}{0.035} & \result{0.121}{0.040} & \result{0.124}{0.032} & \result{0.128}{0.018} & \cellcolor{dvgreen}\best{0.207}{0.030} \\
 & Llama 70B & \result{0.085}{0.022} & \result{0.104}{0.020} & \result{0.097}{0.048} & \result{0.131}{0.027} & \result{0.095}{0.040} & \result{0.136}{0.014} & \result{0.066}{0.011} & \cellcolor{dvgreen}\best{0.204}{0.012} \\
 & Qwen 72B & \result{0.057}{0.027} & \result{0.126}{0.027} & \result{0.075}{0.045} & \result{0.119}{0.046} & \result{0.112}{0.066} & \result{0.107}{0.048} & \result{0.107}{0.037} & \cellcolor{dvgreen}\best{0.203}{0.041} \\
 & GPT 120B & \result{0.080}{0.020} & \result{0.081}{0.032} & \result{0.077}{0.027} & \result{0.082}{0.025} & \result{0.078}{0.047} & \result{0.101}{0.014} & \result{0.054}{0.007} & \cellcolor{dvgreen}\best{0.153}{0.026} \\
 & Qwen 122B & \result{0.060}{0.023} & \result{0.116}{0.064} & \result{0.095}{0.011} & \result{0.158}{0.053} & \result{0.093}{0.055} & \result{0.139}{0.019} & \result{0.085}{0.038} & \cellcolor{dvgreen}\best{0.223}{0.031} \\
\rowcolor{dvlight}  & Average LLM & \result{0.065}{0.006} & \result{0.104}{0.041} & \result{0.090}{0.010} & \result{0.118}{0.007} & \result{0.092}{0.036} & \result{0.107}{0.007} & \result{0.097}{0.025} & \cellcolor{dvgreen}\best{0.195}{0.028} \\
\midrule
MSumBench & Gemma E4B & \result{0.651}{0.058} & \result{0.378}{0.086} & \result{0.651}{0.049} & \result{0.664}{0.033} & \result{0.697}{0.046} & \result{0.697}{0.029} & \result{0.704}{0.050} & \cellcolor{dvgreen}\best{0.717}{0.038} \\
 & Qwen 9B & \result{0.638}{0.075} & \result{0.415}{0.084} & \result{0.636}{0.037} & \result{0.674}{0.046} & \result{0.703}{0.045} & \result{0.678}{0.061} & \result{0.706}{0.046} & \cellcolor{dvgreen}\best{0.717}{0.043} \\
 & Qwen 27B & \result{0.643}{0.054} & \result{0.409}{0.066} & \result{0.674}{0.040} & \result{0.706}{0.045} & \result{0.710}{0.043} & \result{0.694}{0.054} & \result{0.727}{0.039} & \cellcolor{dvgreen}\best{0.735}{0.040} \\
 & Gemma 31B & \result{0.639}{0.056} & \result{0.508}{0.022} & \result{0.557}{0.046} & \result{0.699}{0.040} & \result{0.700}{0.048} & \result{0.696}{0.047} & \result{0.709}{0.050} & \cellcolor{dvgreen}\best{0.729}{0.040} \\
 & Llama 70B & \result{0.647}{0.072} & \result{0.362}{0.072} & \result{0.630}{0.029} & \result{0.684}{0.056} & \result{0.699}{0.054} & \result{0.685}{0.047} & \result{0.696}{0.053} & \cellcolor{dvgreen}\best{0.714}{0.054} \\
 & Qwen 72B & \result{0.654}{0.080} & \result{0.393}{0.056} & \result{0.633}{0.099} & \result{0.723}{0.040} & \result{0.708}{0.043} & \result{0.710}{0.032} & \result{0.717}{0.053} & \cellcolor{dvgreen}\best{0.733}{0.031} \\
 & GPT 120B & \result{0.661}{0.053} & \result{0.440}{0.031} & \result{0.656}{0.024} & \result{0.647}{0.047} & \result{0.676}{0.055} & \result{0.636}{0.071} & \result{0.676}{0.052} & \cellcolor{dvgreen}\best{0.692}{0.049} \\
 & Qwen 122B & \result{0.647}{0.064} & \result{0.417}{0.075} & \result{0.684}{0.036} & \result{0.694}{0.050} & \result{0.706}{0.051} & \result{0.709}{0.045} & \result{0.715}{0.055} & \cellcolor{dvgreen}\best{0.730}{0.037} \\
\rowcolor{dvlight}  & Average LLM & \result{0.647}{0.063} & \result{0.415}{0.059} & \result{0.640}{0.043} & \result{0.686}{0.044} & \result{0.700}{0.048} & \result{0.688}{0.047} & \result{0.706}{0.050} & \cellcolor{dvgreen}\best{0.721}{0.041} \\
\midrule
RATE & Gemma E4B & \result{0.158}{0.023} & \result{0.245}{0.029} & \result{0.241}{0.035} & \result{0.297}{0.008} & \result{0.290}{0.007} & \result{0.280}{0.045} & \result{0.330}{0.017} & \cellcolor{dvgreen}\best{0.350}{0.014} \\
 & Qwen 9B & \result{0.170}{0.016} & \result{0.289}{0.020} & \result{0.312}{0.067} & \result{0.225}{0.021} & \result{0.259}{0.008} & \result{0.295}{0.042} & \result{0.309}{0.022} & \cellcolor{dvgreen}\best{0.354}{0.019} \\
 & Qwen 27B & \result{0.298}{0.050} & \result{0.279}{0.008} & \result{0.299}{0.123} & \result{0.337}{0.029} & \result{0.304}{0.022} & \result{0.336}{0.038} & \result{0.335}{0.009} & \cellcolor{dvgreen}\best{0.408}{0.018} \\
 & Gemma 31B & \result{0.352}{0.033} & \result{0.333}{0.054} & \result{0.307}{0.027} & \result{0.391}{0.021} & \result{0.383}{0.039} & \result{0.342}{0.061} & \result{0.424}{0.051} & \cellcolor{dvgreen}\best{0.453}{0.020} \\
 & Llama 70B & \result{0.150}{0.015} & \result{0.278}{0.013} & \result{0.244}{0.043} & \result{0.272}{0.038} & \result{0.305}{0.013} & \result{0.262}{0.051} & \result{0.286}{0.023} & \cellcolor{dvgreen}\best{0.344}{0.021} \\
 & Qwen 72B & \result{0.210}{0.021} & \result{0.272}{0.039} & \result{0.252}{0.025} & \result{0.282}{0.047} & \result{0.291}{0.020} & \result{0.295}{0.015} & \result{0.341}{0.042} & \cellcolor{dvgreen}\best{0.370}{0.031} \\
 & GPT 120B & \result{0.179}{0.031} & \result{0.245}{0.031} & \result{0.271}{0.050} & \result{0.240}{0.006} & \result{0.253}{0.006} & \result{0.208}{0.040} & \result{0.241}{0.010} & \cellcolor{dvgreen}\best{0.324}{0.020} \\
 & Qwen 122B & \result{0.270}{0.012} & \result{0.291}{0.010} & \result{0.270}{0.033} & \result{0.327}{0.030} & \result{0.334}{0.041} & \result{0.352}{0.020} & \result{0.353}{0.013} & \cellcolor{dvgreen}\best{0.427}{0.014} \\
\rowcolor{dvlight}  & Average LLM & \result{0.223}{0.012} & \result{0.279}{0.014} & \result{0.274}{0.043} & \result{0.296}{0.016} & \result{0.302}{0.006} & \result{0.296}{0.012} & \result{0.327}{0.020} & \cellcolor{dvgreen}\best{0.379}{0.012} \\
\midrule
PRISM & Gemma E4B & \result{0.208}{0.015} & \result{0.292}{0.011} & \result{0.274}{0.016} & \result{0.257}{0.018} & \result{0.261}{0.014} & \result{0.243}{0.015} & \result{0.276}{0.006} & \cellcolor{dvgreen}\best{0.343}{0.021} \\
 & Qwen 9B & \result{0.179}{0.012} & \result{0.290}{0.017} & \result{0.282}{0.020} & \result{0.268}{0.019} & \result{0.267}{0.012} & \result{0.307}{0.024} & \result{0.294}{0.012} & \cellcolor{dvgreen}\best{0.352}{0.008} \\
 & Qwen 27B & \result{0.188}{0.023} & \result{0.305}{0.002} & \result{0.296}{0.005} & \result{0.260}{0.019} & \result{0.256}{0.023} & \result{0.278}{0.018} & \result{0.306}{0.011} & \cellcolor{dvgreen}\best{0.352}{0.020} \\
 & Gemma 31B & \result{0.175}{0.021} & \result{0.287}{0.027} & \result{0.262}{0.034} & \result{0.252}{0.006} & \result{0.273}{0.017} & \result{0.259}{0.007} & \result{0.306}{0.011} & \cellcolor{dvgreen}\best{0.340}{0.022} \\
 & Llama 70B & \result{0.183}{0.009} & \result{0.326}{0.019} & \result{0.269}{0.061} & \result{0.270}{0.017} & \result{0.287}{0.037} & \result{0.287}{0.024} & \result{0.309}{0.013} & \cellcolor{dvgreen}\best{0.376}{0.022} \\
 & Qwen 72B & \result{0.220}{0.015} & \result{0.305}{0.015} & \result{0.244}{0.050} & \result{0.280}{0.018} & \result{0.272}{0.033} & \result{0.268}{0.041} & \result{0.268}{0.016} & \cellcolor{dvgreen}\best{0.356}{0.024} \\
 & GPT 120B & \result{0.168}{0.005} & \result{0.261}{0.010} & \result{0.307}{0.019} & \result{0.235}{0.007} & \result{0.228}{0.038} & \result{0.216}{0.010} & \result{0.229}{0.027} & \cellcolor{dvgreen}\best{0.353}{0.018} \\
 & Qwen 122B & \result{0.182}{0.021} & \result{0.305}{0.016} & \result{0.326}{0.021} & \result{0.268}{0.010} & \result{0.232}{0.013} & \result{0.260}{0.032} & \result{0.273}{0.009} & \cellcolor{dvgreen}\best{0.341}{0.022} \\
\rowcolor{dvlight}  & Average LLM & \result{0.188}{0.010} & \result{0.296}{0.011} & \result{0.283}{0.023} & \result{0.261}{0.009} & \result{0.260}{0.017} & \result{0.265}{0.017} & \result{0.283}{0.011} & \cellcolor{dvgreen}\best{0.352}{0.019} \\
\midrule
DSCodeBench & Gemma E4B & \result{0.190}{0.032} & \result{0.390}{0.131} & \result{0.474}{0.078} & \result{0.403}{0.136} & \result{0.361}{0.062} & \result{0.423}{0.175} & \result{0.399}{0.135} & \cellcolor{dvgreen}\best{0.475}{0.198} \\
 & Qwen 9B & \result{0.173}{0.020} & \result{0.392}{0.140} & \result{0.392}{0.127} & \result{0.323}{0.099} & \result{0.445}{0.165} & \result{0.331}{0.080} & \result{0.440}{0.171} & \cellcolor{dvgreen}\best{0.497}{0.129} \\
 & Qwen 27B & \result{0.206}{0.044} & \result{0.405}{0.149} & \result{0.396}{0.111} & \result{0.428}{0.112} & \result{0.415}{0.179} & \result{0.469}{0.198} & \result{0.480}{0.219} & \cellcolor{dvgreen}\best{0.532}{0.206} \\
 & Gemma 31B & \result{0.232}{0.031} & \result{0.402}{0.135} & \result{0.373}{0.092} & \result{0.490}{0.142} & \result{0.461}{0.163} & \cellcolor{dvyellow}\best{0.549}{0.230} & \result{0.430}{0.146} & \result{0.548}{0.206} \\
 & Llama 70B & \result{0.192}{0.043} & \result{0.393}{0.136} & \result{0.383}{0.047} & \result{0.362}{0.090} & \result{0.401}{0.113} & \result{0.425}{0.120} & \result{0.410}{0.139} & \cellcolor{dvgreen}\best{0.503}{0.197} \\
 & Qwen 72B & \result{0.171}{0.015} & \result{0.403}{0.134} & \result{0.391}{0.138} & \result{0.359}{0.095} & \result{0.354}{0.093} & \result{0.484}{0.218} & \result{0.370}{0.112} & \cellcolor{dvgreen}\best{0.544}{0.183} \\
 & GPT 120B & \result{0.182}{0.009} & \result{0.337}{0.121} & \result{0.413}{0.131} & \result{0.341}{0.077} & \result{0.399}{0.093} & \result{0.411}{0.134} & \result{0.385}{0.124} & \cellcolor{dvgreen}\best{0.473}{0.137} \\
 & Qwen 122B & \result{0.186}{0.024} & \result{0.397}{0.141} & \result{0.460}{0.021} & \result{0.335}{0.083} & \result{0.368}{0.076} & \result{0.460}{0.191} & \result{0.381}{0.090} & \cellcolor{dvgreen}\best{0.530}{0.168} \\
\rowcolor{dvlight}  & Average LLM & \result{0.192}{0.026} & \result{0.390}{0.131} & \result{0.410}{0.082} & \result{0.380}{0.096} & \result{0.401}{0.109} & \result{0.444}{0.157} & \result{0.412}{0.137} & \cellcolor{dvgreen}\best{0.513}{0.171} \\
\midrule
Overall &  & \result{0.263}{0.014} & \result{0.297}{0.033} & \result{0.340}{0.034} & \result{0.348}{0.025} & \result{0.351}{0.024} & \result{0.360}{0.042} & \result{0.365}{0.032} & \cellcolor{dvgreen}\best{0.432}{0.038} \\
\bottomrule
\end{tabular}

%% file: tables/dualview_spearman.tex
\begin{tabular}{llcccccccc}
\toprule
\textbf{Dataset} & \textbf{LLM} & \textbf{SAJA} & \textbf{T-MES} & \textbf{RELISH} & \textbf{MT-ICL} & \textbf{GenRe2} & \textbf{UniSO-N} & \textbf{GAT-AES} & \textbf{\method{} (Ours)} \\
\midrule
CreativeEval & Gemma E4B & \result{0.053}{0.016} & \result{0.079}{0.045} & \result{0.073}{0.022} & \result{0.099}{0.026} & \result{0.048}{0.008} & \result{0.094}{0.040} & \result{0.066}{0.020} & \cellcolor{dvgreen}\best{0.162}{0.036} \\
 & Qwen 9B & \result{0.045}{0.008} & \result{0.086}{0.053} & \result{0.103}{0.032} & \result{0.084}{0.033} & \result{0.085}{0.064} & \result{0.090}{0.029} & \result{0.136}{0.035} & \cellcolor{dvgreen}\best{0.191}{0.043} \\
 & Qwen 27B & \result{0.076}{0.011} & \result{0.109}{0.046} & \result{0.088}{0.045} & \result{0.119}{0.051} & \result{0.080}{0.038} & \result{0.063}{0.017} & \result{0.151}{0.070} & \cellcolor{dvgreen}\best{0.199}{0.045} \\
 & Gemma 31B & \result{0.086}{0.043} & \result{0.108}{0.053} & \result{0.127}{0.035} & \result{0.142}{0.040} & \result{0.112}{0.046} & \result{0.118}{0.043} & \result{0.127}{0.029} & \cellcolor{dvgreen}\best{0.205}{0.043} \\
 & Llama 70B & \result{0.082}{0.029} & \result{0.100}{0.016} & \result{0.095}{0.046} & \result{0.112}{0.030} & \result{0.074}{0.035} & \result{0.128}{0.017} & \result{0.062}{0.003} & \cellcolor{dvgreen}\best{0.193}{0.015} \\
 & Qwen 72B & \result{0.060}{0.024} & \result{0.116}{0.040} & \result{0.086}{0.040} & \result{0.113}{0.050} & \result{0.108}{0.040} & \result{0.107}{0.055} & \result{0.105}{0.024} & \cellcolor{dvgreen}\best{0.207}{0.045} \\
 & GPT 120B & \result{0.071}{0.029} & \result{0.083}{0.029} & \result{0.073}{0.022} & \result{0.083}{0.024} & \result{0.076}{0.044} & \result{0.102}{0.011} & \result{0.057}{0.019} & \cellcolor{dvgreen}\best{0.148}{0.042} \\
 & Qwen 122B & \result{0.071}{0.005} & \result{0.112}{0.061} & \result{0.093}{0.006} & \result{0.156}{0.050} & \result{0.093}{0.049} & \result{0.123}{0.021} & \result{0.083}{0.037} & \cellcolor{dvgreen}\best{0.219}{0.046} \\
\rowcolor{dvlight}  & Average LLM & \result{0.068}{0.002} & \result{0.099}{0.040} & \result{0.092}{0.014} & \result{0.114}{0.017} & \result{0.084}{0.027} & \result{0.103}{0.009} & \result{0.098}{0.019} & \cellcolor{dvgreen}\best{0.190}{0.035} \\
\midrule
MSumBench & Gemma E4B & \result{0.449}{0.023} & \result{0.327}{0.091} & \result{0.496}{0.018} & \result{0.528}{0.017} & \result{0.560}{0.016} & \result{0.574}{0.012} & \result{0.562}{0.024} & \cellcolor{dvgreen}\best{0.591}{0.008} \\
 & Qwen 9B & \result{0.411}{0.073} & \result{0.354}{0.083} & \result{0.475}{0.041} & \result{0.531}{0.011} & \result{0.576}{0.016} & \result{0.541}{0.028} & \result{0.586}{0.017} & \cellcolor{dvgreen}\best{0.589}{0.022} \\
 & Qwen 27B & \result{0.446}{0.026} & \result{0.363}{0.082} & \result{0.536}{0.033} & \result{0.572}{0.018} & \result{0.586}{0.010} & \result{0.579}{0.038} & \result{0.613}{0.024} & \cellcolor{dvgreen}\best{0.632}{0.015} \\
 & Gemma 31B & \result{0.424}{0.044} & \result{0.441}{0.027} & \result{0.419}{0.072} & \result{0.570}{0.023} & \result{0.568}{0.022} & \result{0.578}{0.049} & \result{0.568}{0.028} & \cellcolor{dvgreen}\best{0.613}{0.020} \\
 & Llama 70B & \result{0.445}{0.044} & \result{0.318}{0.082} & \result{0.483}{0.027} & \result{0.528}{0.048} & \result{0.555}{0.029} & \result{0.547}{0.032} & \result{0.547}{0.022} & \cellcolor{dvgreen}\best{0.592}{0.035} \\
 & Qwen 72B & \result{0.468}{0.070} & \result{0.331}{0.064} & \result{0.468}{0.087} & \result{0.590}{0.012} & \result{0.570}{0.021} & \result{0.587}{0.045} & \result{0.576}{0.031} & \cellcolor{dvgreen}\best{0.618}{0.021} \\
 & GPT 120B & \result{0.472}{0.019} & \result{0.396}{0.016} & \result{0.517}{0.014} & \result{0.501}{0.029} & \result{0.516}{0.036} & \result{0.487}{0.049} & \result{0.523}{0.025} & \cellcolor{dvgreen}\best{0.556}{0.021} \\
 & Qwen 122B & \result{0.421}{0.022} & \result{0.353}{0.075} & \result{0.521}{0.032} & \result{0.551}{0.031} & \result{0.579}{0.028} & \result{0.599}{0.021} & \result{0.589}{0.033} & \cellcolor{dvgreen}\best{0.618}{0.009} \\
\rowcolor{dvlight}  & Average LLM & \result{0.442}{0.037} & \result{0.360}{0.064} & \result{0.489}{0.029} & \result{0.546}{0.021} & \result{0.564}{0.019} & \result{0.562}{0.030} & \result{0.571}{0.025} & \cellcolor{dvgreen}\best{0.601}{0.017} \\
\midrule
RATE & Gemma E4B & \result{0.179}{0.033} & \result{0.270}{0.006} & \result{0.268}{0.022} & \result{0.295}{0.045} & \result{0.285}{0.042} & \result{0.292}{0.041} & \result{0.333}{0.043} & \cellcolor{dvgreen}\best{0.351}{0.043} \\
 & Qwen 9B & \result{0.184}{0.030} & \result{0.315}{0.015} & \result{0.288}{0.045} & \result{0.270}{0.028} & \result{0.271}{0.011} & \result{0.296}{0.019} & \result{0.304}{0.035} & \cellcolor{dvgreen}\best{0.355}{0.023} \\
 & Qwen 27B & \result{0.287}{0.016} & \result{0.290}{0.037} & \result{0.299}{0.059} & \result{0.353}{0.040} & \result{0.344}{0.047} & \result{0.360}{0.055} & \result{0.365}{0.029} & \cellcolor{dvgreen}\best{0.422}{0.049} \\
 & Gemma 31B & \result{0.306}{0.011} & \result{0.365}{0.023} & \result{0.342}{0.030} & \result{0.415}{0.005} & \result{0.379}{0.022} & \result{0.336}{0.029} & \result{0.428}{0.046} & \cellcolor{dvgreen}\best{0.448}{0.023} \\
 & Llama 70B & \result{0.167}{0.062} & \result{0.301}{0.020} & \result{0.278}{0.046} & \result{0.284}{0.020} & \result{0.319}{0.019} & \result{0.272}{0.044} & \result{0.306}{0.038} & \cellcolor{dvgreen}\best{0.362}{0.041} \\
 & Qwen 72B & \result{0.202}{0.022} & \result{0.275}{0.035} & \result{0.289}{0.026} & \result{0.317}{0.039} & \result{0.307}{0.046} & \result{0.318}{0.045} & \result{0.345}{0.068} & \cellcolor{dvgreen}\best{0.378}{0.051} \\
 & GPT 120B & \result{0.189}{0.025} & \result{0.280}{0.027} & \result{0.291}{0.066} & \result{0.261}{0.005} & \result{0.276}{0.031} & \result{0.217}{0.028} & \result{0.232}{0.036} & \cellcolor{dvgreen}\best{0.318}{0.001} \\
 & Qwen 122B & \result{0.285}{0.047} & \result{0.310}{0.009} & \result{0.293}{0.028} & \result{0.344}{0.018} & \result{0.348}{0.012} & \result{0.345}{0.025} & \result{0.371}{0.020} & \cellcolor{dvgreen}\best{0.418}{0.023} \\
\rowcolor{dvlight}  & Average LLM & \result{0.225}{0.015} & \result{0.301}{0.016} & \result{0.294}{0.035} & \result{0.317}{0.013} & \result{0.316}{0.021} & \result{0.305}{0.019} & \result{0.335}{0.036} & \cellcolor{dvgreen}\best{0.381}{0.028} \\
\midrule
PRISM & Gemma E4B & \result{0.162}{0.023} & \result{0.283}{0.011} & \result{0.277}{0.014} & \result{0.230}{0.019} & \result{0.240}{0.001} & \result{0.226}{0.010} & \result{0.248}{0.008} & \cellcolor{dvgreen}\best{0.317}{0.012} \\
 & Qwen 9B & \result{0.145}{0.010} & \result{0.282}{0.014} & \result{0.290}{0.014} & \result{0.240}{0.023} & \result{0.240}{0.016} & \result{0.293}{0.020} & \result{0.263}{0.010} & \cellcolor{dvgreen}\best{0.327}{0.008} \\
 & Qwen 27B & \result{0.138}{0.009} & \result{0.297}{0.003} & \result{0.301}{0.006} & \result{0.232}{0.013} & \result{0.226}{0.018} & \result{0.251}{0.008} & \result{0.269}{0.003} & \cellcolor{dvgreen}\best{0.315}{0.021} \\
 & Gemma 31B & \result{0.137}{0.011} & \result{0.274}{0.027} & \result{0.262}{0.038} & \result{0.223}{0.008} & \result{0.243}{0.014} & \result{0.235}{0.008} & \result{0.268}{0.009} & \cellcolor{dvgreen}\best{0.312}{0.019} \\
 & Llama 70B & \result{0.157}{0.003} & \result{0.314}{0.014} & \result{0.278}{0.055} & \result{0.243}{0.015} & \result{0.277}{0.030} & \result{0.259}{0.013} & \result{0.276}{0.001} & \cellcolor{dvgreen}\best{0.356}{0.016} \\
 & Qwen 72B & \result{0.170}{0.006} & \result{0.295}{0.010} & \result{0.247}{0.042} & \result{0.260}{0.014} & \result{0.264}{0.023} & \result{0.250}{0.022} & \result{0.235}{0.013} & \cellcolor{dvgreen}\best{0.341}{0.017} \\
 & GPT 120B & \result{0.144}{0.012} & \result{0.255}{0.011} & \result{0.311}{0.022} & \result{0.214}{0.011} & \result{0.203}{0.032} & \result{0.204}{0.006} & \result{0.200}{0.022} & \cellcolor{dvgreen}\best{0.337}{0.018} \\
 & Qwen 122B & \result{0.133}{0.007} & \result{0.294}{0.013} & \cellcolor{dvyellow}\best{0.345}{0.019} & \result{0.244}{0.005} & \result{0.220}{0.014} & \result{0.243}{0.023} & \result{0.243}{0.013} & \result{0.319}{0.020} \\
\rowcolor{dvlight}  & Average LLM & \result{0.148}{0.005} & \result{0.287}{0.010} & \result{0.289}{0.019} & \result{0.236}{0.010} & \result{0.239}{0.011} & \result{0.245}{0.011} & \result{0.250}{0.004} & \cellcolor{dvgreen}\best{0.328}{0.015} \\
\midrule
DSCodeBench & Gemma E4B & \result{0.295}{0.055} & \result{0.577}{0.019} & \result{0.555}{0.069} & \result{0.583}{0.012} & \result{0.533}{0.032} & \result{0.571}{0.061} & \result{0.625}{0.028} & \cellcolor{dvgreen}\best{0.629}{0.028} \\
 & Qwen 9B & \result{0.295}{0.046} & \result{0.577}{0.009} & \result{0.582}{0.009} & \result{0.501}{0.140} & \result{0.532}{0.019} & \result{0.553}{0.020} & \result{0.624}{0.058} & \cellcolor{dvgreen}\best{0.626}{0.016} \\
 & Qwen 27B & \result{0.333}{0.045} & \result{0.580}{0.009} & \result{0.621}{0.039} & \result{0.660}{0.030} & \result{0.557}{0.014} & \result{0.601}{0.029} & \result{0.639}{0.035} & \cellcolor{dvgreen}\best{0.662}{0.021} \\
 & Gemma 31B & \result{0.362}{0.044} & \result{0.584}{0.023} & \result{0.553}{0.025} & \result{0.674}{0.023} & \result{0.558}{0.012} & \result{0.629}{0.039} & \result{0.625}{0.017} & \cellcolor{dvgreen}\best{0.684}{0.030} \\
 & Llama 70B & \result{0.295}{0.025} & \result{0.559}{0.012} & \result{0.574}{0.072} & \result{0.576}{0.107} & \result{0.519}{0.010} & \result{0.597}{0.038} & \result{0.616}{0.023} & \cellcolor{dvgreen}\best{0.660}{0.011} \\
 & Qwen 72B & \result{0.245}{0.036} & \result{0.557}{0.013} & \result{0.545}{0.039} & \result{0.530}{0.079} & \result{0.537}{0.016} & \result{0.635}{0.028} & \result{0.576}{0.042} & \cellcolor{dvgreen}\best{0.662}{0.022} \\
 & GPT 120B & \result{0.279}{0.011} & \result{0.548}{0.031} & \result{0.625}{0.023} & \result{0.460}{0.149} & \result{0.543}{0.003} & \result{0.515}{0.040} & \result{0.561}{0.027} & \cellcolor{dvgreen}\best{0.632}{0.017} \\
 & Qwen 122B & \result{0.274}{0.021} & \result{0.573}{0.005} & \result{0.595}{0.040} & \result{0.515}{0.084} & \result{0.522}{0.009} & \result{0.617}{0.063} & \result{0.613}{0.045} & \cellcolor{dvgreen}\best{0.651}{0.022} \\
\rowcolor{dvlight}  & Average LLM & \result{0.297}{0.033} & \result{0.569}{0.010} & \result{0.581}{0.022} & \result{0.562}{0.061} & \result{0.537}{0.006} & \result{0.590}{0.026} & \result{0.610}{0.025} & \cellcolor{dvgreen}\best{0.651}{0.017} \\
\midrule
Overall &  & \result{0.236}{0.012} & \result{0.323}{0.003} & \result{0.349}{0.018} & \result{0.355}{0.019} & \result{0.348}{0.005} & \result{0.361}{0.009} & \result{0.373}{0.003} & \cellcolor{dvgreen}\best{0.430}{0.008} \\
\bottomrule
\end{tabular}

%% file: tables/dualview_relative_rmse.tex
\begin{tabular}{llcccccccc}
\toprule
\textbf{Dataset} & \textbf{LLM} & \textbf{SAJA} & \textbf{T-MES} & \textbf{RELISH} & \textbf{MT-ICL} & \textbf{GenRe2} & \textbf{UniSO-N} & \textbf{GAT-AES} & \textbf{\method{} (Ours)} \\
\midrule
CreativeEval & Gemma E4B & \result{1.001}{0.045} & \result{1.057}{0.033} & \result{1.100}{0.013} & \result{1.017}{0.042} & \result{1.012}{0.042} & \result{1.020}{0.032} & \result{1.003}{0.045} & \cellcolor{dvgreen}\best{0.988}{0.037} \\
 & Qwen 9B & \result{1.000}{0.046} & \result{1.025}{0.026} & \result{1.078}{0.026} & \result{1.039}{0.022} & \result{1.013}{0.030} & \result{1.052}{0.041} & \result{0.997}{0.045} & \cellcolor{dvgreen}\best{0.982}{0.035} \\
 & Qwen 27B & \result{0.998}{0.044} & \result{1.023}{0.025} & \result{1.062}{0.033} & \result{1.011}{0.062} & \result{1.027}{0.036} & \result{1.073}{0.045} & \result{0.992}{0.036} & \cellcolor{dvgreen}\best{0.981}{0.034} \\
 & Gemma 31B & \result{1.003}{0.045} & \result{1.029}{0.036} & \result{1.052}{0.054} & \result{0.995}{0.050} & \result{0.999}{0.044} & \result{1.043}{0.102} & \result{0.996}{0.045} & \cellcolor{dvgreen}\best{0.980}{0.043} \\
 & Llama 70B & \result{1.000}{0.049} & \result{1.019}{0.045} & \result{1.048}{0.037} & \result{0.999}{0.045} & \result{1.008}{0.042} & \result{1.023}{0.025} & \result{1.006}{0.044} & \cellcolor{dvgreen}\best{0.982}{0.041} \\
 & Qwen 72B & \result{1.000}{0.047} & \result{1.048}{0.022} & \result{1.038}{0.029} & \result{0.996}{0.040} & \result{1.010}{0.045} & \result{1.050}{0.030} & \result{1.007}{0.044} & \cellcolor{dvgreen}\best{0.981}{0.037} \\
 & GPT 120B & \result{0.996}{0.044} & \result{1.159}{0.020} & \result{1.077}{0.044} & \result{1.024}{0.048} & \result{1.001}{0.049} & \result{1.032}{0.040} & \result{1.007}{0.046} & \cellcolor{dvgreen}\best{0.995}{0.033} \\
 & Qwen 122B & \result{1.001}{0.047} & \result{1.022}{0.024} & \result{1.088}{0.014} & \result{0.990}{0.053} & \result{1.013}{0.039} & \result{1.036}{0.095} & \result{1.004}{0.045} & \cellcolor{dvgreen}\best{0.975}{0.043} \\
\rowcolor{dvlight}  & Average LLM & \result{1.000}{0.046} & \result{1.048}{0.026} & \result{1.068}{0.024} & \result{1.009}{0.044} & \result{1.010}{0.040} & \result{1.041}{0.046} & \result{1.001}{0.043} & \cellcolor{dvgreen}\best{0.983}{0.038} \\
\midrule
MSumBench & Gemma E4B & \result{0.757}{0.074} & \result{0.937}{0.069} & \result{0.785}{0.066} & \result{0.748}{0.049} & \result{0.720}{0.068} & \result{0.720}{0.048} & \result{0.715}{0.061} & \cellcolor{dvgreen}\best{0.694}{0.058} \\
 & Qwen 9B & \result{0.769}{0.085} & \result{0.911}{0.065} & \result{0.821}{0.039} & \result{0.735}{0.059} & \result{0.718}{0.065} & \result{0.749}{0.065} & \result{0.719}{0.057} & \cellcolor{dvgreen}\best{0.693}{0.060} \\
 & Qwen 27B & \result{0.766}{0.069} & \result{0.915}{0.066} & \result{0.762}{0.051} & \result{0.711}{0.062} & \result{0.707}{0.066} & \result{0.728}{0.067} & \result{0.686}{0.052} & \cellcolor{dvgreen}\best{0.673}{0.059} \\
 & Gemma 31B & \result{0.770}{0.072} & \result{0.861}{0.043} & \result{0.876}{0.104} & \result{0.722}{0.062} & \result{0.717}{0.065} & \result{0.726}{0.069} & \result{0.705}{0.067} & \cellcolor{dvgreen}\best{0.682}{0.060} \\
 & Llama 70B & \result{0.761}{0.085} & \result{0.939}{0.056} & \result{0.802}{0.063} & \result{0.737}{0.062} & \result{0.718}{0.074} & \result{0.730}{0.067} & \result{0.726}{0.067} & \cellcolor{dvgreen}\best{0.701}{0.073} \\
 & Qwen 72B & \result{0.752}{0.089} & \result{0.939}{0.049} & \result{0.821}{0.143} & \result{0.694}{0.062} & \result{0.710}{0.061} & \result{0.715}{0.047} & \result{0.701}{0.076} & \cellcolor{dvgreen}\best{0.677}{0.051} \\
 & GPT 120B & \result{0.749}{0.069} & \result{0.913}{0.028} & \result{0.780}{0.046} & \result{0.760}{0.056} & \result{0.737}{0.071} & \result{0.775}{0.079} & \result{0.751}{0.075} & \cellcolor{dvgreen}\best{0.719}{0.065} \\
 & Qwen 122B & \result{0.761}{0.076} & \result{0.918}{0.073} & \result{0.758}{0.074} & \result{0.719}{0.065} & \result{0.713}{0.070} & \result{0.725}{0.055} & \result{0.706}{0.071} & \cellcolor{dvgreen}\best{0.683}{0.057} \\
\rowcolor{dvlight}  & Average LLM & \result{0.761}{0.077} & \result{0.917}{0.056} & \result{0.800}{0.071} & \result{0.728}{0.059} & \result{0.718}{0.067} & \result{0.734}{0.061} & \result{0.714}{0.065} & \cellcolor{dvgreen}\best{0.690}{0.060} \\
\midrule
RATE & Gemma E4B & \result{1.012}{0.016} & \result{0.997}{0.005} & \result{1.031}{0.025} & \result{0.987}{0.006} & \result{0.985}{0.005} & \result{1.042}{0.040} & \result{0.974}{0.022} & \cellcolor{dvgreen}\best{0.963}{0.011} \\
 & Qwen 9B & \result{1.008}{0.006} & \result{0.990}{0.006} & \result{1.022}{0.039} & \result{1.030}{0.023} & \result{1.010}{0.005} & \result{1.018}{0.024} & \result{0.979}{0.022} & \cellcolor{dvgreen}\best{0.961}{0.013} \\
 & Qwen 27B & \result{0.977}{0.007} & \result{0.997}{0.010} & \result{1.017}{0.053} & \result{0.978}{0.024} & \result{1.000}{0.005} & \result{0.990}{0.012} & \result{0.965}{0.016} & \cellcolor{dvgreen}\best{0.933}{0.015} \\
 & Gemma 31B & \result{0.958}{0.020} & \result{1.009}{0.028} & \result{1.021}{0.027} & \result{0.947}{0.019} & \result{0.947}{0.021} & \result{0.993}{0.036} & \result{0.928}{0.039} & \cellcolor{dvgreen}\best{0.911}{0.020} \\
 & Llama 70B & \result{1.016}{0.010} & \result{0.989}{0.005} & \result{1.029}{0.020} & \result{0.996}{0.024} & \result{0.998}{0.011} & \result{1.036}{0.024} & \result{0.993}{0.037} & \cellcolor{dvgreen}\best{0.961}{0.015} \\
 & Qwen 72B & \result{1.003}{0.009} & \result{0.992}{0.010} & \result{1.018}{0.016} & \result{0.993}{0.031} & \result{1.015}{0.023} & \result{1.034}{0.019} & \result{0.967}{0.029} & \cellcolor{dvgreen}\best{0.950}{0.017} \\
 & GPT 120B & \result{1.012}{0.016} & \result{1.025}{0.010} & \result{1.029}{0.026} & \result{1.002}{0.007} & \result{1.000}{0.004} & \result{1.046}{0.010} & \result{1.001}{0.013} & \cellcolor{dvgreen}\best{0.969}{0.011} \\
 & Qwen 122B & \result{0.983}{0.008} & \result{0.985}{0.010} & \result{1.015}{0.023} & \result{0.975}{0.011} & \result{0.974}{0.026} & \result{0.972}{0.018} & \result{0.959}{0.003} & \cellcolor{dvgreen}\best{0.929}{0.011} \\
\rowcolor{dvlight}  & Average LLM & \result{0.996}{0.009} & \result{0.998}{0.006} & \result{1.023}{0.014} & \result{0.988}{0.015} & \result{0.991}{0.006} & \result{1.016}{0.007} & \result{0.971}{0.022} & \cellcolor{dvgreen}\best{0.947}{0.012} \\
\midrule
PRISM & Gemma E4B & \result{1.002}{0.017} & \result{0.981}{0.015} & \result{1.013}{0.017} & \result{0.991}{0.014} & \result{0.996}{0.014} & \result{1.002}{0.013} & \result{0.986}{0.020} & \cellcolor{dvgreen}\best{0.964}{0.018} \\
 & Qwen 9B & \result{1.008}{0.021} & \result{0.983}{0.016} & \result{1.013}{0.027} & \result{0.992}{0.025} & \result{0.998}{0.015} & \result{0.987}{0.012} & \result{0.981}{0.024} & \cellcolor{dvgreen}\best{0.962}{0.020} \\
 & Qwen 27B & \result{1.007}{0.024} & \result{0.979}{0.019} & \result{1.003}{0.026} & \result{0.992}{0.019} & \result{0.998}{0.014} & \result{1.005}{0.013} & \result{0.979}{0.019} & \cellcolor{dvgreen}\best{0.961}{0.022} \\
 & Gemma 31B & \result{1.009}{0.023} & \result{0.986}{0.017} & \result{1.024}{0.010} & \result{0.992}{0.020} & \result{0.994}{0.003} & \result{1.000}{0.023} & \result{0.978}{0.018} & \cellcolor{dvgreen}\best{0.965}{0.019} \\
 & Llama 70B & \result{1.008}{0.021} & \result{0.975}{0.014} & \result{1.006}{0.005} & \result{0.988}{0.018} & \result{0.995}{0.024} & \result{0.996}{0.022} & \result{0.976}{0.020} & \cellcolor{dvgreen}\best{0.951}{0.018} \\
 & Qwen 72B & \result{1.000}{0.022} & \result{0.978}{0.020} & \result{1.038}{0.006} & \result{0.986}{0.019} & \result{0.994}{0.021} & \result{1.000}{0.035} & \result{0.990}{0.024} & \cellcolor{dvgreen}\best{0.959}{0.022} \\
 & GPT 120B & \result{1.010}{0.019} & \result{1.028}{0.019} & \result{1.009}{0.034} & \result{0.998}{0.019} & \result{1.007}{0.022} & \result{1.020}{0.008} & \result{0.998}{0.020} & \cellcolor{dvgreen}\best{0.959}{0.021} \\
 & Qwen 122B & \result{1.007}{0.023} & \result{0.976}{0.017} & \result{0.993}{0.028} & \result{0.991}{0.024} & \result{1.014}{0.020} & \result{0.997}{0.015} & \result{0.988}{0.020} & \cellcolor{dvgreen}\best{0.965}{0.020} \\
\rowcolor{dvlight}  & Average LLM & \result{1.006}{0.021} & \result{0.986}{0.017} & \result{1.012}{0.016} & \result{0.991}{0.020} & \result{1.000}{0.016} & \result{1.001}{0.018} & \result{0.985}{0.020} & \cellcolor{dvgreen}\best{0.961}{0.020} \\
\midrule
DSCodeBench & Gemma E4B & \result{0.891}{0.275} & \result{0.827}{0.278} & \result{0.842}{0.247} & \result{0.850}{0.278} & \result{0.829}{0.278} & \result{0.864}{0.261} & \result{0.836}{0.281} & \cellcolor{dvgreen}\best{0.803}{0.306} \\
 & Qwen 9B & \result{0.891}{0.271} & \result{0.830}{0.291} & \result{0.882}{0.282} & \result{0.866}{0.263} & \result{0.822}{0.296} & \result{0.854}{0.294} & \result{0.818}{0.311} & \cellcolor{dvgreen}\best{0.806}{0.298} \\
 & Qwen 27B & \result{0.884}{0.274} & \result{0.825}{0.297} & \result{0.857}{0.274} & \result{0.821}{0.284} & \result{0.837}{0.334} & \result{0.797}{0.300} & \result{0.793}{0.317} & \cellcolor{dvgreen}\best{0.777}{0.309} \\
 & Gemma 31B & \result{0.863}{0.274} & \result{0.818}{0.290} & \result{0.877}{0.233} & \result{0.815}{0.293} & \result{0.792}{0.289} & \result{0.798}{0.307} & \result{0.810}{0.279} & \cellcolor{dvgreen}\best{0.771}{0.316} \\
 & Llama 70B & \result{0.890}{0.270} & \result{0.841}{0.269} & \result{0.869}{0.280} & \result{0.868}{0.260} & \result{0.829}{0.292} & \result{0.859}{0.244} & \result{0.823}{0.290} & \cellcolor{dvgreen}\best{0.795}{0.301} \\
 & Qwen 72B & \result{0.895}{0.268} & \result{0.824}{0.283} & \result{0.877}{0.276} & \result{0.866}{0.271} & \result{0.838}{0.289} & \result{0.795}{0.299} & \result{0.837}{0.295} & \cellcolor{dvgreen}\best{0.771}{0.303} \\
 & GPT 120B & \result{0.882}{0.271} & \result{0.859}{0.289} & \result{0.828}{0.293} & \result{0.861}{0.236} & \result{0.818}{0.291} & \result{0.836}{0.286} & \result{0.821}{0.288} & \cellcolor{dvgreen}\best{0.798}{0.293} \\
 & Qwen 122B & \result{0.885}{0.269} & \result{0.826}{0.289} & \result{0.843}{0.274} & \result{0.864}{0.269} & \result{0.821}{0.285} & \result{0.840}{0.270} & \result{0.822}{0.276} & \cellcolor{dvgreen}\best{0.778}{0.303} \\
\rowcolor{dvlight}  & Average LLM & \result{0.885}{0.271} & \result{0.831}{0.285} & \result{0.859}{0.270} & \result{0.851}{0.269} & \result{0.823}{0.294} & \result{0.830}{0.282} & \result{0.820}{0.292} & \cellcolor{dvgreen}\best{0.787}{0.303} \\
\midrule
Overall &  & \result{0.930}{0.070} & \result{0.956}{0.065} & \result{0.953}{0.070} & \result{0.914}{0.063} & \result{0.908}{0.072} & \result{0.924}{0.071} & \result{0.898}{0.070} & \cellcolor{dvgreen}\best{0.873}{0.073} \\
\bottomrule
\end{tabular}

%% file: tables/dualview_output_spread_main.tex
\begin{tabular}{@{}l c c c@{}}
\toprule
\textbf{Method} & $\bm{r}$ & $\bm{\rho}$ & \textbf{NRMSE} \\
\midrule
\multicolumn{4}{@{}l}{\textbf{Strict output win rate (\%) $\bm{\uparrow}$}} \\
GenRe2 & \result{5.0}{2.7} & \result{6.8}{2.7} & \result{4.4}{2.0} \\
UniSO-N & \result{4.6}{2.1} & \result{5.0}{4.0} & \result{3.9}{1.8} \\
GAT-AES & \result{8.9}{4.0} & \result{8.5}{4.0} & \result{7.7}{2.3} \\
\textbf{\method{} (Ours)} & \cellcolor{dvgreen}\best{65.9}{5.6} & \cellcolor{dvgreen}\best{60.7}{6.5} & \cellcolor{dvgreen}\best{71.4}{2.4} \\
\midrule
\multicolumn{4}{@{}l}{\textbf{Mean worst-output gap $\bm{\downarrow}$}} \\
GenRe2 & \result{0.178}{0.038} & \result{0.197}{0.005} & \result{0.065}{0.008} \\
UniSO-N & \result{0.153}{0.002} & \result{0.140}{0.017} & \result{0.092}{0.011} \\
GAT-AES & \result{0.149}{0.022} & \result{0.115}{0.010} & \result{0.055}{0.005} \\
\textbf{\method{} (Ours)} & \cellcolor{dvgreen}\best{0.039}{0.005} & \cellcolor{dvgreen}\best{0.033}{0.001} & \cellcolor{dvgreen}\best{0.016}{0.002} \\
\bottomrule
\end{tabular}

%% file: tables/dualview_power_mean_full.tex
\begin{tabular}{@{}lccccc>{\columncolor{gray!10}}c@{}}
\toprule
\textbf{Method} & \textbf{CreativeEval} & \textbf{MSumBench} & \textbf{RATE} & \textbf{PRISM} & \textbf{DSCodeBench} & \textbf{Average Dataset} \\
\midrule
\multicolumn{7}{@{}l}{\textbf{Harmonic Pearson $\uparrow$}} \\
\addlinespace[2pt]
SAJA & $0.039_{\scriptscriptstyle\pm 0.009}$ & $0.644_{\scriptscriptstyle\pm 0.065}$ & $0.162_{\scriptscriptstyle\pm 0.033}$ & $0.182_{\scriptscriptstyle\pm 0.011}$ & $0.080_{\scriptscriptstyle\pm 0.035}$ & $0.221_{\scriptscriptstyle\pm 0.008}$ \\
T-MES & $0.071_{\scriptscriptstyle\pm 0.049}$ & $0.376_{\scriptscriptstyle\pm 0.042}$ & $0.228_{\scriptscriptstyle\pm 0.036}$ & $0.256_{\scriptscriptstyle\pm 0.011}$ & $0.281_{\scriptscriptstyle\pm 0.215}$ & $0.242_{\scriptscriptstyle\pm 0.046}$ \\
RELISH & $0.056_{\scriptscriptstyle\pm 0.018}$ & $0.636_{\scriptscriptstyle\pm 0.042}$ & $0.229_{\scriptscriptstyle\pm 0.063}$ & $0.276_{\scriptscriptstyle\pm 0.023}$ & $0.351_{\scriptscriptstyle\pm 0.109}$ & $0.309_{\scriptscriptstyle\pm 0.037}$ \\
MT-ICL & $0.094_{\scriptscriptstyle\pm 0.010}$ & $0.682_{\scriptscriptstyle\pm 0.042}$ & $0.236_{\scriptscriptstyle\pm 0.051}$ & $0.255_{\scriptscriptstyle\pm 0.008}$ & $0.274_{\scriptscriptstyle\pm 0.186}$ & $0.308_{\scriptscriptstyle\pm 0.040}$ \\
GenRe2 & $0.070_{\scriptscriptstyle\pm 0.035}$ & $0.696_{\scriptscriptstyle\pm 0.047}$ & $0.257_{\scriptscriptstyle\pm 0.027}$ & $0.253_{\scriptscriptstyle\pm 0.016}$ & $0.185_{\scriptscriptstyle\pm 0.136}$ & $0.292_{\scriptscriptstyle\pm 0.027}$ \\
UniSO-N & $0.076_{\scriptscriptstyle\pm 0.011}$ & $0.683_{\scriptscriptstyle\pm 0.046}$ & $0.251_{\scriptscriptstyle\pm 0.030}$ & $0.259_{\scriptscriptstyle\pm 0.018}$ & $0.313_{\scriptscriptstyle\pm 0.232}$ & $0.316_{\scriptscriptstyle\pm 0.058}$ \\
GAT-AES & $0.075_{\scriptscriptstyle\pm 0.023}$ & $0.702_{\scriptscriptstyle\pm 0.048}$ & $0.289_{\scriptscriptstyle\pm 0.024}$ & $0.277_{\scriptscriptstyle\pm 0.010}$ & $0.285_{\scriptscriptstyle\pm 0.220}$ & $0.326_{\scriptscriptstyle\pm 0.052}$ \\
\textbf{\method{} (Ours)} & \cellcolor{dvgreen}$\mathbf{0.185_{\scriptscriptstyle\pm 0.026}}$ & \cellcolor{dvgreen}$\mathbf{0.717_{\scriptscriptstyle\pm 0.040}}$ & \cellcolor{dvgreen}$\mathbf{0.355_{\scriptscriptstyle\pm 0.027}}$ & \cellcolor{dvgreen}$\mathbf{0.348_{\scriptscriptstyle\pm 0.019}}$ & \cellcolor{dvgreen}$\mathbf{0.450_{\scriptscriptstyle\pm 0.242}}$ & \cellcolor{dvgreen}$\mathbf{0.411_{\scriptscriptstyle\pm 0.050}}$ \\
\midrule
\multicolumn{7}{@{}l}{\textbf{Harmonic Spearman $\uparrow$}} \\
\addlinespace[2pt]
SAJA & $0.048_{\scriptscriptstyle\pm 0.007}$ & $0.401_{\scriptscriptstyle\pm 0.035}$ & $0.157_{\scriptscriptstyle\pm 0.034}$ & $0.141_{\scriptscriptstyle\pm 0.006}$ & $0.230_{\scriptscriptstyle\pm 0.052}$ & $0.195_{\scriptscriptstyle\pm 0.012}$ \\
T-MES & $0.060_{\scriptscriptstyle\pm 0.042}$ & $0.312_{\scriptscriptstyle\pm 0.054}$ & $0.207_{\scriptscriptstyle\pm 0.071}$ & $0.246_{\scriptscriptstyle\pm 0.009}$ & $0.542_{\scriptscriptstyle\pm 0.016}$ & $0.274_{\scriptscriptstyle\pm 0.018}$ \\
RELISH & $0.058_{\scriptscriptstyle\pm 0.020}$ & $0.456_{\scriptscriptstyle\pm 0.024}$ & $0.207_{\scriptscriptstyle\pm 0.067}$ & $0.282_{\scriptscriptstyle\pm 0.018}$ & $0.568_{\scriptscriptstyle\pm 0.024}$ & $0.314_{\scriptscriptstyle\pm 0.014}$ \\
MT-ICL & $0.090_{\scriptscriptstyle\pm 0.019}$ & $0.515_{\scriptscriptstyle\pm 0.017}$ & $0.224_{\scriptscriptstyle\pm 0.050}$ & $0.228_{\scriptscriptstyle\pm 0.010}$ & $0.523_{\scriptscriptstyle\pm 0.081}$ & $0.316_{\scriptscriptstyle\pm 0.018}$ \\
GenRe2 & $0.063_{\scriptscriptstyle\pm 0.031}$ & $0.531_{\scriptscriptstyle\pm 0.016}$ & $0.244_{\scriptscriptstyle\pm 0.057}$ & $0.234_{\scriptscriptstyle\pm 0.011}$ & $0.275_{\scriptscriptstyle\pm 0.033}$ & $0.269_{\scriptscriptstyle\pm 0.015}$ \\
UniSO-N & $0.072_{\scriptscriptstyle\pm 0.010}$ & $0.530_{\scriptscriptstyle\pm 0.026}$ & $0.210_{\scriptscriptstyle\pm 0.048}$ & $0.239_{\scriptscriptstyle\pm 0.012}$ & $0.536_{\scriptscriptstyle\pm 0.034}$ & $0.317_{\scriptscriptstyle\pm 0.007}$ \\
GAT-AES & $0.074_{\scriptscriptstyle\pm 0.016}$ & $0.541_{\scriptscriptstyle\pm 0.017}$ & $0.245_{\scriptscriptstyle\pm 0.066}$ & $0.244_{\scriptscriptstyle\pm 0.005}$ & $0.597_{\scriptscriptstyle\pm 0.015}$ & $0.340_{\scriptscriptstyle\pm 0.011}$ \\
\textbf{\method{} (Ours)} & \cellcolor{dvgreen}$\mathbf{0.175_{\scriptscriptstyle\pm 0.033}}$ & \cellcolor{dvgreen}$\mathbf{0.575_{\scriptscriptstyle\pm 0.013}}$ & \cellcolor{dvgreen}$\mathbf{0.314_{\scriptscriptstyle\pm 0.072}}$ & \cellcolor{dvgreen}$\mathbf{0.323_{\scriptscriptstyle\pm 0.016}}$ & \cellcolor{dvgreen}$\mathbf{0.645_{\scriptscriptstyle\pm 0.016}}$ & \cellcolor{dvgreen}$\mathbf{0.406_{\scriptscriptstyle\pm 0.008}}$ \\
\midrule
\multicolumn{7}{@{}l}{\textbf{RMS NRMSE $\downarrow$}} \\
\addlinespace[2pt]
SAJA & $1.001_{\scriptscriptstyle\pm 0.046}$ & $0.764_{\scriptscriptstyle\pm 0.080}$ & $0.997_{\scriptscriptstyle\pm 0.010}$ & $1.007_{\scriptscriptstyle\pm 0.021}$ & $0.971_{\scriptscriptstyle\pm 0.287}$ & $0.948_{\scriptscriptstyle\pm 0.074}$ \\
T-MES & $1.049_{\scriptscriptstyle\pm 0.026}$ & $0.920_{\scriptscriptstyle\pm 0.054}$ & $1.001_{\scriptscriptstyle\pm 0.008}$ & $0.987_{\scriptscriptstyle\pm 0.017}$ & $0.909_{\scriptscriptstyle\pm 0.302}$ & $0.973_{\scriptscriptstyle\pm 0.069}$ \\
RELISH & $1.069_{\scriptscriptstyle\pm 0.023}$ & $0.804_{\scriptscriptstyle\pm 0.075}$ & $1.024_{\scriptscriptstyle\pm 0.015}$ & $1.013_{\scriptscriptstyle\pm 0.016}$ & $0.948_{\scriptscriptstyle\pm 0.272}$ & $0.972_{\scriptscriptstyle\pm 0.071}$ \\
MT-ICL & $1.010_{\scriptscriptstyle\pm 0.044}$ & $0.734_{\scriptscriptstyle\pm 0.060}$ & $0.991_{\scriptscriptstyle\pm 0.017}$ & $0.992_{\scriptscriptstyle\pm 0.019}$ & $0.936_{\scriptscriptstyle\pm 0.290}$ & $0.932_{\scriptscriptstyle\pm 0.068}$ \\
GenRe2 & $1.011_{\scriptscriptstyle\pm 0.040}$ & $0.724_{\scriptscriptstyle\pm 0.068}$ & $0.994_{\scriptscriptstyle\pm 0.009}$ & $1.000_{\scriptscriptstyle\pm 0.016}$ & $0.903_{\scriptscriptstyle\pm 0.318}$ & $0.926_{\scriptscriptstyle\pm 0.078}$ \\
UniSO-N & $1.042_{\scriptscriptstyle\pm 0.046}$ & $0.740_{\scriptscriptstyle\pm 0.062}$ & $1.019_{\scriptscriptstyle\pm 0.009}$ & $1.001_{\scriptscriptstyle\pm 0.017}$ & $0.905_{\scriptscriptstyle\pm 0.309}$ & $0.942_{\scriptscriptstyle\pm 0.077}$ \\
GAT-AES & $1.002_{\scriptscriptstyle\pm 0.043}$ & $0.720_{\scriptscriptstyle\pm 0.067}$ & $0.973_{\scriptscriptstyle\pm 0.024}$ & $0.985_{\scriptscriptstyle\pm 0.020}$ & $0.899_{\scriptscriptstyle\pm 0.316}$ & $0.916_{\scriptscriptstyle\pm 0.076}$ \\
\textbf{\method{} (Ours)} & \cellcolor{dvgreen}$\mathbf{0.984_{\scriptscriptstyle\pm 0.038}}$ & \cellcolor{dvgreen}$\mathbf{0.696_{\scriptscriptstyle\pm 0.061}}$ & \cellcolor{dvgreen}$\mathbf{0.949_{\scriptscriptstyle\pm 0.014}}$ & \cellcolor{dvgreen}$\mathbf{0.961_{\scriptscriptstyle\pm 0.020}}$ & \cellcolor{dvgreen}$\mathbf{0.868_{\scriptscriptstyle\pm 0.324}}$ & \cellcolor{dvgreen}$\mathbf{0.892_{\scriptscriptstyle\pm 0.077}}$ \\
\bottomrule
\end{tabular}

%% file: tables/dualview_power_mean_sensitivity.tex
\begin{tabular}{@{}lccccc>{\columncolor{gray!10}}c@{}}
\toprule
\textbf{Method} & \textbf{CreativeEval} & \textbf{MSumBench} & \textbf{RATE} & \textbf{PRISM} & \textbf{DSCodeBench} & \textbf{Average Dataset} \\
\midrule
\multicolumn{7}{@{}l}{\textbf{Pearson ($q=-2$) $\uparrow$}} \\
\addlinespace[2pt]
SAJA & $0.034_{\scriptscriptstyle\pm 0.010}$ & $0.643_{\scriptscriptstyle\pm 0.065}$ & $0.144_{\scriptscriptstyle\pm 0.034}$ & $0.179_{\scriptscriptstyle\pm 0.012}$ & $0.066_{\scriptscriptstyle\pm 0.032}$ & $0.213_{\scriptscriptstyle\pm 0.007}$ \\
T-MES & $0.062_{\scriptscriptstyle\pm 0.048}$ & $0.366_{\scriptscriptstyle\pm 0.040}$ & $0.217_{\scriptscriptstyle\pm 0.038}$ & $0.253_{\scriptscriptstyle\pm 0.011}$ & $0.262_{\scriptscriptstyle\pm 0.210}$ & $0.232_{\scriptscriptstyle\pm 0.044}$ \\
RELISH & $0.049_{\scriptscriptstyle\pm 0.019}$ & $0.634_{\scriptscriptstyle\pm 0.042}$ & $0.214_{\scriptscriptstyle\pm 0.069}$ & $0.272_{\scriptscriptstyle\pm 0.023}$ & $0.330_{\scriptscriptstyle\pm 0.112}$ & $0.300_{\scriptscriptstyle\pm 0.037}$ \\
MT-ICL & $0.087_{\scriptscriptstyle\pm 0.013}$ & $0.680_{\scriptscriptstyle\pm 0.042}$ & $0.215_{\scriptscriptstyle\pm 0.056}$ & $0.252_{\scriptscriptstyle\pm 0.008}$ & $0.252_{\scriptscriptstyle\pm 0.190}$ & $0.297_{\scriptscriptstyle\pm 0.040}$ \\
GenRe2 & $0.064_{\scriptscriptstyle\pm 0.033}$ & $0.693_{\scriptscriptstyle\pm 0.046}$ & $0.239_{\scriptscriptstyle\pm 0.032}$ & $0.250_{\scriptscriptstyle\pm 0.015}$ & $0.158_{\scriptscriptstyle\pm 0.122}$ & $0.281_{\scriptscriptstyle\pm 0.025}$ \\
UniSO-N & $0.069_{\scriptscriptstyle\pm 0.011}$ & $0.680_{\scriptscriptstyle\pm 0.045}$ & $0.234_{\scriptscriptstyle\pm 0.034}$ & $0.256_{\scriptscriptstyle\pm 0.018}$ & $0.292_{\scriptscriptstyle\pm 0.234}$ & $0.306_{\scriptscriptstyle\pm 0.059}$ \\
GAT-AES & $0.070_{\scriptscriptstyle\pm 0.024}$ & $0.700_{\scriptscriptstyle\pm 0.048}$ & $0.271_{\scriptscriptstyle\pm 0.026}$ & $0.274_{\scriptscriptstyle\pm 0.009}$ & $0.262_{\scriptscriptstyle\pm 0.220}$ & $0.315_{\scriptscriptstyle\pm 0.052}$ \\
\textbf{\method{} (Ours)} & \cellcolor{dvgreen}$\mathbf{0.180_{\scriptscriptstyle\pm 0.025}}$ & \cellcolor{dvgreen}$\mathbf{0.715_{\scriptscriptstyle\pm 0.039}}$ & \cellcolor{dvgreen}$\mathbf{0.343_{\scriptscriptstyle\pm 0.034}}$ & \cellcolor{dvgreen}$\mathbf{0.346_{\scriptscriptstyle\pm 0.020}}$ & \cellcolor{dvgreen}$\mathbf{0.434_{\scriptscriptstyle\pm 0.252}}$ & \cellcolor{dvgreen}$\mathbf{0.404_{\scriptscriptstyle\pm 0.053}}$ \\
\midrule
\multicolumn{7}{@{}l}{\textbf{Spearman ($q=-2$) $\uparrow$}} \\
\addlinespace[2pt]
SAJA & $0.043_{\scriptscriptstyle\pm 0.008}$ & $0.385_{\scriptscriptstyle\pm 0.035}$ & $0.139_{\scriptscriptstyle\pm 0.034}$ & $0.137_{\scriptscriptstyle\pm 0.006}$ & $0.207_{\scriptscriptstyle\pm 0.056}$ & $0.182_{\scriptscriptstyle\pm 0.014}$ \\
T-MES & $0.051_{\scriptscriptstyle\pm 0.040}$ & $0.299_{\scriptscriptstyle\pm 0.055}$ & $0.185_{\scriptscriptstyle\pm 0.075}$ & $0.242_{\scriptscriptstyle\pm 0.010}$ & $0.529_{\scriptscriptstyle\pm 0.022}$ & $0.261_{\scriptscriptstyle\pm 0.019}$ \\
RELISH & $0.051_{\scriptscriptstyle\pm 0.021}$ & $0.442_{\scriptscriptstyle\pm 0.023}$ & $0.182_{\scriptscriptstyle\pm 0.074}$ & $0.279_{\scriptscriptstyle\pm 0.018}$ & $0.562_{\scriptscriptstyle\pm 0.026}$ & $0.303_{\scriptscriptstyle\pm 0.015}$ \\
MT-ICL & $0.083_{\scriptscriptstyle\pm 0.022}$ & $0.503_{\scriptscriptstyle\pm 0.017}$ & $0.193_{\scriptscriptstyle\pm 0.053}$ & $0.225_{\scriptscriptstyle\pm 0.010}$ & $0.504_{\scriptscriptstyle\pm 0.089}$ & $0.302_{\scriptscriptstyle\pm 0.020}$ \\
GenRe2 & $0.057_{\scriptscriptstyle\pm 0.030}$ & $0.517_{\scriptscriptstyle\pm 0.016}$ & $0.218_{\scriptscriptstyle\pm 0.064}$ & $0.231_{\scriptscriptstyle\pm 0.011}$ & $0.200_{\scriptscriptstyle\pm 0.033}$ & $0.245_{\scriptscriptstyle\pm 0.017}$ \\
UniSO-N & $0.066_{\scriptscriptstyle\pm 0.011}$ & $0.517_{\scriptscriptstyle\pm 0.026}$ & $0.180_{\scriptscriptstyle\pm 0.049}$ & $0.235_{\scriptscriptstyle\pm 0.012}$ & $0.513_{\scriptscriptstyle\pm 0.044}$ & $0.302_{\scriptscriptstyle\pm 0.008}$ \\
GAT-AES & $0.068_{\scriptscriptstyle\pm 0.017}$ & $0.529_{\scriptscriptstyle\pm 0.015}$ & $0.214_{\scriptscriptstyle\pm 0.070}$ & $0.241_{\scriptscriptstyle\pm 0.005}$ & $0.591_{\scriptscriptstyle\pm 0.012}$ & $0.328_{\scriptscriptstyle\pm 0.013}$ \\
\textbf{\method{} (Ours)} & \cellcolor{dvgreen}$\mathbf{0.170_{\scriptscriptstyle\pm 0.034}}$ & \cellcolor{dvgreen}$\mathbf{0.563_{\scriptscriptstyle\pm 0.013}}$ & \cellcolor{dvgreen}$\mathbf{0.288_{\scriptscriptstyle\pm 0.082}}$ & \cellcolor{dvgreen}$\mathbf{0.321_{\scriptscriptstyle\pm 0.017}}$ & \cellcolor{dvgreen}$\mathbf{0.641_{\scriptscriptstyle\pm 0.015}}$ & \cellcolor{dvgreen}$\mathbf{0.397_{\scriptscriptstyle\pm 0.009}}$ \\
\bottomrule
\end{tabular}

%% file: tables/dualview_parameters.tex
\begin{tabular}{lcccccccc}
\toprule
\textbf{Method} & \textbf{Gemma E4B} & \textbf{Qwen 9B} & \textbf{Qwen 27B} & \textbf{Gemma 31B} & \textbf{Llama 70B} & \textbf{Qwen 72B} & \textbf{GPT 120B} & \textbf{Qwen 122B} \\
\midrule
T-MES & 1.975--4.609M & 3.160--7.372M & 3.949--9.214M & 4.146--9.675M & 6.318--14.741M & 6.318--14.741M & 2.222--5.185M & 2.370--5.530M \\
RELISH & 3.026--3.027M & 3.419--3.420M & 3.681--3.682M & 3.747--3.748M & 4.468--4.469M & 4.468--4.469M & 3.108--3.109M & 3.157--3.158M \\
MT-ICL & 1.518M & 1.715M & 1.846M & 1.879M & 2.239M & 2.239M & 1.559M & 1.584M \\
GenRe2 & 3.894--3.993M & 4.287--4.386M & 4.549--4.648M & 4.615--4.714M & 5.336--5.435M & 5.336--5.435M & 3.976--4.075M & 4.025--4.124M \\
UniSO-N & 9.443M & 12.589M & 14.686M & 15.210M & 20.978M & 20.978M & 10.099M & 10.492M \\
GAT-AES & 1.842--1.846M & 2.628--2.632M & 3.152--3.156M & 3.283--3.288M & 4.725--4.729M & 4.725--4.729M & 2.006--2.010M & 2.104--2.108M \\
\textbf{\method{} (Ours)} & 5.1--18.0K & 8.2--28.7K & 10.2--35.9K & 10.8--37.7K & 16.4--57.4K & 16.4--57.4K & 5.8--20.2K & 6.2--21.6K \\
\bottomrule
\end{tabular}

%% file: tables/dualview_gemini_by_dataset.tex
\begin{tabular}{lccccccccc}
\toprule
\textbf{Dataset} & \multicolumn{3}{c}{\textbf{NRMSE} $\bm{\downarrow}$} & \multicolumn{3}{c}{\textbf{Pearson} $\bm{r\uparrow}$} & \multicolumn{3}{c}{\textbf{Spearman} $\bm{\rho\uparrow}$} \\
\cmidrule(lr){2-4}\cmidrule(lr){5-7}\cmidrule(lr){8-10}
 & \textbf{Zero shot} & \textbf{Few shot} & \textbf{\method{} (Ours)} & \textbf{Zero shot} & \textbf{Few shot} & \textbf{\method{} (Ours)} & \textbf{Zero shot} & \textbf{Few shot} & \textbf{\method{} (Ours)} \\
\midrule
CreativeEval & \result{1.976}{0.103} & \result{1.192}{0.086} & \cellcolor{dvgreen}\best{0.980}{0.043} & \result{0.099}{0.025} & \result{0.164}{0.028} & \cellcolor{dvgreen}\best{0.207}{0.030} & \result{0.098}{0.024} & \result{0.157}{0.031} & \cellcolor{dvgreen}\best{0.205}{0.043} \\
MSumBench & \result{1.240}{0.126} & \result{1.021}{0.075} & \cellcolor{dvgreen}\best{0.682}{0.060} & \result{0.629}{0.062} & \result{0.630}{0.065} & \cellcolor{dvgreen}\best{0.729}{0.040} & \result{0.493}{0.044} & \result{0.490}{0.038} & \cellcolor{dvgreen}\best{0.613}{0.020} \\
RATE & \result{1.147}{0.035} & \result{1.103}{0.039} & \cellcolor{dvgreen}\best{0.911}{0.020} & \result{0.331}{0.037} & \result{0.342}{0.028} & \cellcolor{dvgreen}\best{0.453}{0.020} & \result{0.340}{0.032} & \result{0.361}{0.020} & \cellcolor{dvgreen}\best{0.448}{0.023} \\
PRISM & \result{1.208}{0.025} & \result{1.167}{0.059} & \cellcolor{dvgreen}\best{0.965}{0.019} & \result{0.162}{0.010} & \result{0.173}{0.030} & \cellcolor{dvgreen}\best{0.340}{0.022} & \result{0.133}{0.013} & \result{0.140}{0.010} & \cellcolor{dvgreen}\best{0.312}{0.019} \\
DSCodeBench & \result{0.896}{0.294} & \result{0.888}{0.292} & \cellcolor{dvgreen}\best{0.771}{0.316} & \result{0.400}{0.147} & \result{0.433}{0.196} & \cellcolor{dvgreen}\best{0.548}{0.206} & \result{0.572}{0.038} & \result{0.562}{0.018} & \cellcolor{dvgreen}\best{0.684}{0.030} \\
\midrule
\textbf{Average} & \result{1.293}{0.084} & \result{1.074}{0.052} & \cellcolor{dvgreen}\best{0.862}{0.073} & \result{0.324}{0.027} & \result{0.348}{0.041} & \cellcolor{dvgreen}\best{0.455}{0.041} & \result{0.327}{0.016} & \result{0.342}{0.009} & \cellcolor{dvgreen}\best{0.452}{0.010} \\
\bottomrule
\end{tabular}

%% file: sections/dataset_prompts.tex
\section{Dataset Prompt Templates}\label{sec:dataset-prompts}
\textbf{Tables~\ref{tab:prompt-creativeeval}--\ref{tab:prompt-dscodebench}} specify the prompts used for each dataset. Within a dataset, prompts across LLMs share the same task instructions, descriptions for all output targets, and input fields. The broad prompt ends after these components. The focused prompts append a short target-specific instruction to this shared prefix. Specifically, the templates reflect $P_i^b=[s(\mathcal Q);x_i]$ and $P_{ij}^f=[s(\mathcal Q);x_i;q(d_j)]$ (\S~\ref{sec:evidence-extraction}), enabling shared-prefix caching. Braces indicate input fields and target-specific substitutions.

For the LLM-as-a-Judge settings, the broad prompt is the zero-shot query. \textbf{Table~\ref{tab:prompt-few-shot-wrapper}} specifies the template for appending demonstrations drawn from the training split within each seed. 

\begin{table*}[t]
\centering
\setlength{\tabcolsep}{8pt}
\begin{tabular}{@{}p{0.96\textwidth}@{}}
\toprule
{\fontsize{9.5}{11}\selectfont\ttfamily\raggedright
Use the labeled examples to help answer the query.
\par\vspace{5pt}
Labeled examples:
\par\vspace{4pt}
Example 1 input:\par
\{broad prompt for training example 1\}\par
Example 1 answer JSON:\par
\{gold-score JSON object for training example 1\}
\par\vspace{4pt}
\textnormal{\itshape Repeat the example block for examples 2, 3, and 4.}
\par\vspace{5pt}
Query:\par
\{broad prompt for the query example\}
\par}\tabularnewline[6pt]
\bottomrule
\end{tabular}
\caption{\textbf{Shared template for few-shot prompting.}}\label{tab:prompt-few-shot-wrapper}
\end{table*}

\begin{table*}[p]\centering\setlength{\tabcolsep}{8pt}
\begin{tabular}{@{}p{0.96\textwidth}@{}}
\toprule
\multicolumn{1}{c}{\textbf{CreativeEval}} \\[4pt]
\midrule
\textbf{Broad Prompt} \\[4pt]
{\fontsize{9.5}{11}\selectfont\ttfamily\raggedright Given a paper introduction, predict the following metrics:
\par\vspace{3pt}
- contribution: a number from 1 to 4, inclusive, measuring significance, originality, and potential impact.\par - soundness: a number from 1 to 4, inclusive, measuring technical validity and support for the stated claims.\par - presentation: a number from 1 to 4, inclusive, measuring clarity, organization, and effectiveness.\par - overall\_score: a number from 1 to 10, inclusive, measuring holistic quality and promise.
\par\vspace{3pt}
Return exactly one JSON object in this order: \{"contribution": <number>, "soundness": <number>, "presentation": <number>, "overall\_score": <number>\}
\par\vspace{3pt}
Do not output any additional text besides the JSON object.\par\vspace{5pt}
Paper introduction:\par
<introduction>\par
\{paper\_introduction\}\par
</introduction>\par}\tabularnewline[6pt]
\midrule
\textbf{Focused Continuations} \\[4pt]
{\fontsize{9.5}{11}\selectfont\ttfamily\raggedright For this paper introduction, assess only \{output\}: \{definition\}.}\tabularnewline[6pt]
\bottomrule
\end{tabular}
\caption{\textbf{Broad and focused prompts for CreativeEval.} The broad prompt contains the shared task instructions and input fields. Each focused prompt reuses that entire prefix and appends the target-specific suffix, instantiated once per output. Braces mark substitutions with actual data.}\label{tab:prompt-creativeeval}
\end{table*}

\begin{table*}[p]\centering\setlength{\tabcolsep}{8pt}
\begin{tabular}{@{}p{0.96\textwidth}@{}}
\toprule
\multicolumn{1}{c}{\textbf{MSumBench}} \\[4pt]
\midrule
\textbf{Broad Prompt} \\[4pt]
{\fontsize{9.5}{11}\selectfont\ttfamily\raggedright Given a reference document and a summary, predict the following metrics on a scale from 0 to 1, inclusive.
\par\vspace{3pt}
- completeness: the fraction of source-document key facts covered by at least one summary sentence.\par - conciseness: the fraction of summary sentences containing at least one source key fact.\par - faithfulness: the fraction of summary sentences containing no factual errors.
\par\vspace{3pt}
Return exactly one JSON object in this order: \{"completeness": <number>, "conciseness": <number>, "faithfulness": <number>\}
\par\vspace{3pt}
Report each metric to three decimal places. Do not output any additional text besides the JSON object.\par\vspace{5pt}
Source document:\par
<document>\par
\{source\_text\}\par
</document>\par
\vspace{3pt}
Candidate summary:\par
<summary>\par
\{summary\_sentences\_text\}\par
</summary>\par}\tabularnewline[6pt]
\midrule
\textbf{Focused Continuations} \\[4pt]
{\fontsize{9.5}{11}\selectfont\ttfamily\raggedright For this source document and summary, assess only \{output\}: \{definition\}.}\tabularnewline[6pt]
\bottomrule
\end{tabular}
\caption{\textbf{Broad and focused prompts for MSumBench.} The broad prompt contains the shared task instructions and input fields. Each focused prompt reuses that entire prefix and appends the target-specific suffix, instantiated once per output. Braces mark substitutions with actual data.}\label{tab:prompt-msumbench}
\end{table*}

\begin{table*}[p]\centering\setlength{\tabcolsep}{8pt}
\begin{tabular}{@{}p{0.96\textwidth}@{}}
\toprule
\multicolumn{1}{c}{\textbf{RATE}} \\[4pt]
\midrule
\textbf{Broad Prompt} \\[4pt]
{\fontsize{9.5}{11}\selectfont\ttfamily\raggedright Given an English source and a Russian translation, predict the following metrics on a scale from 0 to 100, inclusive:
\par\vspace{3pt}
- accuracy: how well the source's meaning is preserved.\par - fluency: the grammaticality, readability, and naturalness of the translation.\par - style: how well the source's style is preserved.
\par\vspace{3pt}
Return exactly one JSON object in this order: \{"accuracy": <number>, "fluency": <number>, "style": <number>\}
\par\vspace{3pt}
Do not output any additional text besides the JSON object.\par\vspace{5pt}
English source:\par
<source>\par
\{source\_text\}\par
</source>\par
\vspace{3pt}
Russian candidate translation:\par
<translation>\par
\{translation\}\par
</translation>\par}\tabularnewline[6pt]
\midrule
\textbf{Focused Continuations} \\[4pt]
{\fontsize{9.5}{11}\selectfont\ttfamily\raggedright For this English source and Russian translation, assess only \{output\}: \{definition\}.}\tabularnewline[6pt]
\bottomrule
\end{tabular}
\caption{\textbf{Broad and focused prompts for RATE.} The broad prompt contains the shared task instructions and input fields. Each focused prompt reuses that entire prefix and appends the target-specific suffix, instantiated once per output. Braces mark substitutions with actual data.}\label{tab:prompt-rate}
\end{table*}

\begin{table*}[p]\centering\setlength{\tabcolsep}{8pt}
\begin{tabular}{@{}p{0.96\textwidth}@{}}
\toprule
\multicolumn{1}{c}{\textbf{PRISM}} \\[4pt]
\midrule
\textbf{Broad Prompt} \\[4pt]
{\fontsize{9.5}{11}\selectfont\ttfamily\raggedright Given a conversation prompt and response, predict the following metrics on a scale from 1 to 100, inclusive:
\par\vspace{3pt}
values: alignment with the participant's stated values or cultural perspective.\par fluency: grammaticality, readability, and coherence.\par factuality: factual accuracy and avoidance of unsupported claims.\par safety: safety and avoidance of harm.\par diversity: coverage of multiple perspectives or worldviews.\par creativity: creativity and originality.\par helpfulness: helpfulness and relevance to the participant's request.
\par\vspace{3pt}
Return only a JSON object containing the numeric prediction for every output and no additional text.\par\vspace{5pt}
Participant self-description:\par
<self\_description>\par
\{self\_description\}\par
</self\_description>\par
\vspace{3pt}
Participant's desired assistant behavior:\par
<desired\_behavior>\par
\{desired\_model\_behavior\}\par
</desired\_behavior>\par
\vspace{3pt}
Opening prompt:\par
<prompt>\par
\{opening\_prompt\}\par
</prompt>\par
\vspace{3pt}
Candidate response:\par
<response>\par
\{response\}\par
</response>\par}\tabularnewline[6pt]
\midrule
\textbf{Focused Continuations} \\[4pt]
{\fontsize{9.5}{11}\selectfont\ttfamily\raggedright For this conversation prompt and response, assess only \{output\}: \{definition\}.}\tabularnewline[6pt]
\bottomrule
\end{tabular}
\caption{\textbf{Broad and focused prompts for PRISM.} The broad prompt contains the shared task instructions and input fields. Each focused prompt reuses that entire prefix and appends the target-specific suffix, instantiated once per output. Braces mark substitutions with actual data.}\label{tab:prompt-prism}
\end{table*}

\begin{table*}[p]\centering\setlength{\tabcolsep}{8pt}
\begin{tabular}{@{}p{0.96\textwidth}@{}}
\toprule
\multicolumn{1}{c}{\textbf{DSCodeBench}} \\[4pt]
\midrule
\textbf{Broad Prompt} \\[4pt]
{\fontsize{9.5}{11}\selectfont\ttfamily\raggedright Given a programming task and candidate code, predict the following metrics without executing the code:
\par\vspace{3pt}
- correctness\_fraction: a number from 0 to 1, inclusive, measuring the fraction of the fixed 200-test workload the code would pass.\par - wall\_time\_s: a nonnegative number measuring the estimated wall-clock seconds for the fixed workload.\par - peak\_rss\_bytes: a nonnegative number measuring the estimated peak process-tree resident memory in bytes.\par - memory\_integral\_byte\_s: a nonnegative number measuring the estimated time-integrated process-tree resident memory in byte-seconds.
\par\vspace{3pt}
Return exactly one JSON object in this order: \{"correctness\_fraction": <number>, "wall\_time\_s": <number>, "peak\_rss\_bytes": <number>, "memory\_integral\_byte\_s": <number>\}
\par\vspace{3pt}
Do not output any additional text besides the JSON object.\par\vspace{5pt}
Target library:\par
\{library\}\par
\vspace{3pt}
Task instruction:\par
<task>\par
\{task\_instruction\}\par
</task>\par
\vspace{3pt}
Execution context:\par
<execution\_context>\par
\{execution\_context\_json\}\par
</execution\_context>\par
\vspace{3pt}
Profile environment:\par
<profile\_environment>\par
\{profile\_environment\_json\}\par
</profile\_environment>\par
\vspace{3pt}
Test workload:\par
<tests>\par
\{test\_script\}\par
</tests>\par
\vspace{3pt}
Candidate code:\par
<code>\par
\{generated\_code\}\par
</code>\par}\tabularnewline[6pt]
\midrule
\textbf{Focused Continuations} \\[4pt]
{\fontsize{9.5}{11}\selectfont\ttfamily\raggedright For this programming task and candidate code, estimate only \{output\}: \{definition\}. Do not execute the code.}\tabularnewline[6pt]
\bottomrule
\end{tabular}
\caption{\textbf{Broad and focused prompts for DSCodeBench.} The broad prompt contains the shared task instructions and input fields. Each focused prompt reuses that entire prefix and appends the target-specific suffix, instantiated once per output. Braces mark substitutions with actual data.}\label{tab:prompt-dscodebench}
\end{table*}

%% file: tables/dualview_targets.tex
\begin{table*}[p]\centering\small\setlength{\tabcolsep}{3pt}
\resizebox{\textwidth}{!}{
\begin{tabular}{llcccccccc}
\toprule
\textbf{Output} & \textbf{Metric} & \textbf{SAJA} & \textbf{T-MES} & \textbf{RELISH} & \textbf{MT-ICL} & \textbf{GenRe2} & \textbf{UniSO-N} & \textbf{GAT-AES} & \textbf{\method{} (Ours)} \\
\midrule
Contribution & NRMSE $\downarrow$ & \result{1.001}{0.055} & \result{1.058}{0.050} & \result{1.073}{0.012} & \result{1.021}{0.059} & \result{1.012}{0.059} & \result{1.022}{0.034} & \result{1.016}{0.055} & \cellcolor{dvgreen}\best{0.988}{0.040} \\
 & $r\uparrow$ & \result{0.066}{0.011} & \result{0.039}{0.030} & \result{0.107}{0.045} & \result{0.126}{0.059} & \result{0.050}{0.011} & \result{0.089}{0.077} & \result{0.091}{0.081} & \cellcolor{dvgreen}\best{0.176}{0.079} \\
 & $\rho\uparrow$ & \result{0.074}{0.012} & \result{0.057}{0.039} & \result{0.102}{0.026} & \result{0.142}{0.071} & \result{0.044}{0.029} & \result{0.095}{0.083} & \result{0.077}{0.059} & \cellcolor{dvgreen}\best{0.184}{0.085} \\
\addlinespace[3pt]
Soundness & NRMSE $\downarrow$ & \result{0.983}{0.048} & \result{1.046}{0.013} & \result{1.096}{0.039} & \result{1.005}{0.044} & \result{0.998}{0.049} & \result{0.998}{0.025} & \result{0.981}{0.049} & \cellcolor{dvgreen}\best{0.970}{0.039} \\
 & $r\uparrow$ & \result{0.032}{0.009} & \result{0.095}{0.076} & \result{0.018}{0.016} & \result{0.072}{0.057} & \result{0.069}{0.033} & \result{0.086}{0.067} & \result{0.048}{0.009} & \cellcolor{dvgreen}\best{0.126}{0.057} \\
 & $\rho\uparrow$ & \result{0.046}{0.018} & \result{0.107}{0.070} & \result{0.019}{0.016} & \result{0.063}{0.035} & \result{0.051}{0.033} & \result{0.073}{0.061} & \result{0.053}{0.053} & \cellcolor{dvgreen}\best{0.126}{0.052} \\
\addlinespace[3pt]
Presentation & NRMSE $\downarrow$ & \result{0.958}{0.031} & \result{1.010}{0.032} & \result{1.096}{0.057} & \result{0.978}{0.032} & \result{0.962}{0.031} & \result{0.980}{0.029} & \result{0.953}{0.034} & \cellcolor{dvgreen}\best{0.944}{0.029} \\
\bottomrule\end{tabular}}
\caption{CreativeEval, Gemma E4B output-level metrics. Green highlights a winning \method{} cell and yellow highlights a winning baseline cell.}
\label{tab:targets-creativeeval-gemma-e4b}
\end{table*}
\begin{table*}[p]\centering\small\setlength{\tabcolsep}{3pt}
\resizebox{\textwidth}{!}{\begin{tabular}{llcccccccc}
\toprule
\textbf{Output} & \textbf{Metric} & \textbf{SAJA} & \textbf{T-MES} & \textbf{RELISH} & \textbf{MT-ICL} & \textbf{GenRe2} & \textbf{UniSO-N} & \textbf{GAT-AES} & \textbf{\method{} (Ours)} \\
\midrule
Contribution & NRMSE $\downarrow$ & \result{1.005}{0.057} & \result{1.021}{0.049} & \result{1.094}{0.056} & \result{1.052}{0.021} & \result{1.001}{0.031} & \result{1.049}{0.048} & \result{1.007}{0.057} & \cellcolor{dvgreen}\best{0.986}{0.043} \\
 & $r\uparrow$ & \result{0.033}{0.005} & \result{0.095}{0.084} & \result{0.066}{0.055} & \result{0.057}{0.071} & \result{0.149}{0.093} & \result{0.087}{0.047} & \result{0.118}{0.098} & \cellcolor{dvgreen}\best{0.194}{0.075} \\
 & $\rho\uparrow$ & \result{0.038}{0.034} & \result{0.100}{0.086} & \result{0.059}{0.055} & \result{0.058}{0.071} & \result{0.134}{0.064} & \result{0.088}{0.035} & \result{0.138}{0.092} & \cellcolor{dvgreen}\best{0.188}{0.075} \\
\addlinespace[3pt]
Soundness & NRMSE $\downarrow$ & \result{0.981}{0.052} & \result{1.005}{0.037} & \result{1.057}{0.045} & \result{1.007}{0.016} & \result{0.996}{0.031} & \result{1.026}{0.075} & \result{0.974}{0.055} & \cellcolor{dvgreen}\best{0.961}{0.035} \\
 & $r\uparrow$ & \result{0.049}{0.045} & \result{0.080}{0.001} & \result{0.048}{0.046} & \result{0.109}{0.081} & \result{0.090}{0.072} & \result{0.090}{0.076} & \result{0.095}{0.101} & \cellcolor{dvgreen}\best{0.201}{0.048} \\
 & $\rho\uparrow$ & \result{0.044}{0.023} & \result{0.103}{0.018} & \result{0.049}{0.062} & \result{0.103}{0.057} & \result{0.085}{0.096} & \result{0.096}{0.077} & \result{0.128}{0.069} & \cellcolor{dvgreen}\best{0.193}{0.055} \\
\addlinespace[3pt]
Presentation & NRMSE $\downarrow$ & \result{0.953}{0.031} & \result{0.973}{0.015} & \result{1.037}{0.020} & \result{0.980}{0.032} & \result{0.979}{0.030} & \result{1.011}{0.043} & \result{0.950}{0.032} & \cellcolor{dvgreen}\best{0.934}{0.026} \\
\bottomrule\end{tabular}}
\caption{CreativeEval, Qwen 9B output-level metrics. Green highlights a winning \method{} cell and yellow highlights a winning baseline cell.}
\label{tab:targets-creativeeval-qwen-9b}
\end{table*}
\begin{table*}[p]\centering\small\setlength{\tabcolsep}{3pt}
\resizebox{\textwidth}{!}{\begin{tabular}{llcccccccc}
\toprule
\textbf{Output} & \textbf{Metric} & \textbf{SAJA} & \textbf{T-MES} & \textbf{RELISH} & \textbf{MT-ICL} & \textbf{GenRe2} & \textbf{UniSO-N} & \textbf{GAT-AES} & \textbf{\method{} (Ours)} \\
\midrule
Contribution & NRMSE $\downarrow$ & \result{1.004}{0.055} & \result{1.009}{0.047} & \result{1.075}{0.084} & \result{1.025}{0.069} & \result{1.037}{0.042} & \result{1.094}{0.044} & \result{1.004}{0.050} & \cellcolor{dvgreen}\best{0.993}{0.029} \\
 & $r\uparrow$ & \result{0.044}{0.033} & \result{0.154}{0.055} & \result{0.070}{0.051} & \result{0.126}{0.064} & \result{0.042}{0.035} & \result{0.024}{0.006} & \result{0.137}{0.116} & \cellcolor{dvgreen}\best{0.196}{0.092} \\
 & $\rho\uparrow$ & \result{0.048}{0.032} & \result{0.136}{0.062} & \result{0.073}{0.040} & \result{0.109}{0.078} & \result{0.064}{0.049} & \result{0.030}{0.008} & \result{0.141}{0.119} & \cellcolor{dvgreen}\best{0.176}{0.119} \\
\addlinespace[3pt]
Soundness & NRMSE $\downarrow$ & \result{0.981}{0.046} & \result{1.023}{0.042} & \result{1.050}{0.063} & \result{0.988}{0.062} & \result{1.010}{0.045} & \result{1.063}{0.066} & \result{0.972}{0.038} & \cellcolor{dvgreen}\best{0.961}{0.045} \\
 & $r\uparrow$ & \result{0.048}{0.028} & \result{0.093}{0.077} & \result{0.076}{0.054} & \result{0.143}{0.069} & \result{0.078}{0.047} & \result{0.041}{0.051} & \result{0.153}{0.051} & \cellcolor{dvgreen}\best{0.204}{0.056} \\
 & $\rho\uparrow$ & \result{0.063}{0.028} & \result{0.083}{0.068} & \result{0.070}{0.059} & \result{0.122}{0.048} & \result{0.056}{0.022} & \result{0.046}{0.030} & \result{0.152}{0.044} & \cellcolor{dvgreen}\best{0.199}{0.057} \\
\addlinespace[3pt]
Presentation & NRMSE $\downarrow$ & \result{0.953}{0.039} & \result{0.948}{0.018} & \result{1.018}{0.037} & \result{0.969}{0.062} & \result{0.967}{0.038} & \result{1.023}{0.037} & \result{0.945}{0.028} & \cellcolor{dvgreen}\best{0.929}{0.029} \\
\bottomrule\end{tabular}}
\caption{CreativeEval, Qwen 27B output-level metrics. Green highlights a winning \method{} cell and yellow highlights a winning baseline cell.}
\label{tab:targets-creativeeval-qwen-27b}
\end{table*}
\begin{table*}[p]\centering\small\setlength{\tabcolsep}{3pt}
\resizebox{\textwidth}{!}{\begin{tabular}{llcccccccc}
\toprule
\textbf{Output} & \textbf{Metric} & \textbf{SAJA} & \textbf{T-MES} & \textbf{RELISH} & \textbf{MT-ICL} & \textbf{GenRe2} & \textbf{UniSO-N} & \textbf{GAT-AES} & \textbf{\method{} (Ours)} \\
\midrule
Contribution & NRMSE $\downarrow$ & \result{1.008}{0.055} & \result{1.004}{0.045} & \result{1.061}{0.082} & \result{1.000}{0.057} & \result{0.991}{0.044} & \result{1.052}{0.101} & \result{0.996}{0.058} & \cellcolor{dvgreen}\best{0.979}{0.051} \\
 & $r\uparrow$ & \result{0.070}{0.086} & \result{0.110}{0.093} & \result{0.120}{0.037} & \result{0.144}{0.057} & \result{0.165}{0.057} & \result{0.143}{0.044} & \result{0.190}{0.058} & \cellcolor{dvgreen}\best{0.225}{0.086} \\
 & $\rho\uparrow$ & \result{0.090}{0.083} & \result{0.108}{0.089} & \result{0.118}{0.042} & \result{0.116}{0.070} & \result{0.135}{0.082} & \result{0.142}{0.063} & \result{0.190}{0.071} & \cellcolor{dvgreen}\best{0.220}{0.104} \\
\addlinespace[3pt]
Soundness & NRMSE $\downarrow$ & \result{0.987}{0.046} & \result{1.013}{0.061} & \result{1.027}{0.032} & \result{0.978}{0.060} & \result{0.981}{0.048} & \result{1.040}{0.127} & \result{0.981}{0.048} & \cellcolor{dvgreen}\best{0.965}{0.051} \\
 & $r\uparrow$ & \result{0.094}{0.014} & \result{0.128}{0.108} & \result{0.138}{0.047} & \result{0.117}{0.048} & \result{0.111}{0.029} & \result{0.085}{0.046} & \result{0.062}{0.045} & \cellcolor{dvgreen}\best{0.181}{0.028} \\
 & $\rho\uparrow$ & \result{0.083}{0.027} & \result{0.135}{0.115} & \result{0.145}{0.043} & \result{0.117}{0.058} & \result{0.123}{0.021} & \result{0.089}{0.062} & \result{0.056}{0.054} & \cellcolor{dvgreen}\best{0.181}{0.026} \\
\addlinespace[3pt]
Presentation & NRMSE $\downarrow$ & \result{0.952}{0.034} & \result{0.995}{0.003} & \result{1.001}{0.043} & \result{0.952}{0.044} & \result{0.964}{0.046} & \result{0.988}{0.093} & \result{0.949}{0.032} & \cellcolor{dvgreen}\best{0.942}{0.032} \\
\bottomrule\end{tabular}}
\caption{CreativeEval, Gemma 31B output-level metrics. Green highlights a winning \method{} cell and yellow highlights a winning baseline cell.}
\label{tab:targets-creativeeval-gemma-31b}
\end{table*}
\begin{table*}[p]\centering\small\setlength{\tabcolsep}{3pt}
\resizebox{\textwidth}{!}{\begin{tabular}{llcccccccc}
\toprule
\textbf{Output} & \textbf{Metric} & \textbf{SAJA} & \textbf{T-MES} & \textbf{RELISH} & \textbf{MT-ICL} & \textbf{GenRe2} & \textbf{UniSO-N} & \textbf{GAT-AES} & \textbf{\method{} (Ours)} \\
\midrule
Contribution & NRMSE $\downarrow$ & \result{1.004}{0.056} & \result{1.036}{0.036} & \result{1.057}{0.055} & \result{1.007}{0.058} & \result{1.014}{0.039} & \result{1.031}{0.028} & \result{1.008}{0.054} & \cellcolor{dvgreen}\best{0.984}{0.039} \\
 & $r\uparrow$ & \result{0.059}{0.014} & \result{0.085}{0.053} & \result{0.080}{0.059} & \result{0.160}{0.018} & \result{0.068}{0.078} & \result{0.176}{0.038} & \result{0.076}{0.058} & \cellcolor{dvgreen}\best{0.213}{0.060} \\
 & $\rho\uparrow$ & \result{0.079}{0.032} & \result{0.090}{0.042} & \result{0.070}{0.054} & \result{0.131}{0.061} & \result{0.048}{0.071} & \result{0.173}{0.050} & \result{0.083}{0.059} & \cellcolor{dvgreen}\best{0.188}{0.063} \\
\addlinespace[3pt]
Soundness & NRMSE $\downarrow$ & \result{0.979}{0.056} & \result{1.003}{0.043} & \result{1.032}{0.032} & \result{0.980}{0.045} & \result{0.982}{0.049} & \result{0.995}{0.043} & \result{0.985}{0.046} & \cellcolor{dvgreen}\best{0.961}{0.049} \\
 & $r\uparrow$ & \result{0.083}{0.069} & \result{0.096}{0.092} & \result{0.089}{0.072} & \result{0.128}{0.083} & \result{0.123}{0.035} & \result{0.139}{0.014} & \result{0.045}{0.013} & \cellcolor{dvgreen}\best{0.198}{0.030} \\
 & $\rho\uparrow$ & \result{0.084}{0.066} & \result{0.093}{0.103} & \result{0.101}{0.089} & \result{0.120}{0.069} & \result{0.084}{0.030} & \result{0.134}{0.024} & \result{0.042}{0.014} & \cellcolor{dvgreen}\best{0.197}{0.028} \\
\addlinespace[3pt]
Presentation & NRMSE $\downarrow$ & \result{0.950}{0.024} & \result{0.959}{0.057} & \result{1.008}{0.024} & \result{0.959}{0.034} & \result{0.964}{0.054} & \result{0.959}{0.037} & \result{0.953}{0.035} & \cellcolor{dvgreen}\best{0.935}{0.032} \\
\bottomrule\end{tabular}}
\caption{CreativeEval, Llama 70B output-level metrics. Green highlights a winning \method{} cell and yellow highlights a winning baseline cell.}
\label{tab:targets-creativeeval-llama-70b}
\end{table*}
\begin{table*}[p]\centering\small\setlength{\tabcolsep}{3pt}
\resizebox{\textwidth}{!}{\begin{tabular}{llcccccccc}
\toprule
\textbf{Output} & \textbf{Metric} & \textbf{SAJA} & \textbf{T-MES} & \textbf{RELISH} & \textbf{MT-ICL} & \textbf{GenRe2} & \textbf{UniSO-N} & \textbf{GAT-AES} & \textbf{\method{} (Ours)} \\
\midrule
Contribution & NRMSE $\downarrow$ & \result{1.003}{0.054} & \result{1.041}{0.020} & \result{1.048}{0.032} & \result{1.004}{0.046} & \result{1.026}{0.049} & \result{1.086}{0.089} & \result{1.012}{0.056} & \cellcolor{dvgreen}\best{0.983}{0.043} \\
 & $r\uparrow$ & \result{0.032}{0.035} & \result{0.074}{0.038} & \result{0.059}{0.047} & \result{0.115}{0.074} & \result{0.077}{0.099} & \result{0.078}{0.091} & \result{0.103}{0.062} & \cellcolor{dvgreen}\best{0.212}{0.051} \\
 & $\rho\uparrow$ & \result{0.042}{0.020} & \result{0.076}{0.066} & \result{0.067}{0.018} & \result{0.095}{0.084} & \result{0.074}{0.091} & \result{0.092}{0.110} & \result{0.112}{0.034} & \cellcolor{dvgreen}\best{0.212}{0.066} \\
\addlinespace[3pt]
Soundness & NRMSE $\downarrow$ & \result{0.982}{0.050} & \result{1.062}{0.028} & \result{1.007}{0.046} & \result{0.979}{0.045} & \result{0.990}{0.050} & \result{1.017}{0.035} & \result{0.983}{0.045} & \cellcolor{dvgreen}\best{0.960}{0.041} \\
 & $r\uparrow$ & \result{0.017}{0.017} & \result{0.114}{0.044} & \result{0.047}{0.045} & \result{0.113}{0.042} & \result{0.113}{0.093} & \result{0.113}{0.062} & \result{0.121}{0.038} & \cellcolor{dvgreen}\best{0.200}{0.044} \\
 & $\rho\uparrow$ & \result{0.019}{0.018} & \result{0.107}{0.059} & \result{0.065}{0.031} & \result{0.127}{0.049} & \result{0.125}{0.044} & \result{0.114}{0.064} & \result{0.132}{0.026} & \cellcolor{dvgreen}\best{0.221}{0.062} \\
\addlinespace[3pt]
Presentation & NRMSE $\downarrow$ & \result{0.951}{0.035} & \result{0.958}{0.053} & \result{1.019}{0.033} & \result{0.945}{0.029} & \result{0.963}{0.039} & \result{0.985}{0.023} & \result{0.953}{0.034} & \cellcolor{dvgreen}\best{0.932}{0.030} \\
\bottomrule\end{tabular}}
\caption{CreativeEval, Qwen 72B output-level metrics. Green highlights a winning \method{} cell and yellow highlights a winning baseline cell.}
\label{tab:targets-creativeeval-qwen-72b}
\end{table*}
\begin{table*}[p]\centering\small\setlength{\tabcolsep}{3pt}
\resizebox{\textwidth}{!}{\begin{tabular}{llcccccccc}
\toprule
\textbf{Output} & \textbf{Metric} & \textbf{SAJA} & \textbf{T-MES} & \textbf{RELISH} & \textbf{MT-ICL} & \textbf{GenRe2} & \textbf{UniSO-N} & \textbf{GAT-AES} & \textbf{\method{} (Ours)} \\
\midrule
Contribution & NRMSE $\downarrow$ & \result{0.999}{0.054} & \result{1.110}{0.026} & \result{1.066}{0.029} & \result{1.026}{0.069} & \result{1.002}{0.054} & \result{1.029}{0.030} & \result{1.010}{0.058} & \cellcolor{dvgreen}\best{0.997}{0.026} \\
 & $r\uparrow$ & \result{0.087}{0.012} & \result{0.064}{0.050} & \result{0.087}{0.035} & \result{0.055}{0.016} & \result{0.070}{0.005} & \result{0.101}{0.086} & \result{0.035}{0.031} & \cellcolor{dvgreen}\best{0.175}{0.090} \\
 & $\rho\uparrow$ & \result{0.075}{0.029} & \result{0.090}{0.038} & \result{0.093}{0.035} & \result{0.048}{0.020} & \result{0.094}{0.041} & \result{0.109}{0.085} & \result{0.048}{0.019} & \cellcolor{dvgreen}\best{0.178}{0.118} \\
\addlinespace[3pt]
Soundness & NRMSE $\downarrow$ & \cellcolor{dvyellow}\best{0.978}{0.044} & \result{1.263}{0.100} & \result{1.053}{0.049} & \result{1.041}{0.050} & \result{0.981}{0.055} & \result{1.022}{0.058} & \result{0.988}{0.047} & \result{0.985}{0.053} \\
 & $r\uparrow$ & \result{0.066}{0.050} & \result{0.088}{0.108} & \result{0.053}{0.028} & \cellcolor{dvyellow}\best{0.089}{0.065} & \result{0.074}{0.064} & \result{0.052}{0.004} & \result{0.049}{0.027} & \result{0.080}{0.069} \\
 & $\rho\uparrow$ & \result{0.056}{0.041} & \result{0.088}{0.111} & \result{0.054}{0.021} & \cellcolor{dvyellow}\best{0.100}{0.074} & \result{0.083}{0.065} & \result{0.064}{0.012} & \result{0.052}{0.046} & \result{0.056}{0.070} \\
\addlinespace[3pt]
Presentation & NRMSE $\downarrow$ & \cellcolor{dvyellow}\best{0.950}{0.030} & \result{1.029}{0.077} & \result{0.997}{0.085} & \result{0.968}{0.046} & \result{0.962}{0.035} & \result{0.982}{0.056} & \result{0.963}{0.035} & \result{0.952}{0.022} \\
\bottomrule\end{tabular}}
\caption{CreativeEval, GPT 120B output-level metrics. Green highlights a winning \method{} cell and yellow highlights a winning baseline cell.}
\label{tab:targets-creativeeval-gpt-120b}
\end{table*}
\begin{table*}[p]\centering\small\setlength{\tabcolsep}{3pt}
\resizebox{\textwidth}{!}{\begin{tabular}{llcccccccc}
\toprule
\textbf{Output} & \textbf{Metric} & \textbf{SAJA} & \textbf{T-MES} & \textbf{RELISH} & \textbf{MT-ICL} & \textbf{GenRe2} & \textbf{UniSO-N} & \textbf{GAT-AES} & \textbf{\method{} (Ours)} \\
\midrule
Contribution & NRMSE $\downarrow$ & \result{1.007}{0.054} & \result{1.032}{0.026} & \result{1.078}{0.033} & \result{0.993}{0.062} & \result{1.026}{0.056} & \result{1.021}{0.114} & \result{1.011}{0.055} & \cellcolor{dvgreen}\best{0.976}{0.051} \\
 & $r\uparrow$ & \result{0.061}{0.049} & \result{0.137}{0.050} & \result{0.098}{0.014} & \result{0.151}{0.040} & \result{0.097}{0.044} & \result{0.161}{0.079} & \result{0.057}{0.010} & \cellcolor{dvgreen}\best{0.234}{0.041} \\
 & $\rho\uparrow$ & \result{0.061}{0.047} & \result{0.138}{0.034} & \result{0.091}{0.027} & \result{0.140}{0.070} & \result{0.115}{0.034} & \result{0.136}{0.116} & \result{0.044}{0.005} & \cellcolor{dvgreen}\best{0.232}{0.070} \\
\addlinespace[3pt]
Soundness & NRMSE $\downarrow$ & \result{0.981}{0.053} & \result{0.999}{0.030} & \result{1.102}{0.017} & \result{0.983}{0.067} & \result{0.995}{0.045} & \result{0.998}{0.059} & \result{0.978}{0.051} & \cellcolor{dvgreen}\best{0.965}{0.049} \\
 & $r\uparrow$ & \result{0.041}{0.020} & \result{0.128}{0.097} & \result{0.103}{0.067} & \result{0.098}{0.117} & \result{0.071}{0.060} & \result{0.088}{0.068} & \result{0.092}{0.054} & \cellcolor{dvgreen}\best{0.172}{0.055} \\
 & $\rho\uparrow$ & \result{0.068}{0.034} & \result{0.110}{0.116} & \result{0.086}{0.077} & \result{0.098}{0.077} & \result{0.053}{0.054} & \result{0.065}{0.061} & \result{0.109}{0.054} & \cellcolor{dvgreen}\best{0.158}{0.051} \\
\addlinespace[3pt]
Presentation & NRMSE $\downarrow$ & \result{0.951}{0.037} & \result{0.972}{0.006} & \result{1.033}{0.018} & \result{0.946}{0.045} & \result{0.968}{0.027} & \result{0.998}{0.080} & \result{0.964}{0.035} & \cellcolor{dvgreen}\best{0.929}{0.031} \\
\bottomrule\end{tabular}}
\caption{CreativeEval, Qwen 122B output-level metrics. Green highlights a winning \method{} cell and yellow highlights a winning baseline cell.}
\label{tab:targets-creativeeval-qwen-122b}
\end{table*}
\begin{table*}[p]\centering\small\setlength{\tabcolsep}{3pt}
\resizebox{\textwidth}{!}{\begin{tabular}{llcccccccc}
\toprule
\textbf{Output} & \textbf{Metric} & \textbf{SAJA} & \textbf{T-MES} & \textbf{RELISH} & \textbf{MT-ICL} & \textbf{GenRe2} & \textbf{UniSO-N} & \textbf{GAT-AES} & \textbf{\method{} (Ours)} \\
\midrule
Completeness & NRMSE $\downarrow$ & \result{0.681}{0.055} & \result{0.872}{0.075} & \result{0.701}{0.009} & \result{0.645}{0.033} & \result{0.599}{0.038} & \result{0.596}{0.031} & \result{0.590}{0.062} & \cellcolor{dvgreen}\best{0.577}{0.051} \\
 & $r\uparrow$ & \result{0.682}{0.065} & \result{0.454}{0.137} & \result{0.706}{0.071} & \result{0.723}{0.054} & \result{0.770}{0.047} & \result{0.771}{0.046} & \cellcolor{dvyellow}\best{0.789}{0.065} & \result{0.783}{0.058} \\
 & $\rho\uparrow$ & \result{0.636}{0.056} & \result{0.430}{0.111} & \result{0.671}{0.070} & \result{0.699}{0.042} & \result{0.746}{0.035} & \result{0.759}{0.037} & \cellcolor{dvyellow}\best{0.765}{0.059} & \result{0.759}{0.052} \\
\addlinespace[3pt]
Conciseness & NRMSE $\downarrow$ & \result{0.824}{0.148} & \result{0.967}{0.102} & \result{0.848}{0.155} & \result{0.801}{0.100} & \result{0.783}{0.125} & \result{0.777}{0.074} & \result{0.787}{0.112} & \cellcolor{dvgreen}\best{0.742}{0.098} \\
 & $r\uparrow$ & \result{0.614}{0.088} & \result{0.356}{0.081} & \result{0.609}{0.048} & \result{0.643}{0.035} & \result{0.673}{0.055} & \result{0.677}{0.005} & \result{0.664}{0.053} & \cellcolor{dvgreen}\best{0.707}{0.032} \\
 & $\rho\uparrow$ & \result{0.335}{0.017} & \result{0.318}{0.089} & \result{0.417}{0.067} & \result{0.476}{0.032} & \result{0.520}{0.032} & \result{0.549}{0.049} & \result{0.508}{0.013} & \cellcolor{dvgreen}\best{0.568}{0.038} \\
\addlinespace[3pt]
Faithfulness & NRMSE $\downarrow$ & \result{0.766}{0.038} & \result{0.971}{0.044} & \result{0.805}{0.040} & \result{0.799}{0.019} & \result{0.779}{0.044} & \result{0.788}{0.039} & \result{0.768}{0.032} & \cellcolor{dvgreen}\best{0.764}{0.026} \\
\bottomrule\end{tabular}}
\caption{MSumBench, Gemma E4B output-level metrics. Green highlights a winning \method{} cell and yellow highlights a winning baseline cell.}
\label{tab:targets-msumbench-gemma-e4b}
\end{table*}
\begin{table*}[p]\centering\small\setlength{\tabcolsep}{3pt}
\resizebox{\textwidth}{!}{\begin{tabular}{llcccccccc}
\toprule
\textbf{Output} & \textbf{Metric} & \textbf{SAJA} & \textbf{T-MES} & \textbf{RELISH} & \textbf{MT-ICL} & \textbf{GenRe2} & \textbf{UniSO-N} & \textbf{GAT-AES} & \textbf{\method{} (Ours)} \\
\midrule
Completeness & NRMSE $\downarrow$ & \result{0.693}{0.066} & \result{0.844}{0.080} & \result{0.753}{0.046} & \result{0.631}{0.058} & \result{0.594}{0.052} & \result{0.648}{0.059} & \result{0.608}{0.050} & \cellcolor{dvgreen}\best{0.582}{0.052} \\
 & $r\uparrow$ & \result{0.668}{0.091} & \result{0.469}{0.147} & \result{0.682}{0.043} & \result{0.738}{0.069} & \result{0.772}{0.063} & \result{0.735}{0.083} & \result{0.777}{0.079} & \cellcolor{dvgreen}\best{0.779}{0.065} \\
 & $\rho\uparrow$ & \result{0.607}{0.084} & \result{0.464}{0.121} & \result{0.650}{0.062} & \result{0.726}{0.072} & \result{0.759}{0.050} & \result{0.717}{0.070} & \cellcolor{dvyellow}\best{0.763}{0.070} & \result{0.761}{0.059} \\
\addlinespace[3pt]
Conciseness & NRMSE $\downarrow$ & \result{0.848}{0.149} & \result{0.964}{0.101} & \result{0.887}{0.085} & \result{0.792}{0.110} & \result{0.783}{0.096} & \result{0.785}{0.112} & \result{0.793}{0.085} & \cellcolor{dvgreen}\best{0.738}{0.098} \\
 & $r\uparrow$ & \result{0.587}{0.098} & \result{0.361}{0.096} & \result{0.598}{0.050} & \result{0.642}{0.054} & \result{0.676}{0.033} & \result{0.675}{0.055} & \result{0.670}{0.028} & \cellcolor{dvgreen}\best{0.702}{0.036} \\
 & $\rho\uparrow$ & \result{0.267}{0.044} & \result{0.292}{0.082} & \result{0.431}{0.014} & \result{0.441}{0.040} & \result{0.533}{0.060} & \result{0.524}{0.013} & \result{0.539}{0.041} & \cellcolor{dvgreen}\best{0.551}{0.019} \\
\addlinespace[3pt]
Faithfulness & NRMSE $\downarrow$ & \result{0.765}{0.058} & \result{0.923}{0.049} & \result{0.822}{0.040} & \result{0.782}{0.013} & \result{0.777}{0.050} & \result{0.816}{0.064} & \cellcolor{dvyellow}\best{0.757}{0.039} & \result{0.757}{0.030} \\
\bottomrule\end{tabular}}
\caption{MSumBench, Qwen 9B output-level metrics. Green highlights a winning \method{} cell and yellow highlights a winning baseline cell.}
\label{tab:targets-msumbench-qwen-9b}
\end{table*}
\begin{table*}[p]\centering\small\setlength{\tabcolsep}{3pt}
\resizebox{\textwidth}{!}{\begin{tabular}{llcccccccc}
\toprule
\textbf{Output} & \textbf{Metric} & \textbf{SAJA} & \textbf{T-MES} & \textbf{RELISH} & \textbf{MT-ICL} & \textbf{GenRe2} & \textbf{UniSO-N} & \textbf{GAT-AES} & \textbf{\method{} (Ours)} \\
\midrule
Completeness & NRMSE $\downarrow$ & \result{0.726}{0.044} & \result{0.830}{0.067} & \result{0.688}{0.062} & \result{0.587}{0.068} & \result{0.587}{0.053} & \result{0.617}{0.081} & \result{0.567}{0.053} & \cellcolor{dvgreen}\best{0.558}{0.068} \\
 & $r\uparrow$ & \result{0.637}{0.066} & \result{0.491}{0.122} & \result{0.723}{0.055} & \result{0.776}{0.071} & \result{0.776}{0.063} & \result{0.753}{0.094} & \cellcolor{dvyellow}\best{0.799}{0.063} & \result{0.798}{0.069} \\
 & $\rho\uparrow$ & \result{0.587}{0.075} & \result{0.492}{0.108} & \result{0.701}{0.053} & \result{0.763}{0.071} & \result{0.764}{0.052} & \result{0.737}{0.088} & \cellcolor{dvyellow}\best{0.783}{0.054} & \result{0.783}{0.062} \\
\addlinespace[3pt]
Conciseness & NRMSE $\downarrow$ & \result{0.832}{0.143} & \result{0.972}{0.085} & \result{0.810}{0.084} & \result{0.783}{0.116} & \result{0.763}{0.123} & \result{0.763}{0.120} & \result{0.755}{0.096} & \cellcolor{dvgreen}\best{0.728}{0.091} \\
 & $r\uparrow$ & \result{0.606}{0.083} & \result{0.346}{0.017} & \result{0.637}{0.036} & \result{0.666}{0.053} & \result{0.689}{0.054} & \result{0.677}{0.066} & \result{0.688}{0.047} & \cellcolor{dvgreen}\best{0.710}{0.032} \\
 & $\rho\uparrow$ & \result{0.341}{0.010} & \result{0.320}{0.061} & \result{0.454}{0.010} & \result{0.492}{0.029} & \result{0.558}{0.016} & \result{0.536}{0.048} & \result{0.554}{0.021} & \cellcolor{dvgreen}\best{0.581}{0.041} \\
\addlinespace[3pt]
Faithfulness & NRMSE $\downarrow$ & \result{0.739}{0.042} & \result{0.943}{0.047} & \result{0.788}{0.054} & \result{0.761}{0.010} & \result{0.771}{0.041} & \result{0.804}{0.004} & \result{0.735}{0.015} & \cellcolor{dvgreen}\best{0.733}{0.021} \\
\bottomrule\end{tabular}}
\caption{MSumBench, Qwen 27B output-level metrics. Green highlights a winning \method{} cell and yellow highlights a winning baseline cell.}
\label{tab:targets-msumbench-qwen-27b}
\end{table*}
\begin{table*}[p]\centering\small\setlength{\tabcolsep}{3pt}
\resizebox{\textwidth}{!}{\begin{tabular}{llcccccccc}
\toprule
\textbf{Output} & \textbf{Metric} & \textbf{SAJA} & \textbf{T-MES} & \textbf{RELISH} & \textbf{MT-ICL} & \textbf{GenRe2} & \textbf{UniSO-N} & \textbf{GAT-AES} & \textbf{\method{} (Ours)} \\
\midrule
Completeness & NRMSE $\downarrow$ & \result{0.728}{0.033} & \result{0.769}{0.067} & \result{0.775}{0.084} & \result{0.606}{0.060} & \result{0.595}{0.039} & \result{0.607}{0.077} & \result{0.579}{0.048} & \cellcolor{dvgreen}\best{0.566}{0.052} \\
 & $r\uparrow$ & \result{0.637}{0.040} & \result{0.592}{0.079} & \result{0.645}{0.046} & \result{0.763}{0.065} & \result{0.772}{0.054} & \result{0.769}{0.075} & \result{0.783}{0.056} & \cellcolor{dvgreen}\best{0.793}{0.057} \\
 & $\rho\uparrow$ & \result{0.596}{0.029} & \result{0.592}{0.065} & \result{0.626}{0.064} & \result{0.750}{0.070} & \result{0.754}{0.047} & \result{0.752}{0.067} & \result{0.765}{0.058} & \cellcolor{dvgreen}\best{0.780}{0.055} \\
\addlinespace[3pt]
Conciseness & NRMSE $\downarrow$ & \result{0.838}{0.152} & \result{0.915}{0.050} & \result{0.963}{0.178} & \result{0.792}{0.103} & \result{0.789}{0.118} & \result{0.807}{0.094} & \result{0.793}{0.122} & \cellcolor{dvgreen}\best{0.741}{0.097} \\
 & $r\uparrow$ & \result{0.597}{0.094} & \result{0.460}{0.038} & \result{0.478}{0.061} & \result{0.667}{0.040} & \result{0.660}{0.053} & \result{0.641}{0.043} & \result{0.661}{0.062} & \cellcolor{dvgreen}\best{0.704}{0.034} \\
 & $\rho\uparrow$ & \result{0.258}{0.094} & \result{0.387}{0.058} & \result{0.279}{0.062} & \result{0.516}{0.024} & \result{0.512}{0.033} & \result{0.507}{0.029} & \result{0.474}{0.021} & \cellcolor{dvgreen}\best{0.568}{0.017} \\
\addlinespace[3pt]
Faithfulness & NRMSE $\downarrow$ & \result{0.743}{0.047} & \result{0.899}{0.017} & \result{0.890}{0.061} & \result{0.769}{0.028} & \result{0.767}{0.052} & \result{0.764}{0.037} & \result{0.745}{0.039} & \cellcolor{dvgreen}\best{0.739}{0.033} \\
\bottomrule\end{tabular}}
\caption{MSumBench, Gemma 31B output-level metrics. Green highlights a winning \method{} cell and yellow highlights a winning baseline cell.}
\label{tab:targets-msumbench-gemma-31b}
\end{table*}
\begin{table*}[p]\centering\small\setlength{\tabcolsep}{3pt}
\resizebox{\textwidth}{!}{\begin{tabular}{llcccccccc}
\toprule
\textbf{Output} & \textbf{Metric} & \textbf{SAJA} & \textbf{T-MES} & \textbf{RELISH} & \textbf{MT-ICL} & \textbf{GenRe2} & \textbf{UniSO-N} & \textbf{GAT-AES} & \textbf{\method{} (Ours)} \\
\midrule
Completeness & NRMSE $\downarrow$ & \result{0.690}{0.071} & \result{0.860}{0.056} & \result{0.665}{0.041} & \result{0.614}{0.041} & \cellcolor{dvyellow}\best{0.575}{0.070} & \result{0.587}{0.058} & \result{0.615}{0.053} & \result{0.577}{0.049} \\
 & $r\uparrow$ & \result{0.672}{0.094} & \result{0.429}{0.149} & \result{0.711}{0.061} & \result{0.765}{0.072} & \result{0.782}{0.076} & \result{0.779}{0.062} & \result{0.770}{0.072} & \cellcolor{dvgreen}\best{0.789}{0.062} \\
 & $\rho\uparrow$ & \result{0.620}{0.103} & \result{0.410}{0.125} & \result{0.675}{0.072} & \result{0.736}{0.078} & \result{0.760}{0.067} & \result{0.756}{0.055} & \result{0.732}{0.071} & \cellcolor{dvgreen}\best{0.769}{0.059} \\
\addlinespace[3pt]
Conciseness & NRMSE $\downarrow$ & \result{0.830}{0.150} & \result{0.992}{0.074} & \result{0.904}{0.132} & \result{0.808}{0.113} & \result{0.794}{0.119} & \result{0.804}{0.113} & \result{0.801}{0.117} & \cellcolor{dvgreen}\best{0.758}{0.128} \\
 & $r\uparrow$ & \result{0.608}{0.089} & \result{0.318}{0.021} & \result{0.571}{0.020} & \result{0.647}{0.071} & \result{0.668}{0.059} & \result{0.641}{0.052} & \result{0.656}{0.052} & \cellcolor{dvgreen}\best{0.693}{0.060} \\
 & $\rho\uparrow$ & \result{0.335}{0.028} & \result{0.302}{0.053} & \result{0.406}{0.044} & \result{0.451}{0.071} & \result{0.487}{0.037} & \result{0.468}{0.020} & \result{0.490}{0.018} & \cellcolor{dvgreen}\best{0.552}{0.018} \\
\addlinespace[3pt]
Faithfulness & NRMSE $\downarrow$ & \cellcolor{dvyellow}\best{0.761}{0.041} & \result{0.965}{0.046} & \result{0.836}{0.035} & \result{0.789}{0.036} & \result{0.787}{0.046} & \result{0.800}{0.032} & \result{0.762}{0.037} & \result{0.769}{0.043} \\
\bottomrule\end{tabular}}
\caption{MSumBench, Llama 70B output-level metrics. Green highlights a winning \method{} cell and yellow highlights a winning baseline cell.}
\label{tab:targets-msumbench-llama-70b}
\end{table*}
\begin{table*}[p]\centering\small\setlength{\tabcolsep}{3pt}
\resizebox{\textwidth}{!}{\begin{tabular}{llcccccccc}
\toprule
\textbf{Output} & \textbf{Metric} & \textbf{SAJA} & \textbf{T-MES} & \textbf{RELISH} & \textbf{MT-ICL} & \textbf{GenRe2} & \textbf{UniSO-N} & \textbf{GAT-AES} & \textbf{\method{} (Ours)} \\
\midrule
Completeness & NRMSE $\downarrow$ & \result{0.651}{0.064} & \result{0.873}{0.044} & \result{0.753}{0.161} & \result{0.568}{0.065} & \result{0.558}{0.050} & \result{0.569}{0.044} & \result{0.552}{0.061} & \cellcolor{dvgreen}\best{0.544}{0.045} \\
 & $r\uparrow$ & \result{0.712}{0.088} & \result{0.474}{0.138} & \result{0.687}{0.116} & \result{0.801}{0.053} & \result{0.799}{0.057} & \result{0.801}{0.054} & \result{0.805}{0.065} & \cellcolor{dvgreen}\best{0.813}{0.050} \\
 & $\rho\uparrow$ & \result{0.683}{0.082} & \result{0.452}{0.120} & \result{0.645}{0.131} & \result{0.782}{0.044} & \result{0.783}{0.050} & \result{0.787}{0.049} & \result{0.788}{0.057} & \cellcolor{dvgreen}\best{0.800}{0.040} \\
\addlinespace[3pt]
Conciseness & NRMSE $\downarrow$ & \result{0.829}{0.150} & \result{0.989}{0.070} & \result{0.883}{0.174} & \result{0.759}{0.101} & \result{0.785}{0.100} & \result{0.806}{0.087} & \result{0.811}{0.138} & \cellcolor{dvgreen}\best{0.743}{0.081} \\
 & $r\uparrow$ & \result{0.609}{0.095} & \result{0.329}{0.046} & \result{0.605}{0.091} & \result{0.678}{0.043} & \result{0.674}{0.043} & \result{0.651}{0.037} & \result{0.660}{0.069} & \cellcolor{dvgreen}\best{0.700}{0.020} \\
 & $\rho\uparrow$ & \result{0.354}{0.063} & \result{0.290}{0.016} & \result{0.415}{0.044} & \result{0.511}{0.026} & \result{0.514}{0.043} & \result{0.496}{0.071} & \result{0.476}{0.024} & \cellcolor{dvgreen}\best{0.579}{0.029} \\
\addlinespace[3pt]
Faithfulness & NRMSE $\downarrow$ & \result{0.776}{0.066} & \result{0.956}{0.051} & \result{0.826}{0.095} & \result{0.756}{0.023} & \result{0.787}{0.049} & \result{0.769}{0.015} & \cellcolor{dvyellow}\best{0.739}{0.034} & \result{0.744}{0.029} \\
\bottomrule\end{tabular}}
\caption{MSumBench, Qwen 72B output-level metrics. Green highlights a winning \method{} cell and yellow highlights a winning baseline cell.}
\label{tab:targets-msumbench-qwen-72b}
\end{table*}
\begin{table*}[p]\centering\small\setlength{\tabcolsep}{3pt}
\resizebox{\textwidth}{!}{\begin{tabular}{llcccccccc}
\toprule
\textbf{Output} & \textbf{Metric} & \textbf{SAJA} & \textbf{T-MES} & \textbf{RELISH} & \textbf{MT-ICL} & \textbf{GenRe2} & \textbf{UniSO-N} & \textbf{GAT-AES} & \textbf{\method{} (Ours)} \\
\midrule
Completeness & NRMSE $\downarrow$ & \result{0.666}{0.039} & \result{0.749}{0.093} & \result{0.689}{0.029} & \result{0.634}{0.029} & \result{0.619}{0.041} & \result{0.627}{0.060} & \result{0.644}{0.075} & \cellcolor{dvgreen}\best{0.588}{0.055} \\
 & $r\uparrow$ & \result{0.700}{0.054} & \result{0.641}{0.121} & \result{0.705}{0.041} & \result{0.735}{0.052} & \result{0.746}{0.056} & \result{0.741}{0.074} & \result{0.752}{0.068} & \cellcolor{dvgreen}\best{0.774}{0.064} \\
 & $\rho\uparrow$ & \result{0.661}{0.073} & \result{0.645}{0.106} & \result{0.688}{0.041} & \result{0.709}{0.041} & \result{0.721}{0.044} & \result{0.716}{0.070} & \result{0.728}{0.064} & \cellcolor{dvgreen}\best{0.758}{0.062} \\
\addlinespace[3pt]
Conciseness & NRMSE $\downarrow$ & \result{0.819}{0.144} & \result{1.003}{0.217} & \result{0.852}{0.120} & \result{0.846}{0.109} & \result{0.805}{0.137} & \result{0.861}{0.143} & \result{0.831}{0.163} & \cellcolor{dvgreen}\best{0.780}{0.113} \\
 & $r\uparrow$ & \result{0.621}{0.087} & \result{0.365}{0.316} & \result{0.617}{0.029} & \result{0.587}{0.062} & \result{0.641}{0.077} & \result{0.571}{0.092} & \result{0.629}{0.070} & \cellcolor{dvgreen}\best{0.664}{0.049} \\
 & $\rho\uparrow$ & \result{0.369}{0.048} & \result{0.303}{0.263} & \result{0.456}{0.081} & \result{0.393}{0.011} & \result{0.440}{0.033} & \result{0.390}{0.047} & \result{0.438}{0.024} & \cellcolor{dvgreen}\best{0.504}{0.032} \\
\addlinespace[3pt]
Faithfulness & NRMSE $\downarrow$ & \cellcolor{dvyellow}\best{0.763}{0.049} & \result{0.986}{0.147} & \result{0.800}{0.033} & \result{0.799}{0.045} & \result{0.788}{0.043} & \result{0.838}{0.038} & \result{0.779}{0.036} & \result{0.788}{0.030} \\
\bottomrule\end{tabular}}
\caption{MSumBench, GPT 120B output-level metrics. Green highlights a winning \method{} cell and yellow highlights a winning baseline cell.}
\label{tab:targets-msumbench-gpt-120b}
\end{table*}
\begin{table*}[p]\centering\small\setlength{\tabcolsep}{3pt}
\resizebox{\textwidth}{!}{\begin{tabular}{llcccccccc}
\toprule
\textbf{Output} & \textbf{Metric} & \textbf{SAJA} & \textbf{T-MES} & \textbf{RELISH} & \textbf{MT-ICL} & \textbf{GenRe2} & \textbf{UniSO-N} & \textbf{GAT-AES} & \textbf{\method{} (Ours)} \\
\midrule
Completeness & NRMSE $\downarrow$ & \result{0.698}{0.053} & \result{0.855}{0.111} & \result{0.648}{0.039} & \result{0.605}{0.064} & \result{0.584}{0.062} & \result{0.597}{0.047} & \result{0.617}{0.080} & \cellcolor{dvgreen}\best{0.572}{0.056} \\
 & $r\uparrow$ & \result{0.668}{0.071} & \result{0.484}{0.170} & \result{0.745}{0.049} & \result{0.764}{0.072} & \result{0.783}{0.067} & \result{0.777}{0.069} & \result{0.777}{0.080} & \cellcolor{dvgreen}\best{0.789}{0.062} \\
 & $\rho\uparrow$ & \result{0.622}{0.069} & \result{0.466}{0.145} & \result{0.717}{0.036} & \result{0.742}{0.063} & \result{0.769}{0.059} & \result{0.766}{0.055} & \result{0.764}{0.070} & \cellcolor{dvgreen}\best{0.775}{0.051} \\
\addlinespace[3pt]
Conciseness & NRMSE $\downarrow$ & \result{0.835}{0.155} & \result{0.963}{0.075} & \result{0.825}{0.131} & \result{0.796}{0.120} & \result{0.796}{0.114} & \result{0.824}{0.085} & \result{0.771}{0.115} & \cellcolor{dvgreen}\best{0.751}{0.095} \\
 & $r\uparrow$ & \result{0.599}{0.099} & \result{0.364}{0.026} & \result{0.648}{0.029} & \result{0.639}{0.064} & \result{0.658}{0.056} & \result{0.663}{0.034} & \result{0.673}{0.064} & \cellcolor{dvgreen}\best{0.697}{0.033} \\
 & $\rho\uparrow$ & \result{0.239}{0.024} & \result{0.306}{0.031} & \result{0.452}{0.090} & \result{0.429}{0.035} & \result{0.480}{0.014} & \result{0.538}{0.020} & \result{0.512}{0.034} & \cellcolor{dvgreen}\best{0.566}{0.036} \\
\addlinespace[3pt]
Faithfulness & NRMSE $\downarrow$ & \result{0.751}{0.038} & \result{0.938}{0.033} & \result{0.802}{0.081} & \result{0.757}{0.018} & \result{0.758}{0.045} & \result{0.752}{0.033} & \result{0.731}{0.035} & \cellcolor{dvgreen}\best{0.725}{0.028} \\
\bottomrule\end{tabular}}
\caption{MSumBench, Qwen 122B output-level metrics. Green highlights a winning \method{} cell and yellow highlights a winning baseline cell.}
\label{tab:targets-msumbench-qwen-122b}
\end{table*}
\begin{table*}[p]\centering\small\setlength{\tabcolsep}{3pt}
\resizebox{\textwidth}{!}{\begin{tabular}{llcccccccc}
\toprule
\textbf{Output} & \textbf{Metric} & \textbf{SAJA} & \textbf{T-MES} & \textbf{RELISH} & \textbf{MT-ICL} & \textbf{GenRe2} & \textbf{UniSO-N} & \textbf{GAT-AES} & \textbf{\method{} (Ours)} \\
\midrule
Accuracy & NRMSE $\downarrow$ & \result{0.997}{0.026} & \result{0.997}{0.014} & \result{1.023}{0.015} & \result{0.979}{0.027} & \result{0.984}{0.010} & \result{1.022}{0.024} & \result{0.980}{0.028} & \cellcolor{dvgreen}\best{0.962}{0.006} \\
 & $r\uparrow$ & \result{0.190}{0.054} & \result{0.219}{0.042} & \result{0.236}{0.050} & \result{0.291}{0.064} & \result{0.275}{0.015} & \result{0.286}{0.020} & \result{0.307}{0.025} & \cellcolor{dvgreen}\best{0.322}{0.017} \\
 & $\rho\uparrow$ & \result{0.229}{0.052} & \result{0.278}{0.044} & \result{0.291}{0.034} & \result{0.318}{0.098} & \result{0.292}{0.027} & \result{0.331}{0.009} & \result{0.345}{0.054} & \cellcolor{dvgreen}\best{0.366}{0.048} \\
\addlinespace[3pt]
Fluency & NRMSE $\downarrow$ & \result{0.976}{0.042} & \result{0.943}{0.054} & \result{0.981}{0.018} & \result{0.931}{0.056} & \result{0.923}{0.048} & \result{1.007}{0.038} & \result{0.912}{0.021} & \cellcolor{dvgreen}\best{0.910}{0.028} \\
 & $r\uparrow$ & \result{0.238}{0.087} & \result{0.374}{0.043} & \result{0.363}{0.050} & \result{0.407}{0.022} & \result{0.400}{0.042} & \result{0.344}{0.041} & \result{0.421}{0.013} & \cellcolor{dvgreen}\best{0.437}{0.019} \\
 & $\rho\uparrow$ & \result{0.258}{0.102} & \result{0.453}{0.041} & \result{0.419}{0.021} & \result{0.461}{0.022} & \result{0.436}{0.038} & \result{0.415}{0.031} & \result{0.481}{0.014} & \cellcolor{dvgreen}\best{0.500}{0.004} \\
\addlinespace[3pt]
Style & NRMSE $\downarrow$ & \result{1.062}{0.050} & \result{1.052}{0.053} & \result{1.088}{0.076} & \result{1.050}{0.055} & \result{1.049}{0.065} & \result{1.096}{0.060} & \result{1.030}{0.059} & \cellcolor{dvgreen}\best{1.016}{0.056} \\
\bottomrule\end{tabular}}
\caption{RATE, Gemma E4B output-level metrics. Green highlights a winning \method{} cell and yellow highlights a winning baseline cell.}
\label{tab:targets-rate-gemma-e4b}
\end{table*}
\begin{table*}[p]\centering\small\setlength{\tabcolsep}{3pt}
\resizebox{\textwidth}{!}{\begin{tabular}{llcccccccc}
\toprule
\textbf{Output} & \textbf{Metric} & \textbf{SAJA} & \textbf{T-MES} & \textbf{RELISH} & \textbf{MT-ICL} & \textbf{GenRe2} & \textbf{UniSO-N} & \textbf{GAT-AES} & \textbf{\method{} (Ours)} \\
\midrule
Accuracy & NRMSE $\downarrow$ & \result{0.975}{0.003} & \result{0.988}{0.015} & \result{1.038}{0.013} & \result{1.029}{0.041} & \result{0.992}{0.023} & \result{1.031}{0.031} & \result{0.964}{0.025} & \cellcolor{dvgreen}\best{0.961}{0.010} \\
 & $r\uparrow$ & \result{0.264}{0.024} & \result{0.271}{0.044} & \result{0.240}{0.073} & \result{0.216}{0.063} & \result{0.280}{0.042} & \result{0.266}{0.028} & \result{0.321}{0.045} & \cellcolor{dvgreen}\best{0.327}{0.033} \\
 & $\rho\uparrow$ & \result{0.280}{0.028} & \result{0.301}{0.027} & \result{0.288}{0.093} & \result{0.268}{0.068} & \result{0.310}{0.070} & \result{0.304}{0.012} & \result{0.346}{0.069} & \cellcolor{dvgreen}\best{0.358}{0.058} \\
\addlinespace[3pt]
Fluency & NRMSE $\downarrow$ & \result{0.986}{0.034} & \result{0.945}{0.058} & \result{0.967}{0.023} & \result{0.974}{0.016} & \result{0.936}{0.067} & \result{0.958}{0.039} & \result{0.924}{0.025} & \cellcolor{dvgreen}\best{0.903}{0.043} \\
 & $r\uparrow$ & \result{0.200}{0.017} & \result{0.366}{0.055} & \result{0.422}{0.025} & \result{0.332}{0.012} & \result{0.390}{0.067} & \result{0.403}{0.044} & \result{0.415}{0.007} & \cellcolor{dvgreen}\best{0.450}{0.030} \\
 & $\rho\uparrow$ & \result{0.187}{0.049} & \result{0.465}{0.054} & \result{0.451}{0.043} & \result{0.409}{0.025} & \result{0.406}{0.062} & \result{0.448}{0.021} & \result{0.462}{0.011} & \cellcolor{dvgreen}\best{0.489}{0.004} \\
\addlinespace[3pt]
Style & NRMSE $\downarrow$ & \result{1.062}{0.051} & \result{1.037}{0.058} & \result{1.061}{0.092} & \result{1.087}{0.068} & \result{1.102}{0.053} & \result{1.066}{0.048} & \result{1.048}{0.067} & \cellcolor{dvgreen}\best{1.018}{0.056} \\
\bottomrule\end{tabular}}
\caption{RATE, Qwen 9B output-level metrics. Green highlights a winning \method{} cell and yellow highlights a winning baseline cell.}
\label{tab:targets-rate-qwen-9b}
\end{table*}
\begin{table*}[p]\centering\small\setlength{\tabcolsep}{3pt}
\resizebox{\textwidth}{!}{\begin{tabular}{llcccccccc}
\toprule
\textbf{Output} & \textbf{Metric} & \textbf{SAJA} & \textbf{T-MES} & \textbf{RELISH} & \textbf{MT-ICL} & \textbf{GenRe2} & \textbf{UniSO-N} & \textbf{GAT-AES} & \textbf{\method{} (Ours)} \\
\midrule
Accuracy & NRMSE $\downarrow$ & \result{0.936}{0.010} & \result{0.990}{0.011} & \result{1.013}{0.060} & \result{0.953}{0.017} & \result{0.973}{0.025} & \result{0.962}{0.028} & \result{0.933}{0.026} & \cellcolor{dvgreen}\best{0.904}{0.020} \\
 & $r\uparrow$ & \result{0.376}{0.042} & \result{0.293}{0.006} & \result{0.276}{0.119} & \result{0.395}{0.017} & \result{0.344}{0.070} & \result{0.385}{0.068} & \result{0.414}{0.036} & \cellcolor{dvgreen}\best{0.449}{0.038} \\
 & $\rho\uparrow$ & \result{0.341}{0.050} & \result{0.329}{0.062} & \result{0.363}{0.068} & \result{0.424}{0.030} & \result{0.385}{0.094} & \result{0.438}{0.098} & \result{0.468}{0.040} & \cellcolor{dvgreen}\best{0.499}{0.068} \\
\addlinespace[3pt]
Fluency & NRMSE $\downarrow$ & \result{0.944}{0.049} & \result{0.956}{0.030} & \result{1.000}{0.047} & \result{0.909}{0.035} & \result{0.935}{0.073} & \result{0.938}{0.033} & \result{0.908}{0.027} & \cellcolor{dvgreen}\best{0.878}{0.038} \\
 & $r\uparrow$ & \result{0.350}{0.057} & \result{0.357}{0.028} & \result{0.369}{0.087} & \result{0.446}{0.036} & \result{0.414}{0.049} & \result{0.429}{0.009} & \result{0.435}{0.015} & \cellcolor{dvgreen}\best{0.491}{0.023} \\
 & $\rho\uparrow$ & \result{0.362}{0.083} & \result{0.418}{0.008} & \result{0.429}{0.053} & \result{0.490}{0.013} & \result{0.455}{0.044} & \result{0.485}{0.010} & \result{0.499}{0.008} & \cellcolor{dvgreen}\best{0.530}{0.023} \\
\addlinespace[3pt]
Style & NRMSE $\downarrow$ & \result{1.051}{0.040} & \result{1.045}{0.051} & \result{1.038}{0.116} & \result{1.071}{0.072} & \result{1.090}{0.049} & \result{1.071}{0.042} & \result{1.052}{0.048} & \cellcolor{dvgreen}\best{1.016}{0.065} \\
\bottomrule\end{tabular}}
\caption{RATE, Qwen 27B output-level metrics. Green highlights a winning \method{} cell and yellow highlights a winning baseline cell.}
\label{tab:targets-rate-qwen-27b}
\end{table*}
\begin{table*}[p]\centering\small\setlength{\tabcolsep}{3pt}
\resizebox{\textwidth}{!}{\begin{tabular}{llcccccccc}
\toprule
\textbf{Output} & \textbf{Metric} & \textbf{SAJA} & \textbf{T-MES} & \textbf{RELISH} & \textbf{MT-ICL} & \textbf{GenRe2} & \textbf{UniSO-N} & \textbf{GAT-AES} & \textbf{\method{} (Ours)} \\
\midrule
Accuracy & NRMSE $\downarrow$ & \result{0.917}{0.019} & \result{0.924}{0.031} & \result{0.981}{0.066} & \result{0.912}{0.018} & \result{0.912}{0.027} & \result{0.964}{0.072} & \result{0.890}{0.048} & \cellcolor{dvgreen}\best{0.880}{0.037} \\
 & $r\uparrow$ & \result{0.424}{0.046} & \result{0.425}{0.036} & \result{0.381}{0.080} & \result{0.450}{0.019} & \result{0.449}{0.038} & \result{0.384}{0.106} & \result{0.476}{0.074} & \cellcolor{dvgreen}\best{0.491}{0.062} \\
 & $\rho\uparrow$ & \result{0.407}{0.024} & \result{0.469}{0.035} & \result{0.441}{0.044} & \result{0.497}{0.033} & \result{0.443}{0.035} & \result{0.408}{0.087} & \result{0.501}{0.058} & \cellcolor{dvgreen}\best{0.515}{0.049} \\
\addlinespace[3pt]
Fluency & NRMSE $\downarrow$ & \result{0.926}{0.008} & \result{0.926}{0.015} & \result{0.986}{0.066} & \result{0.887}{0.049} & \result{0.882}{0.046} & \result{0.938}{0.030} & \result{0.869}{0.016} & \cellcolor{dvgreen}\best{0.851}{0.036} \\
 & $r\uparrow$ & \result{0.394}{0.049} & \result{0.413}{0.023} & \result{0.390}{0.062} & \result{0.506}{0.025} & \result{0.488}{0.033} & \result{0.418}{0.075} & \result{0.516}{0.020} & \cellcolor{dvgreen}\best{0.539}{0.027} \\
 & $\rho\uparrow$ & \result{0.344}{0.008} & \result{0.471}{0.007} & \result{0.440}{0.069} & \result{0.560}{0.032} & \result{0.512}{0.034} & \result{0.443}{0.062} & \result{0.558}{0.014} & \cellcolor{dvgreen}\best{0.569}{0.039} \\
\addlinespace[3pt]
Style & NRMSE $\downarrow$ & \result{1.031}{0.051} & \result{1.178}{0.061} & \result{1.098}{0.052} & \result{1.042}{0.072} & \result{1.046}{0.083} & \result{1.076}{0.035} & \result{1.025}{0.085} & \cellcolor{dvgreen}\best{1.002}{0.064} \\
\bottomrule\end{tabular}}
\caption{RATE, Gemma 31B output-level metrics. Green highlights a winning \method{} cell and yellow highlights a winning baseline cell.}
\label{tab:targets-rate-gemma-31b}
\end{table*}
\begin{table*}[p]\centering\small\setlength{\tabcolsep}{3pt}
\resizebox{\textwidth}{!}{\begin{tabular}{llcccccccc}
\toprule
\textbf{Output} & \textbf{Metric} & \textbf{SAJA} & \textbf{T-MES} & \textbf{RELISH} & \textbf{MT-ICL} & \textbf{GenRe2} & \textbf{UniSO-N} & \textbf{GAT-AES} & \textbf{\method{} (Ours)} \\
\midrule
Accuracy & NRMSE $\downarrow$ & \result{0.997}{0.040} & \result{0.990}{0.004} & \result{1.018}{0.019} & \result{0.968}{0.022} & \result{0.987}{0.009} & \result{1.008}{0.024} & \result{0.964}{0.027} & \cellcolor{dvgreen}\best{0.937}{0.010} \\
 & $r\uparrow$ & \result{0.212}{0.083} & \result{0.245}{0.022} & \result{0.208}{0.052} & \result{0.318}{0.055} & \result{0.318}{0.009} & \result{0.279}{0.046} & \result{0.325}{0.035} & \cellcolor{dvgreen}\best{0.380}{0.018} \\
 & $\rho\uparrow$ & \result{0.189}{0.038} & \result{0.263}{0.043} & \result{0.294}{0.075} & \result{0.335}{0.055} & \result{0.358}{0.003} & \result{0.311}{0.066} & \result{0.366}{0.038} & \cellcolor{dvgreen}\best{0.421}{0.045} \\
\addlinespace[3pt]
Fluency & NRMSE $\downarrow$ & \result{0.987}{0.031} & \result{0.936}{0.037} & \result{0.978}{0.028} & \result{0.952}{0.049} & \result{0.944}{0.026} & \result{0.979}{0.039} & \result{0.954}{0.017} & \cellcolor{dvgreen}\best{0.913}{0.027} \\
 & $r\uparrow$ & \result{0.186}{0.080} & \result{0.380}{0.019} & \result{0.353}{0.083} & \result{0.365}{0.048} & \result{0.393}{0.008} & \result{0.387}{0.025} & \result{0.377}{0.024} & \cellcolor{dvgreen}\best{0.422}{0.012} \\
 & $\rho\uparrow$ & \result{0.215}{0.122} & \result{0.472}{0.048} & \result{0.411}{0.063} & \result{0.417}{0.089} & \result{0.433}{0.004} & \result{0.433}{0.019} & \result{0.426}{0.037} & \cellcolor{dvgreen}\best{0.475}{0.020} \\
\addlinespace[3pt]
Style & NRMSE $\downarrow$ & \result{1.064}{0.058} & \result{1.041}{0.047} & \result{1.091}{0.055} & \result{1.067}{0.079} & \result{1.064}{0.062} & \result{1.122}{0.043} & \result{1.063}{0.069} & \cellcolor{dvgreen}\best{1.032}{0.058} \\
\bottomrule\end{tabular}}
\caption{RATE, Llama 70B output-level metrics. Green highlights a winning \method{} cell and yellow highlights a winning baseline cell.}
\label{tab:targets-rate-llama-70b}
\end{table*}
\begin{table*}[p]\centering\small\setlength{\tabcolsep}{3pt}
\resizebox{\textwidth}{!}{\begin{tabular}{llcccccccc}
\toprule
\textbf{Output} & \textbf{Metric} & \textbf{SAJA} & \textbf{T-MES} & \textbf{RELISH} & \textbf{MT-ICL} & \textbf{GenRe2} & \textbf{UniSO-N} & \textbf{GAT-AES} & \textbf{\method{} (Ours)} \\
\midrule
Accuracy & NRMSE $\downarrow$ & \result{0.973}{0.010} & \result{0.999}{0.015} & \result{1.035}{0.016} & \result{0.977}{0.024} & \result{1.004}{0.022} & \result{1.013}{0.027} & \result{0.945}{0.031} & \cellcolor{dvgreen}\best{0.924}{0.022} \\
 & $r\uparrow$ & \result{0.295}{0.035} & \result{0.228}{0.049} & \result{0.215}{0.019} & \result{0.329}{0.038} & \result{0.308}{0.019} & \result{0.335}{0.056} & \result{0.368}{0.066} & \cellcolor{dvgreen}\best{0.404}{0.049} \\
 & $\rho\uparrow$ & \result{0.308}{0.032} & \result{0.227}{0.078} & \result{0.298}{0.042} & \result{0.384}{0.046} & \result{0.345}{0.045} & \result{0.358}{0.107} & \result{0.394}{0.096} & \cellcolor{dvgreen}\best{0.438}{0.064} \\
\addlinespace[3pt]
Fluency & NRMSE $\downarrow$ & \result{0.979}{0.022} & \result{0.935}{0.041} & \result{0.959}{0.033} & \result{0.924}{0.008} & \result{0.948}{0.060} & \result{0.946}{0.047} & \result{0.923}{0.005} & \cellcolor{dvgreen}\best{0.897}{0.032} \\
 & $r\uparrow$ & \result{0.244}{0.015} & \result{0.370}{0.025} & \result{0.367}{0.050} & \result{0.414}{0.029} & \result{0.389}{0.041} & \result{0.426}{0.030} & \result{0.421}{0.020} & \cellcolor{dvgreen}\best{0.460}{0.010} \\
 & $\rho\uparrow$ & \result{0.242}{0.023} & \result{0.445}{0.019} & \result{0.442}{0.055} & \result{0.472}{0.021} & \result{0.431}{0.012} & \result{0.474}{0.009} & \result{0.471}{0.038} & \cellcolor{dvgreen}\best{0.508}{0.015} \\
\addlinespace[3pt]
Style & NRMSE $\downarrow$ & \result{1.057}{0.044} & \result{1.043}{0.059} & \result{1.059}{0.071} & \result{1.079}{0.083} & \result{1.093}{0.037} & \result{1.143}{0.039} & \result{1.033}{0.060} & \cellcolor{dvgreen}\best{1.030}{0.065} \\
\bottomrule\end{tabular}}
\caption{RATE, Qwen 72B output-level metrics. Green highlights a winning \method{} cell and yellow highlights a winning baseline cell.}
\label{tab:targets-rate-qwen-72b}
\end{table*}
\begin{table*}[p]\centering\small\setlength{\tabcolsep}{3pt}
\resizebox{\textwidth}{!}{\begin{tabular}{llcccccccc}
\toprule
\textbf{Output} & \textbf{Metric} & \textbf{SAJA} & \textbf{T-MES} & \textbf{RELISH} & \textbf{MT-ICL} & \textbf{GenRe2} & \textbf{UniSO-N} & \textbf{GAT-AES} & \textbf{\method{} (Ours)} \\
\midrule
Accuracy & NRMSE $\downarrow$ & \result{0.976}{0.009} & \result{0.977}{0.012} & \result{1.023}{0.023} & \result{1.000}{0.002} & \result{0.997}{0.008} & \result{1.054}{0.068} & \result{0.987}{0.032} & \cellcolor{dvgreen}\best{0.956}{0.016} \\
 & $r\uparrow$ & \result{0.262}{0.017} & \result{0.304}{0.032} & \result{0.263}{0.026} & \result{0.225}{0.024} & \result{0.247}{0.016} & \result{0.211}{0.083} & \result{0.266}{0.068} & \cellcolor{dvgreen}\best{0.330}{0.021} \\
 & $\rho\uparrow$ & \result{0.268}{0.031} & \result{0.331}{0.043} & \cellcolor{dvyellow}\best{0.343}{0.053} & \result{0.245}{0.040} & \result{0.297}{0.037} & \result{0.217}{0.104} & \result{0.246}{0.114} & \result{0.338}{0.016} \\
\addlinespace[3pt]
Fluency & NRMSE $\downarrow$ & \result{0.986}{0.027} & \result{0.920}{0.048} & \result{0.985}{0.076} & \result{0.931}{0.045} & \result{0.931}{0.052} & \result{0.988}{0.035} & \result{0.953}{0.044} & \cellcolor{dvgreen}\best{0.911}{0.044} \\
 & $r\uparrow$ & \result{0.221}{0.039} & \result{0.399}{0.062} & \result{0.367}{0.073} & \result{0.388}{0.029} & \result{0.386}{0.040} & \result{0.314}{0.039} & \result{0.343}{0.038} & \cellcolor{dvgreen}\best{0.432}{0.035} \\
 & $\rho\uparrow$ & \result{0.269}{0.036} & \result{0.472}{0.047} & \result{0.399}{0.100} & \result{0.433}{0.046} & \result{0.431}{0.042} & \result{0.380}{0.034} & \result{0.400}{0.016} & \cellcolor{dvgreen}\best{0.482}{0.053} \\
\addlinespace[3pt]
Style & NRMSE $\downarrow$ & \result{1.073}{0.063} & \result{1.178}{0.061} & \result{1.079}{0.044} & \result{1.074}{0.056} & \result{1.073}{0.056} & \result{1.097}{0.065} & \result{1.063}{0.046} & \cellcolor{dvgreen}\best{1.040}{0.061} \\
\bottomrule\end{tabular}}
\caption{RATE, GPT 120B output-level metrics. Green highlights a winning \method{} cell and yellow highlights a winning baseline cell.}
\label{tab:targets-rate-gpt-120b}
\end{table*}
\begin{table*}[p]\centering\small\setlength{\tabcolsep}{3pt}
\resizebox{\textwidth}{!}{\begin{tabular}{llcccccccc}
\toprule
\textbf{Output} & \textbf{Metric} & \textbf{SAJA} & \textbf{T-MES} & \textbf{RELISH} & \textbf{MT-ICL} & \textbf{GenRe2} & \textbf{UniSO-N} & \textbf{GAT-AES} & \textbf{\method{} (Ours)} \\
\midrule
Accuracy & NRMSE $\downarrow$ & \result{0.948}{0.014} & \result{0.980}{0.016} & \result{1.025}{0.033} & \result{0.949}{0.024} & \result{0.955}{0.025} & \result{0.943}{0.029} & \result{0.929}{0.010} & \cellcolor{dvgreen}\best{0.893}{0.015} \\
 & $r\uparrow$ & \result{0.350}{0.013} & \result{0.282}{0.032} & \result{0.236}{0.083} & \result{0.363}{0.047} & \result{0.365}{0.049} & \result{0.395}{0.046} & \result{0.407}{0.018} & \cellcolor{dvgreen}\best{0.476}{0.029} \\
 & $\rho\uparrow$ & \result{0.374}{0.015} & \result{0.324}{0.050} & \result{0.320}{0.052} & \result{0.385}{0.056} & \result{0.390}{0.023} & \result{0.431}{0.050} & \result{0.443}{0.012} & \cellcolor{dvgreen}\best{0.505}{0.027} \\
\addlinespace[3pt]
Fluency & NRMSE $\downarrow$ & \result{0.940}{0.035} & \result{0.936}{0.045} & \result{0.956}{0.052} & \result{0.913}{0.054} & \result{0.902}{0.042} & \result{0.903}{0.051} & \result{0.904}{0.037} & \cellcolor{dvgreen}\best{0.879}{0.053} \\
 & $r\uparrow$ & \result{0.359}{0.018} & \result{0.372}{0.034} & \result{0.385}{0.062} & \result{0.456}{0.030} & \result{0.448}{0.035} & \result{0.470}{0.026} & \result{0.462}{0.028} & \cellcolor{dvgreen}\best{0.504}{0.041} \\
 & $\rho\uparrow$ & \result{0.367}{0.021} & \result{0.446}{0.025} & \result{0.434}{0.054} & \result{0.510}{0.057} & \result{0.475}{0.038} & \result{0.491}{0.036} & \result{0.503}{0.044} & \cellcolor{dvgreen}\best{0.537}{0.055} \\
\addlinespace[3pt]
Style & NRMSE $\downarrow$ & \result{1.061}{0.048} & \result{1.040}{0.052} & \result{1.063}{0.089} & \result{1.064}{0.064} & \result{1.064}{0.076} & \result{1.070}{0.042} & \result{1.045}{0.054} & \cellcolor{dvgreen}\best{1.016}{0.051} \\
\bottomrule\end{tabular}}
\caption{RATE, Qwen 122B output-level metrics. Green highlights a winning \method{} cell and yellow highlights a winning baseline cell.}
\label{tab:targets-rate-qwen-122b}
\end{table*}
\begin{table*}[p]\centering\small\setlength{\tabcolsep}{3pt}
\resizebox{\textwidth}{!}{\begin{tabular}{llcccccccc}
\toprule
\textbf{Output} & \textbf{Metric} & \textbf{SAJA} & \textbf{T-MES} & \textbf{RELISH} & \textbf{MT-ICL} & \textbf{GenRe2} & \textbf{UniSO-N} & \textbf{GAT-AES} & \textbf{\method{} (Ours)} \\
\midrule
Values & NRMSE $\downarrow$ & \result{1.023}{0.031} & \result{0.996}{0.027} & \result{1.068}{0.064} & \result{1.001}{0.035} & \result{1.010}{0.029} & \result{1.015}{0.029} & \result{1.000}{0.037} & \cellcolor{dvgreen}\best{0.980}{0.030} \\
 & $r\uparrow$ & \result{0.176}{0.041} & \result{0.283}{0.024} & \result{0.256}{0.009} & \result{0.266}{0.029} & \result{0.258}{0.015} & \result{0.229}{0.020} & \result{0.266}{0.021} & \cellcolor{dvgreen}\best{0.331}{0.013} \\
 & $\rho\uparrow$ & \result{0.151}{0.056} & \result{0.287}{0.019} & \result{0.272}{0.015} & \result{0.255}{0.018} & \result{0.242}{0.016} & \result{0.232}{0.002} & \result{0.253}{0.025} & \cellcolor{dvgreen}\best{0.319}{0.008} \\
\addlinespace[3pt]
Fluency & NRMSE $\downarrow$ & \result{0.998}{0.039} & \result{0.991}{0.035} & \result{1.020}{0.037} & \result{0.988}{0.039} & \result{0.999}{0.046} & \result{0.997}{0.046} & \result{0.992}{0.036} & \cellcolor{dvgreen}\best{0.963}{0.041} \\
 & $r\uparrow$ & \result{0.223}{0.013} & \result{0.259}{0.013} & \result{0.247}{0.017} & \result{0.266}{0.009} & \result{0.257}{0.015} & \result{0.252}{0.033} & \result{0.257}{0.009} & \cellcolor{dvgreen}\best{0.342}{0.019} \\
 & $\rho\uparrow$ & \result{0.124}{0.020} & \result{0.240}{0.013} & \result{0.254}{0.012} & \result{0.213}{0.009} & \result{0.238}{0.016} & \result{0.204}{0.023} & \result{0.222}{0.018} & \cellcolor{dvgreen}\best{0.304}{0.021} \\
\addlinespace[3pt]
Factuality & NRMSE $\downarrow$ & \result{0.986}{0.027} & \result{0.973}{0.027} & \result{0.984}{0.039} & \result{0.978}{0.029} & \result{0.983}{0.027} & \result{0.989}{0.026} & \result{0.973}{0.036} & \cellcolor{dvgreen}\best{0.951}{0.034} \\
\bottomrule\end{tabular}}
\caption{PRISM, Gemma E4B output-level metrics. Green highlights a winning \method{} cell and yellow highlights a winning baseline cell.}
\label{tab:targets-prism-gemma-e4b}
\end{table*}
\begin{table*}[p]\centering\small\setlength{\tabcolsep}{3pt}
\resizebox{\textwidth}{!}{%
}
\caption{PRISM, Qwen 9B output-level metrics. Green highlights a winning \method{} cell and yellow highlights a winning baseline cell.}
\label{tab:targets-prism-qwen-9b}
\end{table*}
\begin{table*}[p]\centering\small\setlength{\tabcolsep}{3pt}
\resizebox{\textwidth}{!}{%
}
\caption{PRISM, Qwen 27B output-level metrics. Green highlights a winning \method{} cell and yellow highlights a winning baseline cell.}
\label{tab:targets-prism-qwen-27b}
\end{table*}
\begin{table*}[p]\centering\small\setlength{\tabcolsep}{3pt}
\resizebox{\textwidth}{!}{%
}
\caption{PRISM, Gemma 31B output-level metrics. Green highlights a winning \method{} cell and yellow highlights a winning baseline cell.}
\label{tab:targets-prism-gemma-31b}
\end{table*}
\begin{table*}[p]\centering\small\setlength{\tabcolsep}{3pt}
\resizebox{\textwidth}{!}{%
}
\caption{PRISM, Llama 70B output-level metrics. Green highlights a winning \method{} cell and yellow highlights a winning baseline cell.}
\label{tab:targets-prism-llama-70b}
\end{table*}
\begin{table*}[p]\centering\small\setlength{\tabcolsep}{3pt}
\resizebox{\textwidth}{!}{%
}
\caption{PRISM, Qwen 72B output-level metrics. Green highlights a winning \method{} cell and yellow highlights a winning baseline cell.}
\label{tab:targets-prism-qwen-72b}
\end{table*}
\begin{table*}[p]\centering\small\setlength{\tabcolsep}{3pt}
\resizebox{\textwidth}{!}{%
}
\caption{PRISM, GPT 120B output-level metrics. Green highlights a winning \method{} cell and yellow highlights a winning baseline cell.}
\label{tab:targets-prism-gpt-120b}
\end{table*}
\begin{table*}[p]\centering\small\setlength{\tabcolsep}{3pt}
\resizebox{\textwidth}{!}{%
}
\caption{DSCodeBench, Gemma E4B output-level metrics. Green highlights a winning \method{} cell and yellow highlights a winning baseline cell.}
\label{tab:targets-dscodebench-gemma-e4b}
\end{table*}
\begin{table*}[p]\centering\small\setlength{\tabcolsep}{3pt}
\resizebox{\textwidth}{!}{%
}
\caption{DSCodeBench, Qwen 9B output-level metrics. Green highlights a winning \method{} cell and yellow highlights a winning baseline cell.}
\label{tab:targets-dscodebench-qwen-9b}
\end{table*}
\begin{table*}[p]\centering\small\setlength{\tabcolsep}{3pt}
\resizebox{\textwidth}{!}{%
}
\caption{DSCodeBench, Qwen 27B output-level metrics. Green highlights a winning \method{} cell and yellow highlights a winning baseline cell.}
\label{tab:targets-dscodebench-qwen-27b}
\end{table*}
\begin{table*}[p]\centering\small\setlength{\tabcolsep}{3pt}
\resizebox{\textwidth}{!}{%
}
\caption{DSCodeBench, Gemma 31B output-level metrics. Green highlights a winning \method{} cell and yellow highlights a winning baseline cell.}
\label{tab:targets-dscodebench-gemma-31b}
\end{table*}
\begin{table*}[p]\centering\small\setlength{\tabcolsep}{3pt}
\resizebox{\textwidth}{!}{%
}
\caption{DSCodeBench, Llama 70B output-level metrics. Green highlights a winning \method{} cell and yellow highlights a winning baseline cell.}
\label{tab:targets-dscodebench-llama-70b}
\end{table*}
\begin{table*}[p]\centering\small\setlength{\tabcolsep}{3pt}
\resizebox{\textwidth}{!}{%
}
\caption{DSCodeBench, Qwen 72B output-level metrics. Green highlights a winning \method{} cell and yellow highlights a winning baseline cell.}
\label{tab:targets-dscodebench-qwen-72b}
\end{table*}
\begin{table*}[p]\centering\small\setlength{\tabcolsep}{3pt}
\resizebox{\textwidth}{!}{%
}
\caption{DSCodeBench, GPT 120B output-level metrics. Green highlights a winning \method{} cell and yellow highlights a winning baseline cell.}
\label{tab:targets-dscodebench-gpt-120b}
\end{table*}
\begin{table*}[p]\centering\small\setlength{\tabcolsep}{3pt}
\resizebox{\textwidth}{!}{%
}
\caption{DSCodeBench, Qwen 122B output-level metrics. Green highlights a winning \method{} cell and yellow highlights a winning baseline cell.}
\label{tab:targets-dscodebench-qwen-122b}
\end{table*}

%% file: references.bib
@inproceedings{piskorz2026eliciting,
 author = {Piskorz, Julianna and Kobalczyk, Katarzyna and van der Schaar, Mihaela},
 booktitle = {International Conference on Learning Representations},
 editor = {C. Vondrick and B. Hariharan and C. Raffel and L. Pinto and D. Yang and A. Faust},
 pages = {135557--135582},
 title = {Eliciting Numerical Predictive Distributions of LLMs Without Auto-Regression},
 url = {https://proceedings.iclr.cc/paper_files/paper/2026/file/db79b6c50430a52fbab9d63efb8433ca-Paper-Conference.pdf},
 volume = {2026},
 year = {2026}
}

@inproceedings{zhu-etal-2026-text,
  address = {San Diego, California, United States},
  author = {Zhu, Yilun  and
Zhuang, Yuan  and
Vedula, Nikhita  and
Dhyani, Dushyanta  and
Xu, Shaoyuan  and
Bayati, Mohsen  and
Wang, Bryan  and
Malmasi, Shervin},
  booktitle = {Proceedings of the 64th Annual Meeting of the {A}ssociation for {C}omputational {L}inguistics (Volume 1: Long Papers)},
  doi = {10.18653/v1/2026.acl-long.758},
  editor = {Liakata, Maria  and
Moreira, Viviane P.  and
Zhang, Jiajun  and
Jurgens, David},
  isbn = {979-8-89176-390-6},
  month = {July},
  pages = {16632--16648},
  publisher = {Association for Computational Linguistics},
  title = {Text-to-Distribution Prediction with Quantile Tokens and Neighbor Context},
  url = {https://aclanthology.org/2026.acl-long.758/},
  year = {2026}
}

@article{IZENMAN1975248,
  author = {Izenman, Alan Julian},
  doi = {10.1016/0047-259x(75)90042-1},
  issn = {0047-259X},
  journal = {Journal of Multivariate Analysis},
  month = {June},
  number = {2},
  pages = {248--264},
  publisher = {Elsevier BV},
  title = {Reduced-rank regression for the multivariate linear model},
  url = {http://dx.doi.org/10.1016/0047-259X(75)90042-1},
  volume = {5},
  year = {1975}
}

@inproceedings{wang-etal-2024-interpretable,
  address = {Miami, Florida, USA},
  author = {Wang, Haoxiang  and
Xiong, Wei  and
Xie, Tengyang  and
Zhao, Han  and
Zhang, Tong},
  booktitle = {Findings of the Association for Computational Linguistics: EMNLP 2024},
  doi = {10.18653/v1/2024.findings-emnlp.620},
  editor = {Al-Onaizan, Yaser  and
Bansal, Mohit  and
Chen, Yun-Nung},
  month = {November},
  pages = {10582--10592},
  publisher = {Association for Computational Linguistics},
  title = {Interpretable Preferences via Multi-Objective Reward Modeling and Mixture-of-Experts},
  url = {https://aclanthology.org/2024.findings-emnlp.620/},
  year = {2024}
}

@inproceedings{li-ng-2025-graph,
  address = {Suzhou, China},
  author = {Li, Shengjie  and
Ng, Vincent},
  booktitle = {Proceedings of the 2025 Conference on Empirical Methods in Natural Language Processing},
  doi = {10.18653/v1/2025.emnlp-main.1691},
  editor = {Christodoulopoulos, Christos  and
Chakraborty, Tanmoy  and
Rose, Carolyn  and
Peng, Violet},
  isbn = {979-8-89176-332-6},
  month = {November},
  pages = {33325--33351},
  publisher = {Association for Computational Linguistics},
  title = {Graph-Based Multi-Trait Essay Scoring},
  url = {https://aclanthology.org/2025.emnlp-main.1691/},
  year = {2025}
}

@inproceedings{kim-etal-2025-biggen,
  address = {Albuquerque, New Mexico},
  author = {Kim, Seungone  and
Suk, Juyoung  and
Cho, Ji Yong  and
Longpre, Shayne  and
Kim, Chaeeun  and
Yoon, Dongkeun  and
Son, Guijin  and
Cho, Yejin  and
Shafayat, Sheikh  and
Baek, Jinheon  and
Park, Sue Hyun  and
Hwang, Hyeonbin  and
Jo, Jinkyung  and
Cho, Hyowon  and
Shin, Haebin  and
Lee, Seongyun  and
Oh, Hanseok  and
Lee, Noah  and
Ho, Namgyu  and
Joo, Se June  and
Ko, Miyoung  and
Lee, Yoonjoo  and
Chae, Hyungjoo  and
Shin, Jamin  and
Jang, Joel  and
Ye, Seonghyeon  and
Lin, Bill Yuchen  and
Welleck, Sean  and
Neubig, Graham  and
Lee, Moontae  and
Lee, Kyungjae  and
Seo, Minjoon},
  booktitle = {Proceedings of the 2025 Conference of the Nations of the Americas Chapter of the Association for Computational Linguistics: Human Language Technologies (Volume 1: Long Papers)},
  doi = {10.18653/v1/2025.naacl-long.303},
  editor = {Chiruzzo, Luis  and
Ritter, Alan  and
Wang, Lu},
  isbn = {979-8-89176-189-6},
  month = {April},
  pages = {5877--5919},
  publisher = {Association for Computational Linguistics},
  title = {The {B}i{GG}en Bench: A Principled Benchmark for Fine-grained Evaluation of Language Models with Language Models},
  url = {https://aclanthology.org/2025.naacl-long.303/},
  year = {2025}
}

@inproceedings{hashemi-etal-2024-llm,
  address = {Bangkok, Thailand},
  author = {Hashemi, Helia  and
Eisner, Jason  and
Rosset, Corby  and
Van Durme, Benjamin  and
Kedzie, Chris},
  booktitle = {Proceedings of the 62nd Annual Meeting of the Association for Computational Linguistics (Volume 1: Long Papers)},
  doi = {10.18653/v1/2024.acl-long.745},
  editor = {Ku, Lun-Wei  and
Martins, Andre  and
Srikumar, Vivek},
  month = {August},
  pages = {13806--13834},
  publisher = {Association for Computational Linguistics},
  title = {{LLM}-Rubric: A Multidimensional, Calibrated Approach to Automated Evaluation of Natural Language Texts},
  url = {https://aclanthology.org/2024.acl-long.745/},
  year = {2024}
}

@inproceedings{Chung2026-nj,
  author = {Chung, Jiwan and KIM, SAEJIN and Jo, Yongrae and Park, Jaewoo and Min, Dongjun and Yu, Youngjae},
  booktitle = {International Conference on Learning Representations},
  editor = {C. Vondrick and B. Hariharan and C. Raffel and L. Pinto and D. Yang and A. Faust},
  pages = {11633--11655},
  title = {Teaching Metric Distance to Discrete Autoregressive Language Models},
  url = {https://proceedings.iclr.cc/paper_files/paper/2026/file/13b45b44e26c353c64cba9529bf4724f-Paper-Conference.pdf},
  volume = {2026},
  year = {2026}
}

@inproceedings{Zuo2026-jc,
  author = {Zhuo Zuo and Li Yue and Wenhao Zheng and Chenpeng Wang and Xianggen Liu},
  booktitle = {Proceedings of the 43rd International Conference on Machine Learning},
  month = {July},
  title = {Enhancing Numerical Prediction in LLMs via Smooth MMD Alignment},
  url = {https://icml.cc/virtual/2026/poster/64143},
  year = {2026}
}

@inproceedings{Zhang2026-mc,
  author = {Yasi Zhang and Tianyu Chen and Mingyuan Zhou and Oscar Leong and Ying Nian Wu and Michal Lukasik},
  booktitle = {Proceedings of the 43rd International Conference on Machine Learning},
  month = {July},
  title = {REAL: Regression-Aware Reinforcement Learning for LLM-as-a-Judge},
  url = {https://icml.cc/virtual/2026/poster/62715},
  year = {2026}
}

@inproceedings{akhauri2026regression,
  author = {Yash Akhauri and Xingyou Song and Arissa Wongpanich and Bryan Lewandowski and Mohamed S. Abdelfattah},
  booktitle = {Proceedings of the 43rd International Conference on Machine Learning},
  month = {July},
  title = {Regression Language Models for Code},
  url = {https://icml.cc/virtual/2026/poster/60978},
  year = {2026}
}

@inproceedings{
su2026relish,
title={{RELISH}: {LLM} {RE}gression with a Latent Iterative State Head},
author={Yiheng Su and Matthew Lease},
booktitle={Third Conference on Language Modeling},
year={2026},
url={https://openreview.net/forum?id=IebWFoQwdz}
}

@inproceedings{chen2026genre2,
  author = {Ming Chen and Sheng Tang and Rong-Xi Tan and Ziniu Li and Jiacheng Chen and Ke Xue and Chao Qian},
  booktitle = {Proceedings of the 43rd International Conference on Machine Learning},
  month = {July},
  title = {Beyond Token-level Supervision: Unlocking the Potential of Decoding-based Regression via Reinforcement Learning},
  url = {https://icml.cc/virtual/2026/poster/60986},
  year = {2026}
}

@inproceedings{lukasik2025raft,
  author = {Lukasik, Michal and Meng, Zhao and Narasimhan, Harikrishna and Chang, Yin-Wen and Menon, Aditya Krishna and Yu, Felix and Kumar, Sanjiv},
  booktitle = {International Conference on Learning Representations},
  editor = {Y. Yue and A. Garg and N. Peng and F. Sha and R. Yu},
  pages = {69871--69901},
  title = {Better autoregressive regression with LLMs via regression-aware fine-tuning},
  url = {https://proceedings.iclr.cc/paper_files/paper/2025/file/ae3bb0adcdce28afb5f7e86a79b6cffa-Paper-Conference.pdf},
  volume = {2025},
  year = {2025}
}

@article{song2025decoding,
  author = {Xingyou Song and Dara Bahri},
  issn = {2835-8856},
  journal = {Transactions on Machine Learning Research},
  title = {Decoding-based Regression},
  url = {https://openreview.net/forum?id=avUQ8jguxg},
  year = {2025}
}

@inproceedings{wang-liu-2025-mes,
  address = {Abu Dhabi, UAE},
  author = {Wang, Jiong  and
Liu, Jie},
  booktitle = {Proceedings of the 31st International Conference on Computational Linguistics},
  editor = {Rambow, Owen  and
Wanner, Leo  and
Apidianaki, Marianna  and
Al-Khalifa, Hend  and
Eugenio, Barbara Di  and
Schockaert, Steven},
  month = {January},
  pages = {1224--1236},
  publisher = {Association for Computational Linguistics},
  title = {{T}-{MES}: Trait-Aware Mix-of-Experts Representation Learning for Multi-trait Essay Scoring},
  url = {https://aclanthology.org/2025.coling-main.81/},
  year = {2025}
}

@inproceedings{yuchi-etal-2026-llms,
  address = {Rabat, Morocco},
  author = {Yuchi, Fengting  and
Du, Li  and
Eisner, Jason},
  booktitle = {Proceedings of the 19th Conference of the {E}uropean Chapter of the {A}ssociation for {C}omputational {L}inguistics (Volume 2: Short Papers)},
  doi = {10.18653/v1/2026.eacl-short.47},
  editor = {Demberg, Vera  and
Inui, Kentaro  and
Marquez, Llu{\'i}s},
  isbn = {979-8-89176-381-4},
  month = {March},
  pages = {659--673},
  publisher = {Association for Computational Linguistics},
  title = {{LLM}s Know More About Numbers than They Can Say},
  url = {https://aclanthology.org/2026.eacl-short.47/},
  year = {2026}
}

@inproceedings{liu2023geval,
  address = {Singapore},
  author = {Liu, Yang  and
Iter, Dan  and
Xu, Yichong  and
Wang, Shuohang  and
Xu, Ruochen  and
Zhu, Chenguang},
  booktitle = {Proceedings of the 2023 Conference on Empirical Methods in Natural Language Processing},
  doi = {10.18653/v1/2023.emnlp-main.153},
  editor = {Bouamor, Houda  and
Pino, Juan  and
Bali, Kalika},
  month = {December},
  pages = {2511--2522},
  publisher = {Association for Computational Linguistics},
  title = {{G}-Eval: {NLG} Evaluation using Gpt-4 with Better Human Alignment},
  url = {https://aclanthology.org/2023.emnlp-main.153/},
  year = {2023}
}

@inproceedings{kola-etal-2026-saja,
  address = {San Diego, California, USA},
  author = {Kola, Sneha  and
Sharma, Pankaj Kumar  and
Dey, Soumyadeep  and
Mishra, Bamdev  and
Datar, Mayur},
  booktitle = {Proceedings of the 64th Annual Meeting of the {A}ssociation for {C}omputational {L}inguistics (Volume 6: Industry Track)},
  doi = {10.18653/v1/2026.acl-industry.45},
  editor = {Li, Yunyao  and
Rehm, Georg  and
Tu, Mei},
  isbn = {979-8-89176-394-4},
  month = {July},
  pages = {646--664},
  publisher = {Association for Computational Linguistics},
  title = {{SAJA}: A Simple Approach to Judge Alignment for {LLM}-as-a-Judge},
  url = {https://aclanthology.org/2026.acl-industry.45/},
  year = {2026}
}

@inproceedings{hu-etal-2025-impartial,
  address = {Vienna, Austria},
  author = {Hu, Dou  and
Wei, Lingwei  and
Zhou, Wei  and
Hu, Songlin},
  booktitle = {Proceedings of the 63rd Annual Meeting of the Association for Computational Linguistics (Volume 1: Long Papers)},
  doi = {10.18653/v1/2025.acl-long.975},
  editor = {Che, Wanxiang  and
Nabende, Joyce  and
Shutova, Ekaterina  and
Pilehvar, Mohammad Taher},
  isbn = {979-8-89176-251-0},
  month = {July},
  pages = {19883--19897},
  publisher = {Association for Computational Linguistics},
  title = {Impartial Multi-task Representation Learning via Variance-invariant Probabilistic Decoding},
  url = {https://aclanthology.org/2025.acl-long.975/},
  year = {2025}
}

@inproceedings{hu-etal-2026-multi,
  address = {San Diego, California, United States},
  author = {Hu, Dou  and
Wei, Lingwei  and
Xiao, Hongjiang  and
Hu, Songlin  and
Zhang, Yuan},
  booktitle = {Proceedings of the 64th Annual Meeting of the {A}ssociation for {C}omputational {L}inguistics (Volume 1: Long Papers)},
  doi = {10.18653/v1/2026.acl-long.2120},
  editor = {Liakata, Maria  and
Moreira, Viviane P.  and
Zhang, Jiajun  and
Jurgens, David},
  isbn = {979-8-89176-390-6},
  month = {July},
  pages = {45722--45740},
  publisher = {Association for Computational Linguistics},
  title = {Multi-Task Representation Alignment on Language Understanding: A Mutual Information Perspective},
  url = {https://aclanthology.org/2026.acl-long.2120/},
  year = {2026}
}

@inproceedings{tan2025uniso,
  author = {Tan, Rong-Xi and Chen, Ming and Xue, Ke and Wang, Yao and Wang, Yaoyuan and Sheng, Fu and Qian, Chao},
  booktitle = {Proceedings of the 42nd International Conference on Machine Learning},
  editor = {Singh, Aarti and Fazel, Maryam and Hsu, Daniel and Lacoste-Julien, Simon and Berkenkamp, Felix and Maharaj, Tegan and Wagstaff, Kiri and Zhu, Jerry},
  month = {13--19 Jul},
  pages = {58499--58544},
  publisher = {PMLR},
  series = {Proceedings of Machine Learning Research},
  title = {Towards Universal Offline Black-Box Optimization via Learning Language Model Embeddings},
  url = {https://proceedings.mlr.press/v267/tan25b.html},
  volume = {267},
  year = {2025}
}

@inproceedings{zhu2026multitask,
  author = {Qingyang Zhu and Eric Karl Oermann and Kyunghyun Cho},
  booktitle = {Proceedings of the 43rd International Conference on Machine Learning},
  month = {July},
  title = {Multi-Task Bayesian In-Context Learning},
  url = {https://icml.cc/virtual/2026/poster/63501},
  year = {2026}
}

@article{fabbri-etal-2021-summeval,
  address = {Cambridge, MA},
  author = {Fabbri, Alexander R.  and
Kry{\'s}ci{\'n}ski, Wojciech  and
McCann, Bryan  and
Xiong, Caiming  and
Socher, Richard  and
Radev, Dragomir},
  doi = {10.1162/tacl_a_00373},
  editor = {Roark, Brian  and
Nenkova, Ani},
  journal = {Transactions of the Association for Computational Linguistics},
  pages = {391--409},
  publisher = {MIT Press},
  title = {{S}umm{E}val: Re-evaluating Summarization Evaluation},
  url = {https://aclanthology.org/2021.tacl-1.24/},
  volume = {9},
  year = {2021}
}

@inproceedings{do-etal-2024-autoregressive,
  address = {St. Julian{'}s, Malta},
  author = {Do, Heejin  and
Kim, Yunsu  and
Lee, Gary},
  booktitle = {Findings of the Association for Computational Linguistics: EACL 2024},
  doi = {10.18653/v1/2024.findings-eacl.115},
  editor = {Graham, Yvette  and
Purver, Matthew},
  month = {March},
  pages = {1659--1666},
  publisher = {Association for Computational Linguistics},
  title = {Autoregressive Score Generation for Multi-trait Essay Scoring},
  url = {https://aclanthology.org/2024.findings-eacl.115/},
  year = {2024}
}

@inproceedings{lukasik-etal-2024-regression,
  address = {Miami, Florida, USA},
  author = {Lukasik, Michal  and
Narasimhan, Harikrishna  and
Menon, Aditya Krishna  and
Yu, Felix  and
Kumar, Sanjiv},
  booktitle = {Findings of the Association for Computational Linguistics: EMNLP 2024},
  doi = {10.18653/v1/2024.findings-emnlp.799},
  editor = {Al-Onaizan, Yaser  and
Bansal, Mohit  and
Chen, Yun-Nung},
  month = {November},
  pages = {13667--13678},
  publisher = {Association for Computational Linguistics},
  title = {Regression Aware Inference with {LLM}s},
  url = {https://aclanthology.org/2024.findings-emnlp.799/},
  year = {2024}
}

@inproceedings{nafar-etal-2025-learning,
  address = {Albuquerque, New Mexico},
  author = {Nafar, Aliakbar  and
Venable, K. Brent  and
Kordjamshidi, Parisa},
  booktitle = {Proceedings of the 2025 Conference of the Nations of the Americas Chapter of the Association for Computational Linguistics: Human Language Technologies (Volume 1: Long Papers)},
  doi = {10.18653/v1/2025.naacl-long.417},
  editor = {Chiruzzo, Luis  and
Ritter, Alan  and
Wang, Lu},
  isbn = {979-8-89176-189-6},
  month = {April},
  pages = {8206--8229},
  publisher = {Association for Computational Linguistics},
  title = {Learning vs Retrieval: The Role of In-Context Examples in Regression with Large Language Models},
  url = {https://aclanthology.org/2025.naacl-long.417/},
  year = {2025}
}

@inproceedings{chiang-etal-2025-tract,
  address = {Vienna, Austria},
  author = {Chiang, Cheng-Han  and
Lee, Hung-yi  and
Lukasik, Michal},
  booktitle = {Proceedings of the 63rd Annual Meeting of the Association for Computational Linguistics (Volume 1: Long Papers)},
  doi = {10.18653/v1/2025.acl-long.147},
  editor = {Che, Wanxiang  and
Nabende, Joyce  and
Shutova, Ekaterina  and
Pilehvar, Mohammad Taher},
  isbn = {979-8-89176-251-0},
  month = {July},
  pages = {2934--2952},
  publisher = {Association for Computational Linguistics},
  title = {{TRACT}: Regression-Aware Fine-tuning Meets Chain-of-Thought Reasoning for {LLM}-as-a-Judge},
  url = {https://aclanthology.org/2025.acl-long.147/},
  year = {2025}
}

@article{tang2025understanding,
  author = {Eric Tang and Bangding Yang and Xingyou Song},
  issn = {2835-8856},
  journal = {Transactions on Machine Learning Research},
  title = {Understanding {LLM} Embeddings for Regression},
  url = {https://openreview.net/forum?id=Wt6Iz5XNIO},
  year = {2025}
}

@inproceedings{vacareanu2024words,
  author = {Robert Vacareanu and Vlad-Andrei Negru and Vasile Suciu and Mihai Surdeanu},
  booktitle = {First Conference on Language Modeling},
  title = {From Words to Numbers: Your Large Language Model Is Secretly A Capable Regressor When Given In-Context Examples},
  url = {https://openreview.net/forum?id=LzpaUxcNFK},
  year = {2024}
}

@inproceedings{sener2018multiobjective,
  author = {Sener, Ozan and Koltun, Vladlen},
  booktitle = {Advances in Neural Information Processing Systems},
  editor = {S. Bengio and H. Wallach and H. Larochelle and K. Grauman and N. Cesa-Bianchi and R. Garnett},
  publisher = {Curran Associates, Inc.},
  title = {Multi-Task Learning as Multi-Objective Optimization},
  url = {https://proceedings.neurips.cc/paper_files/paper/2018/file/432aca3a1e345e339f35a30c8f65edce-Paper.pdf},
  volume = {31},
  year = {2018}
}

@inproceedings{yu2020pcgrad,
  author = {Yu, Tianhe and Kumar, Saurabh and Gupta, Abhishek and Levine, Sergey and Hausman, Karol and Finn, Chelsea},
  booktitle = {Advances in Neural Information Processing Systems},
  editor = {H. Larochelle and M. Ranzato and R. Hadsell and M.F. Balcan and H. Lin},
  pages = {5824--5836},
  publisher = {Curran Associates, Inc.},
  title = {Gradient Surgery for Multi-Task Learning},
  url = {https://proceedings.neurips.cc/paper_files/paper/2020/file/3fe78a8acf5fda99de95303940a2420c-Paper.pdf},
  volume = {33},
  year = {2020}
}

@inproceedings{liu2021cagrad,
  author = {Liu, Bo and Liu, Xingchao and Jin, Xiaojie and Stone, Peter and Liu, Qiang},
  booktitle = {Advances in Neural Information Processing Systems},
  editor = {M. Ranzato and A. Beygelzimer and Y. Dauphin and P.S. Liang and J. Wortman Vaughan},
  pages = {18878--18890},
  publisher = {Curran Associates, Inc.},
  title = {Conflict-Averse Gradient Descent for Multi-task learning},
  url = {https://proceedings.neurips.cc/paper_files/paper/2021/file/9d27fdf2477ffbff837d73ef7ae23db9-Paper.pdf},
  volume = {34},
  year = {2021}
}

@inproceedings{navon2022nash,
  author = {Navon, Aviv and Shamsian, Aviv and Achituve, Idan and Maron, Haggai and Kawaguchi, Kenji and Chechik, Gal and Fetaya, Ethan},
  booktitle = {Proceedings of the 39th International Conference on Machine Learning},
  editor = {Chaudhuri, Kamalika and Jegelka, Stefanie and Song, Le and Szepesvari, Csaba and Niu, Gang and Sabato, Sivan},
  month = {17--23 Jul},
  pages = {16428--16446},
  publisher = {PMLR},
  series = {Proceedings of Machine Learning Research},
  title = {Multi-Task Learning as a Bargaining Game},
  url = {https://proceedings.mlr.press/v162/navon22a.html},
  volume = {162},
  year = {2022}
}

@inproceedings{liu2023famo,
  author = {Liu, Bo and Feng, Yihao and Stone, Peter and Liu, Qiang},
  booktitle = {Advances in Neural Information Processing Systems},
  doi = {10.52202/075280-2500},
  editor = {A. Oh and T. Naumann and A. Globerson and K. Saenko and M. Hardt and S. Levine},
  pages = {57226--57243},
  publisher = {Curran Associates, Inc.},
  title = {FAMO: Fast Adaptive Multitask Optimization},
  url = {https://proceedings.neurips.cc/paper_files/paper/2023/file/b2fe1ee8d936ac08dd26f2ff58986c8f-Paper-Conference.pdf},
  volume = {36},
  year = {2023}
}

@inproceedings{jeong2025selective,
  author = {Jeong, Wooseong and Yoon, Kuk-Jin},
  booktitle = {International Conference on Learning Representations},
  editor = {Y. Yue and A. Garg and N. Peng and F. Sha and R. Yu},
  pages = {95046--95079},
  title = {Selective Task Group Updates for Multi-Task Optimization},
  url = {https://proceedings.iclr.cc/paper_files/paper/2025/file/ed3cd2520148b577039adfade82a5566-Paper-Conference.pdf},
  volume = {2025},
  year = {2025}
}

@inproceedings{efroni2025aligned,
  author = {Efroni, Yonathan and Kretzu, Ben and Jiang, Daniel R. and Bhandari, Jalaj and Zhu, Zheqing and Ullrich, Karen},
  booktitle = {Proceedings of the 42nd International Conference on Machine Learning},
  editor = {Singh, Aarti and Fazel, Maryam and Hsu, Daniel and Lacoste-Julien, Simon and Berkenkamp, Felix and Maharaj, Tegan and Wagstaff, Kiri and Zhu, Jerry},
  month = {13--19 Jul},
  pages = {14989--15017},
  publisher = {PMLR},
  series = {Proceedings of Machine Learning Research},
  title = {Aligned Multi Objective Optimization},
  url = {https://proceedings.mlr.press/v267/efroni25a.html},
  volume = {267},
  year = {2025}
}

@inproceedings{ayman2026etap,
  author = {Ayman, Afiya and Mukhopadhyay, Ayan and Laszka, Aron},
  booktitle = {International Conference on Learning Representations},
  editor = {C. Vondrick and B. Hariharan and C. Raffel and L. Pinto and D. Yang and A. Faust},
  pages = {111821--111843},
  title = {Ensemble Prediction of Task Affinity for Efficient Multi-Task Learning},
  url = {https://proceedings.iclr.cc/paper_files/paper/2026/file/b595b039d8f3e04649e45879f487e707-Paper-Conference.pdf},
  volume = {2026},
  year = {2026}
}

@inproceedings{li-etal-2025-gradient,
  address = {Vienna, Austria},
  author = {Li, Chengao  and
Zhang, Hanyu  and
Xu, Yunkun  and
Xue, Hongyan  and
Ao, Xiang  and
He, Qing},
  booktitle = {Proceedings of the 63rd Annual Meeting of the Association for Computational Linguistics (Volume 1: Long Papers)},
  doi = {10.18653/v1/2025.acl-long.549},
  editor = {Che, Wanxiang  and
Nabende, Joyce  and
Shutova, Ekaterina  and
Pilehvar, Mohammad Taher},
  isbn = {979-8-89176-251-0},
  month = {July},
  pages = {11214--11232},
  publisher = {Association for Computational Linguistics},
  title = {Gradient-Adaptive Policy Optimization: Towards Multi-Objective Alignment of Large Language Models},
  url = {https://aclanthology.org/2025.acl-long.549/},
  year = {2025}
}

@inproceedings{ban2024fairgrad,
  author = {Ban, Hao and Ji, Kaiyi},
  booktitle = {Proceedings of the 41st International Conference on Machine Learning},
  editor = {Salakhutdinov, Ruslan and Kolter, Zico and Heller, Katherine and Weller, Adrian and Oliver, Nuria and Scarlett, Jonathan and Berkenkamp, Felix},
  month = {21--27 Jul},
  pages = {2715--2731},
  publisher = {PMLR},
  series = {Proceedings of Machine Learning Research},
  title = {Fair Resource Allocation in Multi-Task Learning},
  url = {https://proceedings.mlr.press/v235/ban24a.html},
  volume = {235},
  year = {2024}
}

@inproceedings{he2024excessmtl,
  author = {He, Yifei and Zhou, Shiji and Zhang, Guojun and Yun, Hyokun and Xu, Yi and Zeng, Belinda and Chilimbi, Trishul and Zhao, Han},
  booktitle = {Proceedings of the 41st International Conference on Machine Learning},
  editor = {Salakhutdinov, Ruslan and Kolter, Zico and Heller, Katherine and Weller, Adrian and Oliver, Nuria and Scarlett, Jonathan and Berkenkamp, Felix},
  month = {21--27 Jul},
  pages = {18094--18114},
  publisher = {PMLR},
  series = {Proceedings of Machine Learning Research},
  title = {Robust Multi-Task Learning with Excess Risks},
  url = {https://proceedings.mlr.press/v235/he24n.html},
  volume = {235},
  year = {2024}
}

@inproceedings{do-etal-2024-autoregressive-multi,
  address = {Miami, Florida, USA},
  author = {Do, Heejin  and
Ryu, Sangwon  and
Lee, Gary},
  booktitle = {Proceedings of the 2024 Conference on Empirical Methods in Natural Language Processing},
  doi = {10.18653/v1/2024.emnlp-main.917},
  editor = {Al-Onaizan, Yaser  and
Bansal, Mohit  and
Chen, Yun-Nung},
  month = {November},
  pages = {16427--16438},
  publisher = {Association for Computational Linguistics},
  title = {Autoregressive Multi-trait Essay Scoring via Reinforcement Learning with Scoring-aware Multiple Rewards},
  url = {https://aclanthology.org/2024.emnlp-main.917/},
  year = {2024}
}

@inproceedings{saad-falcon-etal-2025-lmunit,
  address = {Suzhou, China},
  author = {Saad-Falcon, Jon  and
Vivek, Rajan Pathe  and
Berrios, William  and
Naik, Nandita Shankar  and
Franklin, Matija  and
Vidgen, Bertie  and
Singh, Amanpreet  and
Kiela, Douwe  and
Mehri, Shikib},
  booktitle = {Findings of the Association for Computational Linguistics: EMNLP 2025},
  doi = {10.18653/v1/2025.findings-emnlp.176},
  editor = {Christodoulopoulos, Christos  and
Chakraborty, Tanmoy  and
Rose, Carolyn  and
Peng, Violet},
  isbn = {979-8-89176-335-7},
  month = {November},
  pages = {3303--3324},
  publisher = {Association for Computational Linguistics},
  title = {{LMUNIT}: Fine-grained Evaluation with Natural Language Unit Tests},
  url = {https://aclanthology.org/2025.findings-emnlp.176/},
  year = {2025}
}

@inproceedings{ding-etal-2023-score,
  address = {Toronto, Canada},
  author = {Ding, Yuning  and
Bexte, Marie  and
Horbach, Andrea},
  booktitle = {Findings of the Association for Computational Linguistics: ACL 2023},
  doi = {10.18653/v1/2023.findings-acl.825},
  editor = {Rogers, Anna  and
Boyd-Graber, Jordan  and
Okazaki, Naoaki},
  month = {July},
  pages = {13052--13063},
  publisher = {Association for Computational Linguistics},
  title = {Score It All Together: A Multi-Task Learning Study on Automatic Scoring of Argumentative Essays},
  url = {https://aclanthology.org/2023.findings-acl.825/},
  year = {2023}
}

@inproceedings{zhang2026complementary,
  author = {Zhang, Zhiwei and Liu, Hui and Li, Xiaomin and Dai, Zhenwei and Zeng, Jingying and Wang, Fali and Lin, Minhua and Chandradevan, Ramraj and Wu, Linlin and Li, Zhen and Luo, Chen and Wu, Zongyu and Tang, Xianfeng and He, Qi and Wang, Suhang},
  booktitle = {International Conference on Learning Representations},
  editor = {C. Vondrick and B. Hariharan and C. Raffel and L. Pinto and D. Yang and A. Faust},
  pages = {22736--22776},
  title = {Bradley-Terry and Multi-Objective Reward Modeling Are Complementary},
  url = {https://proceedings.iclr.cc/paper_files/paper/2026/file/2703e3329fef63ed7690f1b8145f25aa-Paper-Conference.pdf},
  volume = {2026},
  year = {2026}
}

@inproceedings{chen-etal-2026-multi-agent,
  address = {San Diego, California, United States},
  author = {Chen, Jiaju  and
Lu, Yuxuan  and
Wang, Xiaojie  and
Zeng, Huimin  and
Huang, Jing  and
Gesi, Jiri  and
Xu, Ying  and
Yao, Bingsheng  and
Wang, Dakuo},
  booktitle = {Proceedings of the 64th Annual Meeting of the {A}ssociation for {C}omputational {L}inguistics (Volume 1: Long Papers)},
  doi = {10.18653/v1/2026.acl-long.790},
  editor = {Liakata, Maria  and
Moreira, Viviane P.  and
Zhang, Jiajun  and
Jurgens, David},
  isbn = {979-8-89176-390-6},
  month = {July},
  pages = {17389--17416},
  publisher = {Association for Computational Linguistics},
  title = {Multi-Agent-as-Judge: Aligning {LLM}-Agent-Based Automated Evaluation with Multi-Dimensional Human Evaluation},
  url = {https://aclanthology.org/2026.acl-long.790/},
  year = {2026}
}

@inproceedings{ICLR2025_06694da0,
  author = {Qiu, Ruizhong and Zeng, Weiliang and Ezick, James and Lott, Christopher and Tong, Hanghang},
  booktitle = {International Conference on Learning Representations},
  editor = {Y. Yue and A. Garg and N. Peng and F. Sha and R. Yu},
  pages = {2233--2261},
  title = {How efficient is LLM-generated code? A rigorous \& high-standard benchmark},
  url = {https://proceedings.iclr.cc/paper_files/paper/2025/file/06694da057cb15fef11542270a592627-Paper-Conference.pdf},
  volume = {2025},
  year = {2025}
}

@inproceedings{ICLR2025_ea94957d,
  author = {Si, Chenglei and Yang, Diyi and Hashimoto, Tatsunori},
  booktitle = {International Conference on Learning Representations},
  editor = {Y. Yue and A. Garg and N. Peng and F. Sha and R. Yu},
  pages = {94003--94092},
  title = {Can LLMs Generate Novel Research Ideas? A Large-Scale Human Study with 100+ NLP Researchers},
  url = {https://proceedings.iclr.cc/paper_files/paper/2025/file/ea94957d81b1c1caf87ef5319fa6b467-Paper-Conference.pdf},
  volume = {2025},
  year = {2025}
}

@inproceedings{NEURIPS2023_3c2b60a3,
  author = {Allison, Robert and Stephenson, Anthony and F, Samuel and Pyzer-Knapp, Edward O},
  booktitle = {Advances in Neural Information Processing Systems},
  doi = {10.52202/075280-0829},
  editor = {A. Oh and T. Naumann and A. Globerson and K. Saenko and M. Hardt and S. Levine},
  pages = {18906--18931},
  publisher = {Curran Associates, Inc.},
  title = {Leveraging Locality and Robustness to Achieve Massively Scalable Gaussian Process Regression},
  url = {https://proceedings.neurips.cc/paper_files/paper/2023/file/3c2b60a3f269c404e9329ee119f2d34a-Paper-Conference.pdf},
  volume = {36},
  year = {2023}
}

@inproceedings{NEURIPS2024_2ee1c872,
  author = {Holzm\"{u}ller, David and Grinsztajn, L\'{e}o and Steinwart, Ingo},
  booktitle = {Advances in Neural Information Processing Systems},
  doi = {10.52202/079017-0837},
  editor = {A. Globerson and L. Mackey and D. Belgrave and A. Fan and U. Paquet and J. Tomczak and C. Zhang},
  pages = {26577--26658},
  publisher = {Curran Associates, Inc.},
  title = {Better by default: Strong pre-tuned MLPs and boosted trees on tabular data},
  url = {https://proceedings.neurips.cc/paper_files/paper/2024/file/2ee1c87245956e3eaa71aaba5f5753eb-Paper-Conference.pdf},
  volume = {37},
  year = {2024}
}

@inproceedings{lin-etal-2023-counterfactual,
  address = {Singapore},
  author = {Lin, Victoria  and
Morency, Louis-Philippe  and
Dimitriadis, Dimitrios  and
Sharma, Srinagesh},
  booktitle = {Findings of the Association for Computational Linguistics: EMNLP 2023},
  doi = {10.18653/v1/2023.findings-emnlp.43},
  editor = {Bouamor, Houda  and
Pino, Juan  and
Bali, Kalika},
  month = {December},
  pages = {592--606},
  publisher = {Association for Computational Linguistics},
  title = {Counterfactual Augmentation for Multimodal Learning Under Presentation Bias},
  url = {https://aclanthology.org/2023.findings-emnlp.43/},
  year = {2023}
}

@inproceedings{baek-etal-2025-researchagent,
  address = {Albuquerque, New Mexico},
  author = {Baek, Jinheon  and
Jauhar, Sujay Kumar  and
Cucerzan, Silviu  and
Hwang, Sung Ju},
  booktitle = {Proceedings of the 2025 Conference of the Nations of the Americas Chapter of the Association for Computational Linguistics: Human Language Technologies (Volume 1: Long Papers)},
  doi = {10.18653/v1/2025.naacl-long.342},
  editor = {Chiruzzo, Luis  and
Ritter, Alan  and
Wang, Lu},
  isbn = {979-8-89176-189-6},
  month = {April},
  pages = {6709--6738},
  publisher = {Association for Computational Linguistics},
  title = {{R}esearch{A}gent: Iterative Research Idea Generation over Scientific Literature with Large Language Models},
  url = {https://aclanthology.org/2025.naacl-long.342/},
  year = {2025}
}

@inproceedings{popov-etal-2025-refined,
  address = {Suzhou, China},
  author = {Popov, Dmitry  and
Negodin, Vladislav  and
Enikeeva, Ekaterina  and
Matrosova, Iana  and
Karpachev, Nikolay  and
Ryabinin, Max},
  booktitle = {Findings of the Association for Computational Linguistics: EMNLP 2025},
  doi = {10.18653/v1/2025.findings-emnlp.1203},
  editor = {Christodoulopoulos, Christos  and
Chakraborty, Tanmoy  and
Rose, Carolyn  and
Peng, Violet},
  isbn = {979-8-89176-335-7},
  month = {November},
  pages = {22079--22095},
  publisher = {Association for Computational Linguistics},
  title = {Refined Assessment for Translation Evaluation: Rethinking Machine Translation Evaluation in the Era of Human-Level Systems},
  url = {https://aclanthology.org/2025.findings-emnlp.1203/},
  year = {2025}
}

@inproceedings{shayegh-etal-2025-feeding,
  address = {Suzhou, China},
  author = {Shayegh, Behzad  and
Peter, Jan-Thorsten  and
Vilar, David  and
Domhan, Tobias  and
Juraska, Juraj  and
Freitag, Markus  and
Mou, Lili},
  booktitle = {Proceedings of the Tenth Conference on Machine Translation},
  doi = {10.18653/v1/2025.wmt-1.16},
  editor = {Haddow, Barry  and
Kocmi, Tom  and
Koehn, Philipp  and
Monz, Christof},
  isbn = {979-8-89176-341-8},
  month = {November},
  pages = {269--285},
  publisher = {Association for Computational Linguistics},
  title = {Feeding Two Birds or Favoring One? Adequacy{--}Fluency Tradeoffs in Evaluation and Meta-Evaluation of Machine Translation},
  url = {https://aclanthology.org/2025.wmt-1.16/},
  year = {2025}
}

@inproceedings{ye2026problemoriented,
  author = {Ye, Tong and Ma, Tengfei and Zhang, Xuhong and Yu, Hang and Yin, Jianwei and Wang, Wenhai},
  booktitle = {International Conference on Learning Representations},
  editor = {C. Vondrick and B. Hariharan and C. Raffel and L. Pinto and D. Yang and A. Faust},
  pages = {47925--47957},
  title = {A Problem-Oriented Perspective and Anchor Verification for Code Optimization},
  url = {https://proceedings.iclr.cc/paper_files/paper/2026/file/4f5aa418348643f2041a55bba9afa7dc-Paper-Conference.pdf},
  volume = {2026},
  year = {2026}
}

@inproceedings{min-etal-2025-towards,
  address = {Vienna, Austria},
  author = {Min, Hyangsuk  and
Lee, Yuho  and
Ban, Minjeong  and
Deng, Jiaqi  and
Kim, Nicole Hee-Yeon  and
Yun, Taewon  and
Su, Hang  and
Cai, Jason  and
Song, Hwanjun},
  booktitle = {Proceedings of the 63rd Annual Meeting of the Association for Computational Linguistics (Volume 1: Long Papers)},
  doi = {10.18653/v1/2025.acl-long.702},
  editor = {Che, Wanxiang  and
Nabende, Joyce  and
Shutova, Ekaterina  and
Pilehvar, Mohammad Taher},
  isbn = {979-8-89176-251-0},
  month = {July},
  pages = {14417--14450},
  publisher = {Association for Computational Linguistics},
  title = {Towards Multi-dimensional Evaluation of {LLM} Summarization across Domains and Languages},
  url = {https://aclanthology.org/2025.acl-long.702/},
  year = {2025}
}

@inproceedings{lu-etal-2026-rethinking,
  address = {Rabat, Morocco},
  author = {Lu, Li-Chun  and
Liu, Miri  and
Lu, Pin Chun  and
Tian, Yufei  and
Sun, Shao-Hua  and
Peng, Nanyun},
  booktitle = {Proceedings of the 19th Conference of the {E}uropean Chapter of the {A}ssociation for {C}omputational {L}inguistics (Volume 1: Long Papers)},
  doi = {10.18653/v1/2026.eacl-long.297},
  editor = {Demberg, Vera  and
Inui, Kentaro  and
Marquez, Llu{\'i}s},
  isbn = {979-8-89176-380-7},
  month = {March},
  pages = {6329--6352},
  publisher = {Association for Computational Linguistics},
  title = {Rethinking Creativity Evaluation: A Critical Analysis of Existing Creativity Evaluations},
  url = {https://aclanthology.org/2026.eacl-long.297/},
  year = {2026}
}

@inproceedings{NEURIPS2024_be2e1b68,
  author = {Kirk, Hannah Rose and Whitefield, Alexander and R\"{o}ttger, Paul and Bean, Andrew and Margatina, Katerina and Ciro, Juan and Mosquera, Rafael and Bartolo, Max and Williams, Adina and He, He and Vidgen, Bertie and Hale, Scott},
  booktitle = {Advances in Neural Information Processing Systems},
  doi = {10.52202/079017-3342},
  editor = {A. Globerson and L. Mackey and D. Belgrave and A. Fan and U. Paquet and J. Tomczak and C. Zhang},
  pages = {105236--105344},
  publisher = {Curran Associates, Inc.},
  title = {The PRISM Alignment Dataset: What Participatory, Representative and Individualised Human Feedback Reveals About the Subjective and Multicultural Alignment of Large Language Models},
  url = {https://proceedings.neurips.cc/paper_files/paper/2024/file/be2e1b68b44f2419e19f6c35a1b8cf35-Paper-Datasets_and_Benchmarks_Track.pdf},
  volume = {37},
  year = {2024}
}

@article{ouyang2026dscodebench,
  author = {Ouyang, Shuyin and HUANG, Dong and Guo, Jingwen and Sun, Zeyu and Zhu, Qihao and Zhang, Jie M.},
  doi = {10.1609/aaai.v40i38.40540},
  issn = {2159-5399},
  journal = {Proceedings of the AAAI Conference on Artificial Intelligence},
  month = {March},
  number = {38},
  pages = {32628--32636},
  publisher = {Association for the Advancement of Artificial Intelligence (AAAI)},
  title = {DSCodeBench: A Realistic Benchmark for Data Science Code Generation},
  url = {http://dx.doi.org/10.1609/aaai.v40i38.40540},
  volume = {40},
  year = {2026}
}

@misc{gemmateam2026gemma4,
  author = {{Gemma Team}},
  howpublished = {arXiv preprint arXiv:2607.02770},
  title = {{Gemma 4 Technical Report}},
  url = {https://arxiv.org/abs/2607.02770},
  year = {2026}
}

@article{team2026qwen3,
  title={Qwen3. 5-omni technical report},
  author={{Qwen Team}},
  journal={arXiv preprint arXiv:2604.15804},
  year={2026}
}

@misc{qwen2024qwen25,
  author = {{Qwen Team}},
  howpublished = {arXiv preprint arXiv:2412.15115},
  title = {{Qwen2.5 Technical Report}},
  url = {https://arxiv.org/abs/2412.15115},
  year = {2025}
}

@misc{openai2025gptoss,
  author = {{OpenAI}},
  howpublished = {arXiv preprint arXiv:2508.10925},
  title = {{gpt-oss-120b \& gpt-oss-20b Model Card}},
  url = {https://arxiv.org/abs/2508.10925},
  year = {2025}
}

@inproceedings{vedula-etal-2025-quantile,
  address = {Vienna, Austria},
  author = {Vedula, Nikhita  and
Dhyani, Dushyanta  and
Jalali, Laleh  and
Oreshkin, Boris N.  and
Bayati, Mohsen  and
Malmasi, Shervin},
  booktitle = {Findings of the Association for Computational Linguistics: ACL 2025},
  doi = {10.18653/v1/2025.findings-acl.641},
  editor = {Che, Wanxiang  and
Nabende, Joyce  and
Shutova, Ekaterina  and
Pilehvar, Mohammad Taher},
  isbn = {979-8-89176-256-5},
  month = {July},
  pages = {12396--12415},
  publisher = {Association for Computational Linguistics},
  title = {Quantile Regression with Large Language Models for Price Prediction},
  url = {https://aclanthology.org/2025.findings-acl.641/},
  year = {2025}
}

@article{grattafiori2024llama,
  title = {The {Llama 3} Herd of Models},
  author = {Grattafiori, Aaron and Dubey, Abhimanyu and Jauhri, Abhinav and Pandey, Abhinav and Kadian, Abhishek and Al-Dahle, Ahmad and Letman, Aiesha and Mathur, Akhil and Schelten, Alan and Vaughan, Alex and others},
  journal = {arXiv preprint arXiv:2407.21783},
  year = {2024},
  url = {https://arxiv.org/abs/2407.21783}
}
